\documentclass{article}

\usepackage{arxiv}

\usepackage[utf8]{inputenc}
\usepackage{amsmath}
\usepackage{mathrsfs}
\usepackage{amsmath,bm}
\usepackage{amssymb}
\usepackage{bbm}
\usepackage{graphicx}
\usepackage{esint}
\usepackage{enumitem}
\usepackage{amsthm}
\usepackage{amsfonts}
\usepackage{booktabs}
\usepackage{subfigure}
\usepackage{epstopdf}
\usepackage[svgnames]{xcolor}
\usepackage{listings}
\usepackage{mathtools}
\usepackage{authblk}
\usepackage[ruled,vlined]{algorithm2e}
\usepackage[colorlinks=true,linkcolor=blue,citecolor=blue]{hyperref}
\usepackage[sort, numbers]{natbib}
\usepackage{tikz}
\usetikzlibrary{shapes,arrows,positioning,fit,calc}
\tikzset{block/.style={draw, thick, text width=3cm, minimum height=1.5cm, align=center},   
line/.style={-latex}   
}  
\tikzstyle{block} = [rectangle, draw, 
    text width=5em, text centered, rounded corners, minimum height=4em]
\tikzstyle{line} = [draw, -latex']

\SetKwInput{kwInit}{initilize}
\title{Multifidelity data fusion in convolutional encoder/decoder networks}
\author[1]{Lauren Partin}
\author[2]{Gianluca Geraci}
\author[3]{Ahmad Rushdi}
\author[2]{Michael S. Eldred}
\author[1]{Daniele E. Schiavazzi}
\affil[1]{Department of Applied and Computational Mathematics and Statistics\protect\\University of Notre Dame, Notre Dame, IN, USA}
\affil[2]{Center for Computing Research, Optimization and UQ Group\protect\\Sandia National Laboratories, Albuquerque, NM, USA}
\affil[3]{Institute for Human-Centered Artificial Intelligence\protect\\Stanford University, Stanford, CA, USA}
\date{ }

\begin{document}
\maketitle


\begin{abstract}
\noindent We analyze the regression accuracy of convolutional neural networks assembled from encoders, decoders and skip connections and trained with multifidelity data. 
Besides requiring significantly less trainable parameters than equivalent fully connected networks, encoder, decoder, encoder-decoder or decoder-encoder architectures can learn the mapping between inputs to outputs of arbitrary dimensionality.
We demonstrate their accuracy when trained on a few high-fidelity and many low-fidelity data generated from models ranging from one-dimensional functions to Poisson equation solvers in two-dimensions. 
We finally discuss a number of implementation choices that improve the reliability of the uncertainty estimates generated by Monte Carlo DropBlocks, and compare uncertainty estimates among low-, high- and multifidelity approaches.
\end{abstract}

\section{Introduction} \label{LMP:sec:intro}

\noindent Analyzing physical phenomena through their mathematical or numerical modeling is a common practice in engineering and science, providing the analyst with the ability to predict the behavior of a system outside of a limited number of observations.
Simulation of complex phenomena, for example characterized by multiple interacting physics, may require a substantial computational effort, and the availability of sufficient resources may be a key factor in the ability to answer the scientific questions of interest. 
However, it is often possible to combine accurate but expensive high-fidelity simulations with lower-fidelity simulations that provide approximations at a reduced cost, in order to optimize efficiency while retaining accuracy.

This study focuses on generating multifidelity (MF) surrogate models designed to combine information from a few high-fidelity (HF) model solutions and many low-fidelity (LF) approximations of varying accuracy. More specifically, we focus on data-driven multifidelity surrogates in the machine learning context.
Previous work considered student-teacher networks with the ability to handle datasets with variable annotation quality~\cite{dehghani2017fidelity}, surrogates trained using transfer learning between two model fidelities~\cite{de2020transfer}, and fully connected neural networks combining three sub-networks designed to learn a LF representation, the correlation between a LF and a HF representation, and to minimize a physics-based residual loss~\cite{meng2020composite}. 
Other approaches utilize Bayesian neural networks~\cite{meng2021multi}, or combine convolutional and fully connected networks to learn the discrepancy between increasingly fine discretizations of a given PDE solution, projected on a common mesh~\cite{van2020multi}.

Our approach is inspired by the recent successes in image classification and segmentation tasks shown by deep convolutional encoder-decoder networks (see, e.g.,~\cite{zhu2018bayesian,minaee2021image}). While multifidelity data fusion has been mainly demonstrated for fully connected networks or for ensembles of hybrid convolutional and fully connected networks~\cite{van2020multi}, no approach has focused on convolutional networks assembled from encoders, decoders and skip connections, where the model fidelities are learned simultaneously, following an \emph{all-at-once} training paradigm. 
Convolutions are essential to reduce the number of weights with respect to fully connected networks when the input, the output or both are high-dimensional, as discussed in our recent work~\cite{PartinCSRI2021,partin2022multifidelity}.

We also focus on quantifying the {\em predictive uncertainty} in the network outputs, i.e., we want to characterize the variability of the predicted quantities of interest, a paradigm commonly referred to as ``UQ for ML'', analyzing the uncertainty in predictions that are inherent when using a machine-learned surrogate model.
We consider this as a \emph{model form} uncertainty that relates to how the information flows through the selected multifidelity network, as opposed to other paradigms where a deterministic machine learning model is employed as an inexpensive surrogate for uncertainty quantification studies (referred to as ``ML for UQ'').

Many different approaches have been proposed in the literature to quantify predictive uncertainty in neural network outputs~\cite{hullermeier2021aleatoric,kendall2017uncertainties,abdar2020review}.  
Among these, dropout layers~\cite{srivastava2014dropout} offer a simple and computationally appealing solution to drop neurons at random, providing, at the same time, regularization and variance estimates. Their interpretation in terms of an ensemble of network architectures has also been investigated in the literature~\cite{baldi2013understanding,gal2016dropout}. 
However, their performance has been mainly assessed on neural networks with fully connected layers. In this study, we use DropBlocks~\cite{ghiasi2018dropblock}, i.e., adaptations of dropout layers showing improved performance on convolutional architectures.
This study extends previous results from our research group in two directions. In~\cite{PartinCSRI2021,partin2022multifidelity}, we focused on two separate questions, i.e., the problem of identifying network hyperparameters leading to optimal accuracy, and the problem of understanding the effect of hyperparameter selection on the variability of network predictions. Here we unify these two perspectives by studying networks providing the best trade off between accuracy and uncertainty. In addition, we show new results for the characterization of uncertainty in the one-dimensional and low- to high-dimensional test cases.

This paper is organized as follows: Section~\ref{LMP:sec:problems} introduces the problems of interest including one-dimensional function approximation from MF datasets, and prediction of high-dimensional responses from computational fluid dynamics solvers.
Section~\ref{LMP:sec:nn_arch} introduces the convolutional network architectures used in the study.
Uncertainty estimates through Monte Carlo DropBlock are discussed in Section~\ref{LMP:sec:uncertainty}.
Results are summarized in Section~\ref{LMP:sec:results}, while conclusions and future work are finally discussed in Section~\ref{LMP:sec:conclusionsfuture}.
For the interested reader, implementation details are reported in the appendix.

\section{Problem description} \label{LMP:sec:problems}

\noindent We study the approximation performance of multifidelity networks on three different problem instances. 
We begin with function approximation, where we show how a convolutional architecture can be designed to accurately learn a map between inputs and outputs, even in a single dimension.
We then consider \emph{dense regression} problems where inputs and outputs are images of the same size (i.e., having equal dimensionality).
Finally, we present a multifidelity architecture for \emph{low- to high-dimensional regression}, where a high-dimensional output is predicted from a low-dimensional input.
For all these cases, we examine how low-fidelity representations can be leveraged to accelerate training and improve the accuracy of high-fidelity predictions from limited data.

\subsection{One-dimensional multifidelity function approximation}\label{sec:problem_ld}

\begin{figure}[!ht]
\centering
\subfigure[]{\includegraphics[width=0.45\textwidth]{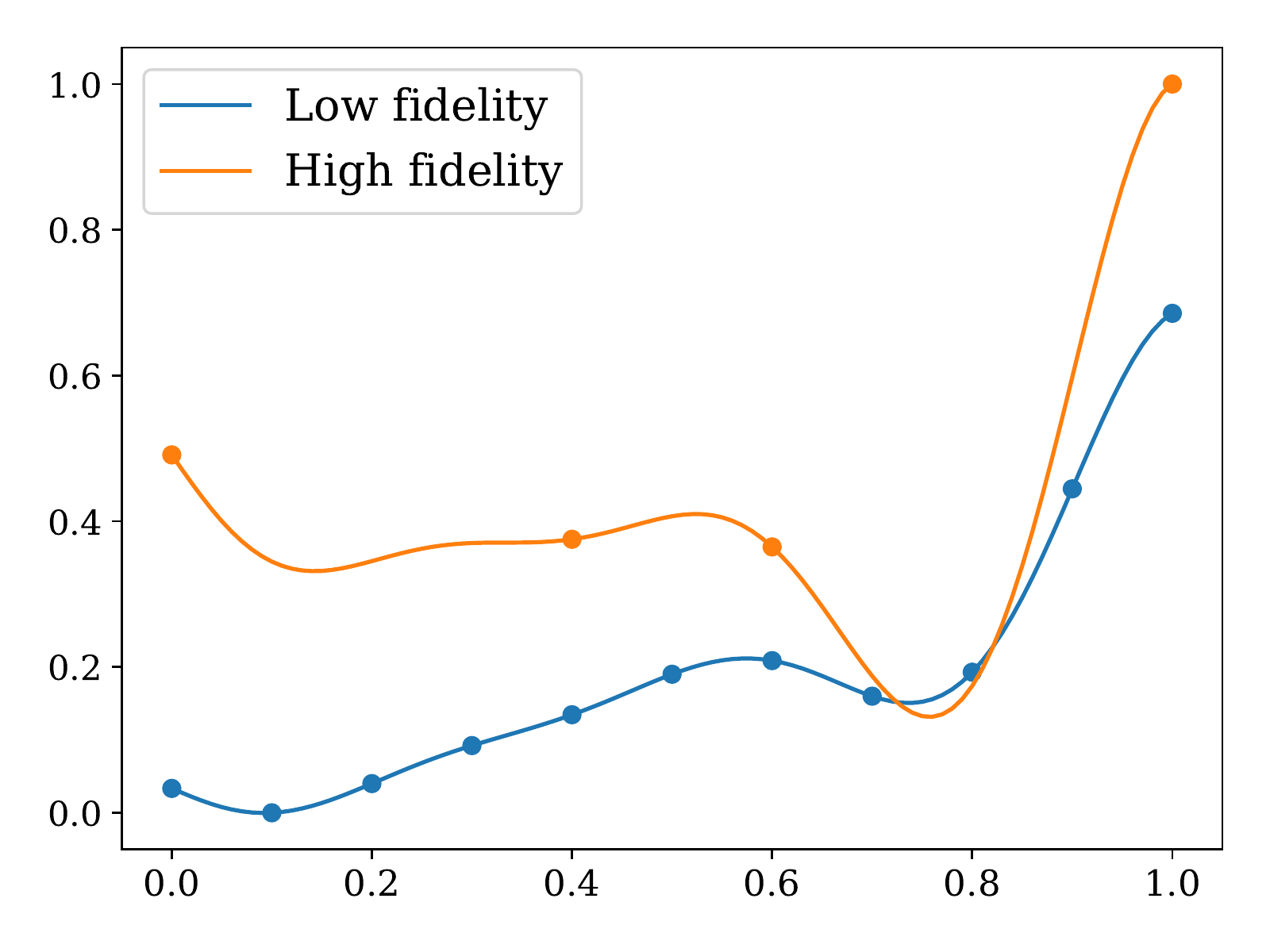}\label{LMP:fig:ex1_dataset}}
\subfigure[]{\includegraphics[width=0.45\textwidth]{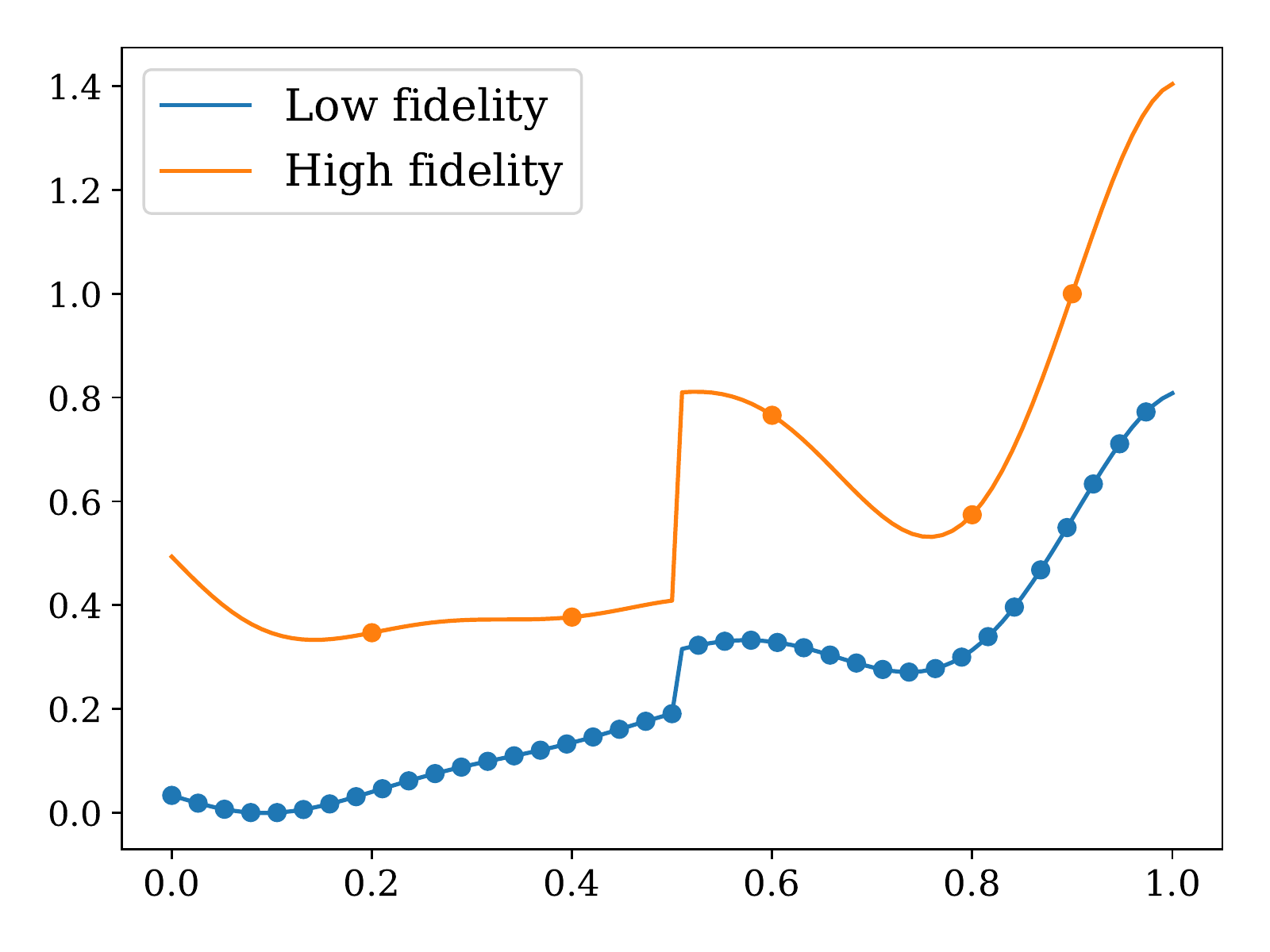}\label{LMP:fig:ex2_dataset}}
\caption{LF and HF functions and respective training locations from (a) Eqs.~\eqref{LMP:equ:example1_lf}, ~\eqref{LMP:equ:example1_hf} and (b) Eqs.~\eqref{LMP:equ:example2_lf}, \eqref{LMP:equ:example2_hf}. Function values are rescaled such that $y_H(x), y_L(x) \in [0,1]$ for $x$ in the training set.}\label{LMP:fig:datasets_1d}
\end{figure}

\noindent We consider two examples, each consisting of two correlated LF and HF functions, with very few available HF and relatively more LF training examples~\cite{meng2020composite}.
The first example consists of two linearly correlated continuous functions, defined as
\begin{eqnarray}
    y_L(x) &=& (1/2)(6x - 2)^2\sin (12x - 4) + 10(x - 1/2) - 5 \label{LMP:equ:example1_lf} \\
    y_H(x) &=& (6x - 2)^2\sin (12x - 4) \label{LMP:equ:example1_hf},
\end{eqnarray}
where $11$ and $4$ samples, are provided for $y_L$ and $y_H$, respectively, as shown in Fig.~\ref{LMP:fig:ex1_dataset}.
In the second example, we consider two linearly correlated discontinuous functions expressed as
\begin{eqnarray}
y_L(x) &=& 
\begin{cases}
l(x) = 0.5(6x-2)^2\sin(12x-4) + 10(x-0.5) - 5 & 0 \leq x \leq 0.5 \\
3 + l(x) & 0.5 < x \leq 1 \\
\end{cases}
\label{LMP:equ:example2_lf}\\
y_H(x) &=& 
\begin{cases}
h(x) = 2 y_L(x) - 20x + 20 & 0 \leq x \leq 0.5 \\
4 + h(x) & 0.5 < x \leq 1,\\
\end{cases} \label{LMP:equ:example2_hf} 
\end{eqnarray}
with $38$ and $5$ training samples for $y_L$ and $y_H$, respectively, as shown in Fig.~\ref{LMP:fig:ex2_dataset}. 
We only consider linear correlations in these examples, since these suffice to show the performance of the proposed network for one-dimensional problems and provide a point of comparison with existing literature~\cite{meng2020composite}. 
However, this is not explored further, since fully-connected and convolutional architectures contain a similar number of weights for one-dimensional regression problems, and therefore convolutional networks provide no practical computational advantage.

\subsection{Dense regression} \label{sec:problem_hd}

\noindent For the high-dimensional dense regression case, we focus on an application in computational fluid dynamics, where we are interested in predicting the pressure distribution in a fluid domain $\Omega_{f}$ from a noisy binary mask  and its velocities. The binary mask identifies the fluid region and is referred to as the scalar \emph{concentrations}.
The pressure, up to a constant, can be computed through a reformulation of the incompressible Navier Stokes equations as a Poisson pressure equation with appropriate boundary conditions~\cite{schiavazzi2017effect}. This approach, however, may require the solution of a partial differential equation on a large computational grid and training a neural network could provide a much faster and computationally attractive alternative.
Note that, unlike physics-informed neural networks (PINN~\cite{raissi2019physics}), we would like to learn a relation between high-dimensional concentration/velocity inputs and high-dimensional pressure outputs (unlike the one-dimensional problem discussed in Section~\ref{sec:problem_ld}), rather than pressure as a function of space and time.

Under the assumption that no body force is acting on the fluid, the Poisson equation for the pressure $p$ can be written as
\begin{equation}
\Delta p = \boldsymbol{\nabla}\cdot\mathbf{f} = \boldsymbol{\nabla}\cdot\left[ -\rho\,\left(\dfrac{\partial\,\mathbf{u}}{\partial t} + \mathbf{u}\cdot\nabla \mathbf{u}\right) + \mu\,\Delta\mathbf{u}\right],
\end{equation}
%
%
and only Neumann boundary conditions are applied equal to the flux of $\mathbf{f}$ across the smooth boundary $\partial\Omega_{f}$ of $\Omega_{f}$~\cite{schiavazzi2017effect}.

The problem is discretized through a structured grid $\Omega$ with $m_{x}\cdot m_{y}\cdot m_{z}=m$ cells, where the cells belonging to the fluid region are identified by components equal to 1 in a binary \emph{concentration} array $c_{b}\in\{0,1\}^{m}$. However, in practice, noise makes the concentration non-binary, with \emph{measured} concentration expressed as $c\in\mathbb{R}^{m}$. Our objective is to train a neural network surrogate to quickly evaluate the map
\begin{equation}
f(\mathbf{c},\mathbf{u}) = \mathbf{p},\,\,\text{where}\,\,f: \mathbb{R}^{m}\times \mathbb{R}^{m\times 3}\rightarrow\mathbb{R}^{m},
\end{equation}
which, given the noisy concentrations and velocity distributions over $\Omega$, returns the spatial pressure distribution on $\Omega_{f}$. 
Our main goal in this paper is to investigate the possibility to increase training efficiency and accuracy using a multifidelity dataset containing pressure representations with increasingly coarser resolution. 
A multifidelity training set is obtained by combining a small number of HF examples, resulting from a Poisson pressure equation finite element solver, with a large number of solutions from the same solver, but evaluated on coarser discretizations. 
Although the parabolic velocities and linear relative pressures associated with Poiseuille flow represent a rather smooth field to be emulated by the network, and therefore may appear to offer a rather simplistic test case, we note how the change in the flow domain introduces a fair amount of complexity. 
This study therefore focuses on the development of highly accurate U-Net-like convolutional architectures for this flow, and provides an initial exploration of the method's robustness to noise and bias in the low-fidelity approximants.
Future work will focus on quantifying the performance of the proposed approach on more complex flow configurations.

\subsection{Low- to high-dimensional regression}\label{sec:problem_ld_hd}

\begin{figure}[!ht]
\centering
\subfigure[]{\includegraphics[width=0.25\textwidth]{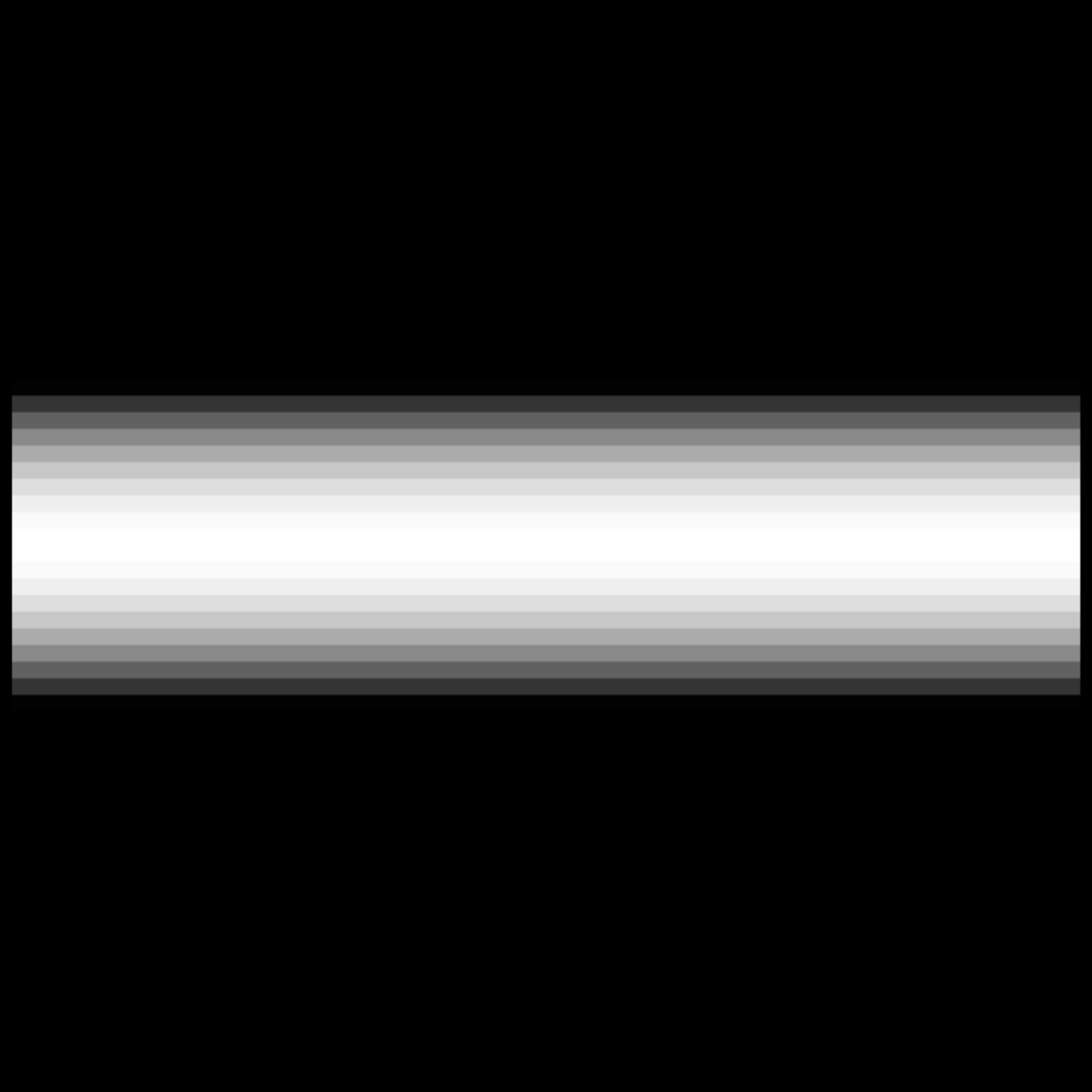}}$\quad$
\subfigure[]{\includegraphics[width=0.25\textwidth]{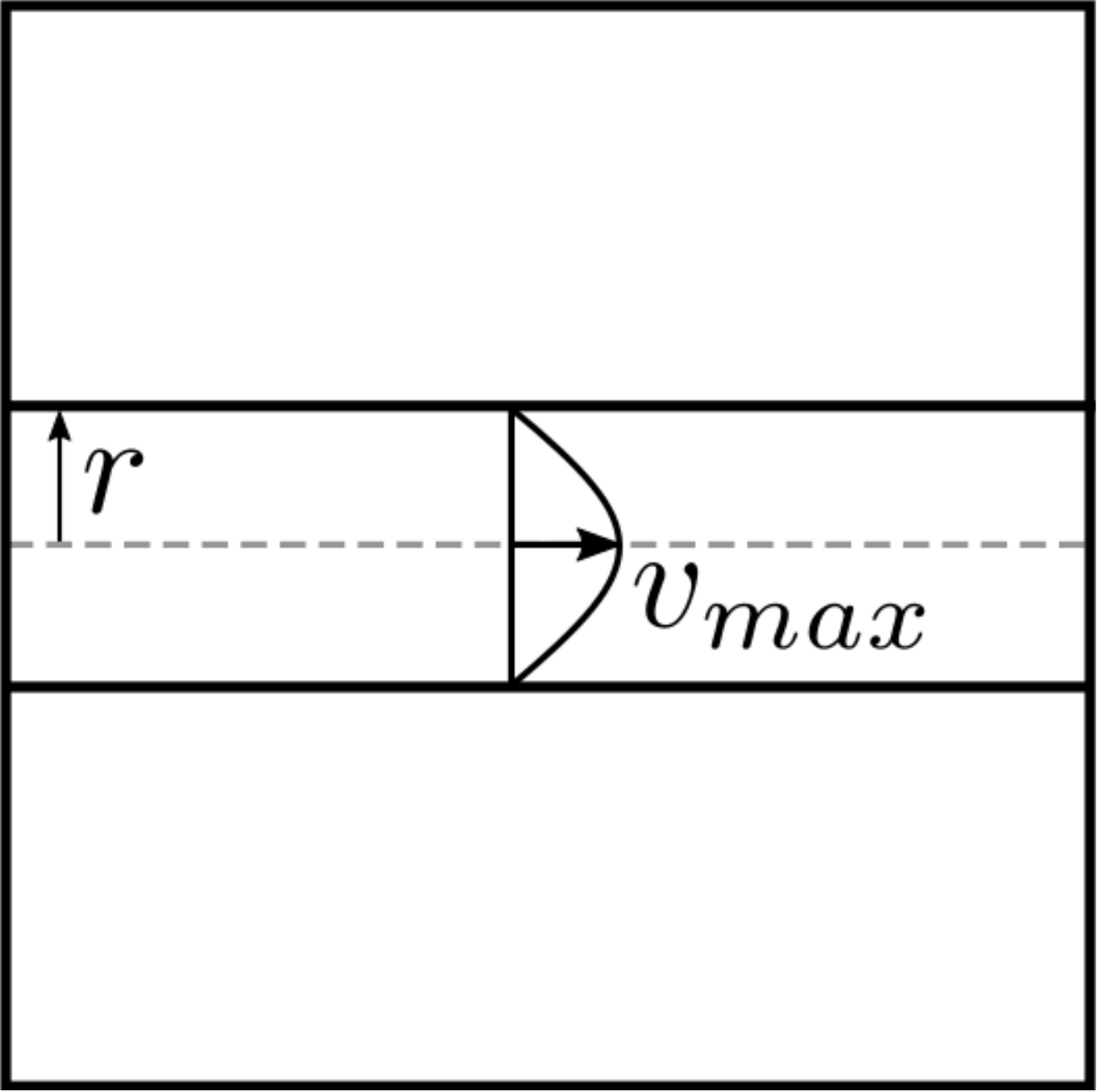}\label{LMP:fig:poiseuille_testcase_2}}$\quad$
\subfigure[]{\includegraphics[width=0.25\textwidth]{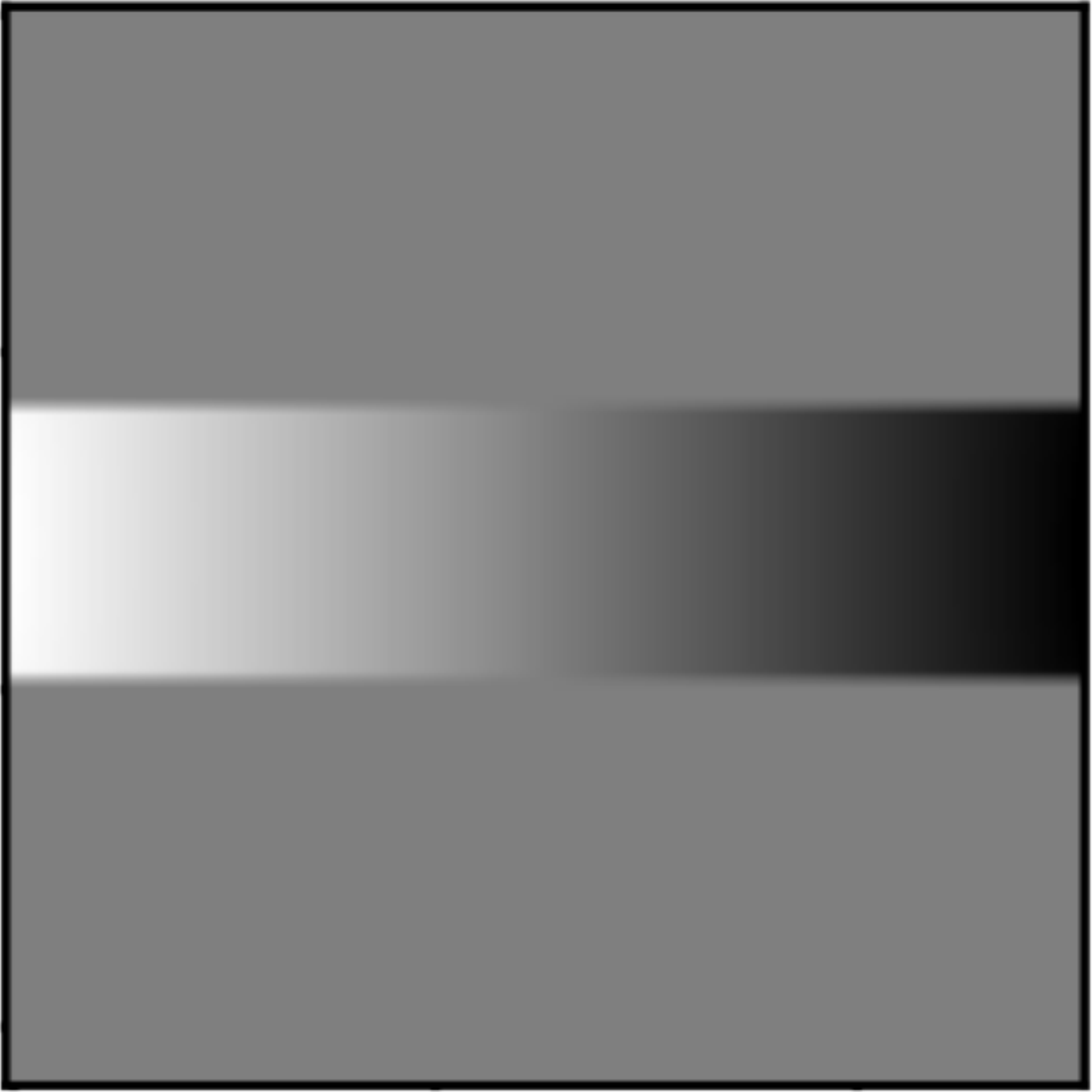}}
\caption{Example of Poiseuille flow. Velocity profile (a). Test case parameterization in terms of the fluid region radius and maximum velocity (b). Pressure result (c).}
\label{LMP:fig:poiseuille_testcase}
\end{figure}

\noindent We consider a problem with the same pressure outputs as that in Section~\ref{sec:problem_hd}, but with only two inputs, i.e., the radius $r$ of the cylindrical fluid domain and the maximum velocity $v_{max}$, as these parameters are sufficient to fully specify the geometry and velocity distribution in a Hagen-Poiseuille flow, as shown in Fig.~\ref{LMP:fig:poiseuille_testcase_2}.
In other words, this case represents the surrogate construction for a random field (the pressure) given only two input parameters. 
This low-to-high dimensional regression problem may not be easy to solve for traditional surrogate-based approaches. For example, techniques like generalized polynomial chaos (gPC~\cite{xiu2002wiener}) could be used to obtain scalar pressure estimates at each pixel in the fluid domain, but additional structure (e.g. \emph{modes} informed from dimensionality reduction algorithms) would need to be specified to account for the spatial correlation of the resulting pressure field.

\section{Network architectures}\label{LMP:sec:nn_arch}

\noindent For all problems discussed above, we employ a network architecture assembled from convolutional encoders, decoders and skip connections.
A convolutional encoder~\cite{lecun1989backpropagation} is composed of alternating layers of convolutions and pooling (i.e., downsampling), generating a compressed feature representation.
A convolutional decoder, on the other hand, is composed of alternating layers of convolutions and upsampling. 
Skip connections are finally added to mitigate the loss of information due to downsampling, and counteract the vanishing gradient problem (see, e.g.~\cite{li2018visualizing}).
For dense regression problems, the encoder and decoder are symmetric, and padding is applied so that the number of pixels in the network input and output is the same.

In the next sections, we describe the three specific network architectures used to address the problems presented in Section~\ref{sec:problem_ld}, Section~\ref{sec:problem_hd} and Section~\ref{sec:problem_ld_hd}, respectively.
Section~\ref{LMP:sec:uncertainty} provides an overview of DropBlock layers and their use in uncertainty quantification.
In addition, Section~\ref{LMP:sec:mf_coupling} discusses how the information from low- and high-fidelity models is assembled in a network.

\subsection{Decoder-encoder architecture for one-dimensional regression}\label{sec:low_dim_network}

\noindent For one-dimensional regression (Section~\ref{sec:problem_ld}), a single LF predictor (as opposed to multiple LF predictors for the networks discussed in the next sections) is generated by the network. 
Denote $z$ as the concatenation between the LF predictor and its $x$ coordinate. The HF predictor is obtained by summing two contributions
\begin{enumerate}
\item a convolution applied to $z$, designed to capture the linear correlation between the HF and LF outputs; and
\item the convolution of $z$ in two layers which are separated by a nonlinear activation, designed to capture the nonlinear correlation between the HF and LF outputs.
\end{enumerate}
Additive skip connections are used to facilitate the exchange of information between the decoder and the encoder. 
The network layout is illustrated in Fig.~\ref{LMP:fig:low_multifidelity_network}.
Additional details are included, for the interested reader, in the appendix. 
\begin{figure}[!ht]
    \centering
    \includegraphics[width=\textwidth]{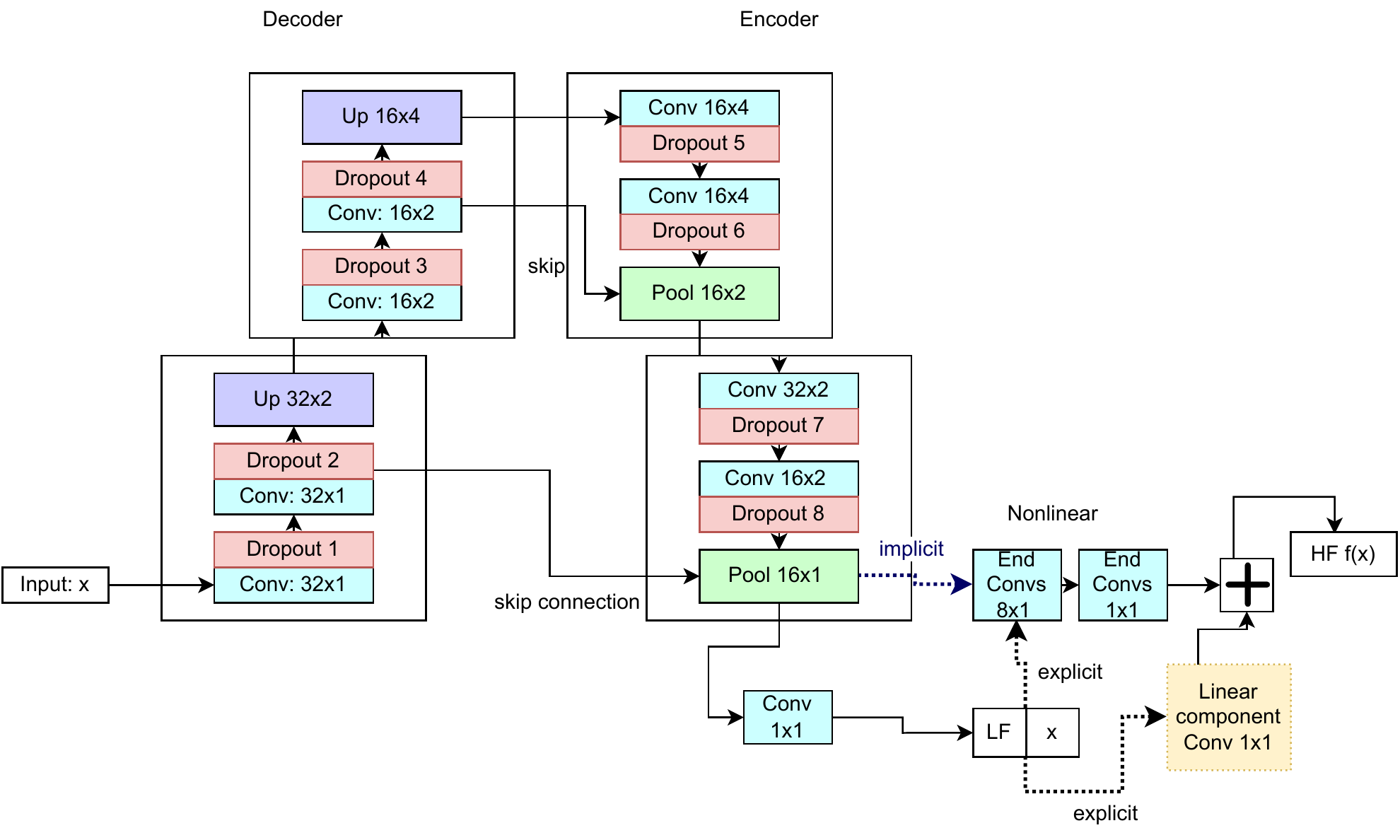}
    \caption{Multifidelity decoder-encoder convolutional network architecture for one-dimensional regression. The HF output is a linear combination of the nonlinear and linear portion of the network, $\text{HF} = \gamma\cdot\text{Linear}(\text{LF}, x) + (1-\gamma)\cdot\text{Nonlinear}(\text{LF}, x)$, where $\gamma$ is a parameter learned during training.}
    \label{LMP:fig:low_multifidelity_network}
\end{figure}

\subsection{Encoder-decoder architecture for dense regression}\label{sec:high_dim_network}

\noindent This architecture resembles the popular U-Net~\cite{ronneberger2015u} that has shown remarkable performance in terms of accuracy and training speed for segmentation tasks, even under limited training data~\cite{isensee2021nnu}.
We consider both input and output images with $64 \times 64$ pixels, where the input is characterized by three channels (one concentration and two velocity components) and the output by a single channel (the pressure).
A simple identity replaces the ReLU activation after the final convolution layer.

\begin{figure}[!ht]
\centering
\includegraphics[width=\textwidth]{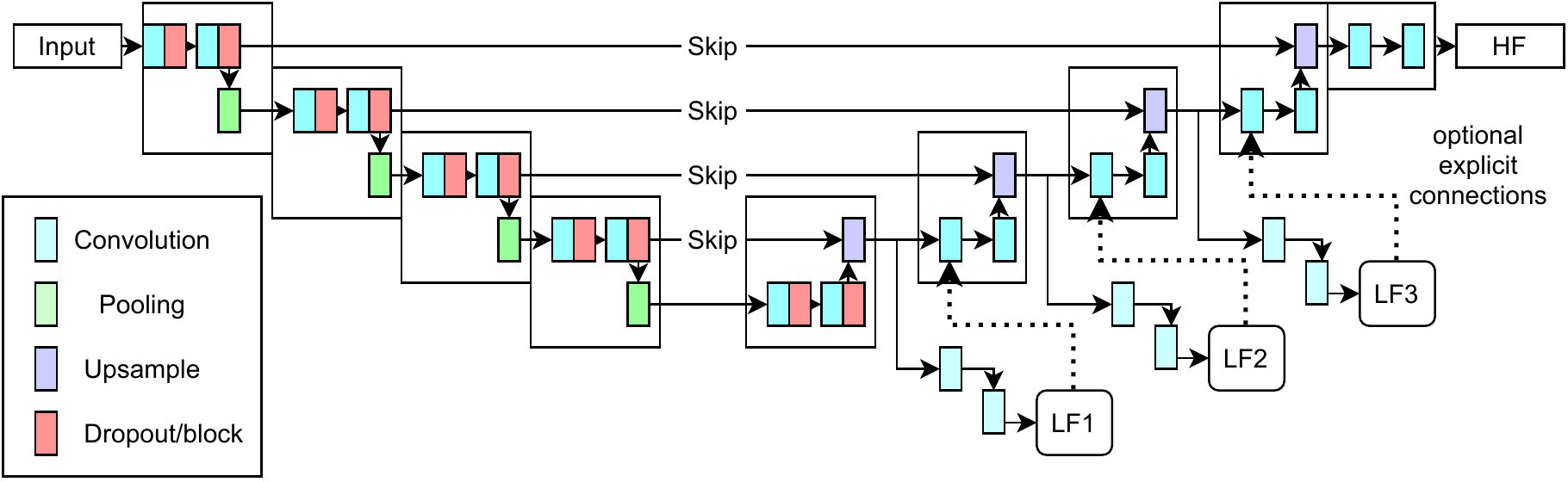}
\caption{Multifidelity decoder-encoder convolutional network architecture for high-dimensional dense regression.}
\label{LMP:fig:multifidelity_network_2d}
\end{figure}

A multifidelity network is obtained by extracting a LF representation of increasing resolution at each stage of the decoder. 
A term for each LF predictor is then added to the loss function, so these LF representations are accurately learned. This is depicted in Fig.~\ref{LMP:fig:multifidelity_network_2d} where the models are ordered as LF1, LF2, LF3, HF, i.e., from the coarsest to the finest resolutions.
For the interested reader, additional details on this network are included in the appendix.

\subsection{Decoder architecture for low- to high-dimensional regression}
\label{sec:lowtohigh_dim_network}

\begin{figure}[!ht]
    \centering
    \includegraphics[width=\textwidth]{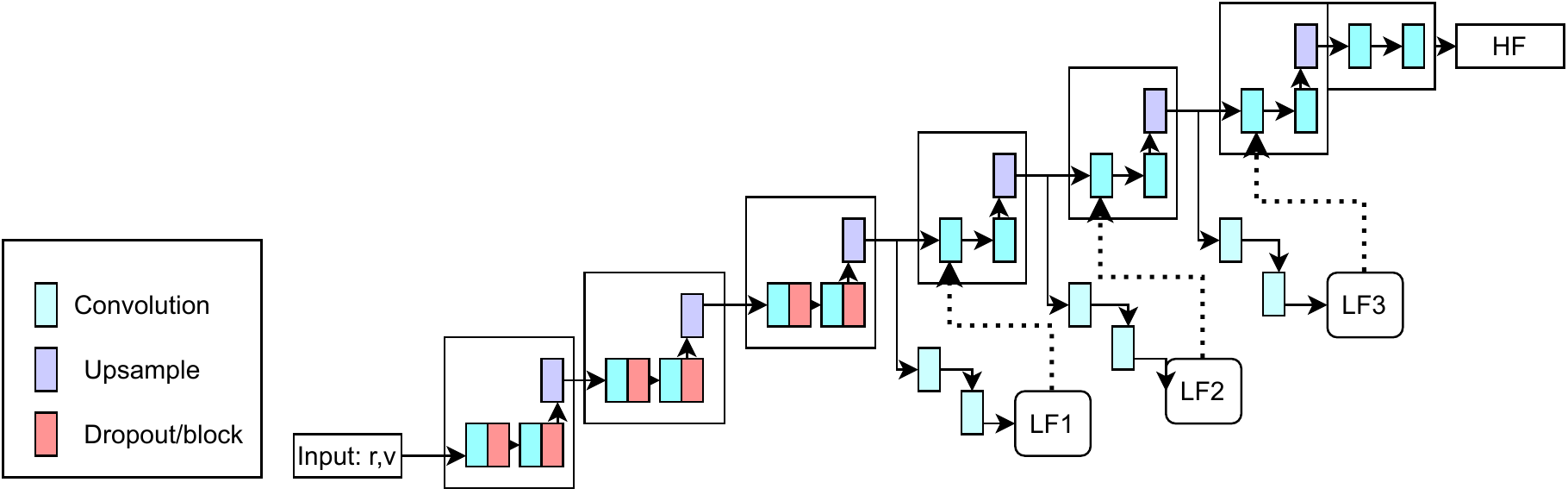}
    \caption{Multifidelity convolutional decoder architecture for low- to high-dimensional regression.}
    \label{LMP:fig:multifidelity_network_decoder}
\end{figure}

\noindent The network selected for low- to high-dimensional regression is shown in Fig.~\ref{LMP:fig:multifidelity_network_decoder}. 
The two dimensional input is upsampled to generate $64 \times 64$ output images. 
This is achieved with a single decoder, choosing the network depth and padding (see appendix) to enforce the correct output dimensions, i.e. 6 upsampling layers, each with a scale factor of 2. 

Similar to the network in the previous section, a LF prediction is generated at each decoder stage, ordered as LF1, LF2, LF3, HF, i.e., from the coarsest to the finest resolution, respectively.
For the interested reader, additional details on this network are included in the appendix.

\subsection{DropBlock layers for regularization and prediction uncertainty}\label{LMP:sec:uncertainty}

\noindent Dropout layers~\cite{srivastava2014dropout} are a widely used form of regularization for artificial neural networks, designed to avoid overfitting. These layers operate by dropping neurons at random during training, so the learned representation relies only weakly on the correlations between neurons.
This continuous change in network connectivity can be interpreted as the simultaneous training of an ensemble of architectures at a fraction of the computational cost of processing them individually.
However, dropout layers are ineffective for convolutional neural networks due to the spatial correlation present in images, where relevant details consist of multiple correlated pixels.
For this reason, DropBlock~\cite{ghiasi2018dropblock} layers are designed to drop a continuous group of pixels.

As these layers are still parameterized in terms of drop probability $p$, the relation between $p$ and the actual ratio of features being dropped should be first clarified.
A Bernoulli mask is generated in~\cite{ghiasi2018dropblock}, using a probability $\gamma$ expressed as
\begin{equation}\label{LMP:equ:dropblock_gamma}
\gamma = \dfrac{p F^d} {b^d (F - b + 1)^d}\,,
\end{equation}
where $p$ is the drop probability, $F$ and $b$ are the feature and block size, respectively, and $d$ is the feature space dimensionality.
Note that the drop probability $p$ may not represent the actual percentage of elements dropped due to overlapping between blocks.
As discussed in the appendix, the actual drop ratio is closest to $p$ when $F=b$ (assuming no partial blocks are considered at the edges of the feature map); otherwise, we typically see a lower drop ratio due to overlapping blocks.
In ~\cite{ghiasi2018dropblock}, a distinction is made between the DropBlock mask being independent or shared across feature channels; we considered both approaches and chose the one producing the best accuracy for each test case; 
we also compared the implications of both choices in Section~\ref{sec:res_uq}.

When used during the \emph{evaluation} of an optimally trained network, DropBlock layers can also be used to \emph{inject stochasticity} in the network predictions, and, combined with Monte Carlo sampling, provide a tool to quantify output uncertainty. We refer to this technique as \emph{MC DropBlock}, which is similar, in principle, to the MC dropout approach discussed in~\cite{gal2016dropout}.
In this study, we consider network output \emph{ensembles} of size $N_{UQ}=1000$.
As noted in Section~\ref{LMP:sec:intro}, it is important to emphasize that this notion of uncertainty reflects the variability introduced in the network by the hyper-parameters (in this case changes in the network architecture induced by randomly dropping groups of features) and not the impact of any uncertainty either in the network weights (as in Bayesian neural networks, see~\cite{langseth2007bayesian, meng2021multi}) or its inputs.

We also wanted our network to promote accuracy in \emph{each} MC-DropBlock realization rather than only on their mean. 
To do so, we activated DropBlock layers both in training and when \textit{evaluating} the network (e.g. when calculating the validation loss).
This is in contrast with the practice of keeping these layers off (i.e., drop probability $p=0$) when generating network outputs, commonly adopted when using DropBlocks for mere regularization purposes.
Keeping $p=0$ during network evaluation leads to accurate \emph{mean} predictions, but nothing prevents a single dropout sample from being inaccurate or having large oscillations, resulting in a significant increase in the prediction uncertainty.
Except for DropBlock layers, no other form of regularization was used for all the networks discussed in this work.
Additional details on the implementation and hyperparameter selection for DropBlock layers are reported in the appendix.

\subsection{High- to low-fidelity representation coupling}\label{LMP:sec:mf_coupling}

\noindent For each of these three networks, we consider an \emph{implicit} and an \emph{explicit} coupling between the LF and the HF representations.
In the first implicit case, the LF predictors are not directly propagated towards the network output, as shown in Fig.~\ref{LMP:fig:low_multifidelity_network}, \ref{LMP:fig:multifidelity_network_2d} and \ref{LMP:fig:multifidelity_network_decoder}, where the black dotted arrows are omitted. However, forcing the upstream stages to learn accurate coarse pressure representations clearly affects the accuracy of the high-fidelity prediction.
Propagation of information through the dotted arrows is instead allowed for the explicit feedback mechanism, meaning that the LF predictions are propagated through the following stages of the decoder.
When the HF and LF truths belong to the same space and are correlated, an explicit connection helps in capturing the relationship between the LF and HF, such as in the two one-dimensional regression problems discussed in Section~\ref{sec:problem_ld}. 
However, it is unclear that an explicit connection would be beneficial for dense and low- to high-dimensional regression, since the LFs and HF live on different spaces. 
Since the LFs are of lower dimensionality than the HF, no bijective mapping exists between their respective spaces.
In such a case, as it will be discussed in Section~\ref{LMP:sec:res_hd}, our results seem to indicate the tendency of the network between two successive LFs to learn a \emph{discrepancy} between their corresponding feature maps, ultimately leading to improved HF predictions.

\section{Results}\label{LMP:sec:results}

\subsection{One-dimensional regression}\label{LMP:sec:res_ld}

\noindent We first trained the network only with LF data, to ensure there was no detriment to using a convolutional network as opposed to a fully-connected network (as in~\cite{meng2020composite}) for one dimensional regression.
After obtaining accurate predictions for the LF functions, we focused on multifidelity datasets.
The explicit network outperformed the implicit network, which seems reasonable given the significant correlation between the LF and HF models.

\begin{figure}[!ht]
\centering
\subfigure[]{\includegraphics[width=0.45\textwidth]{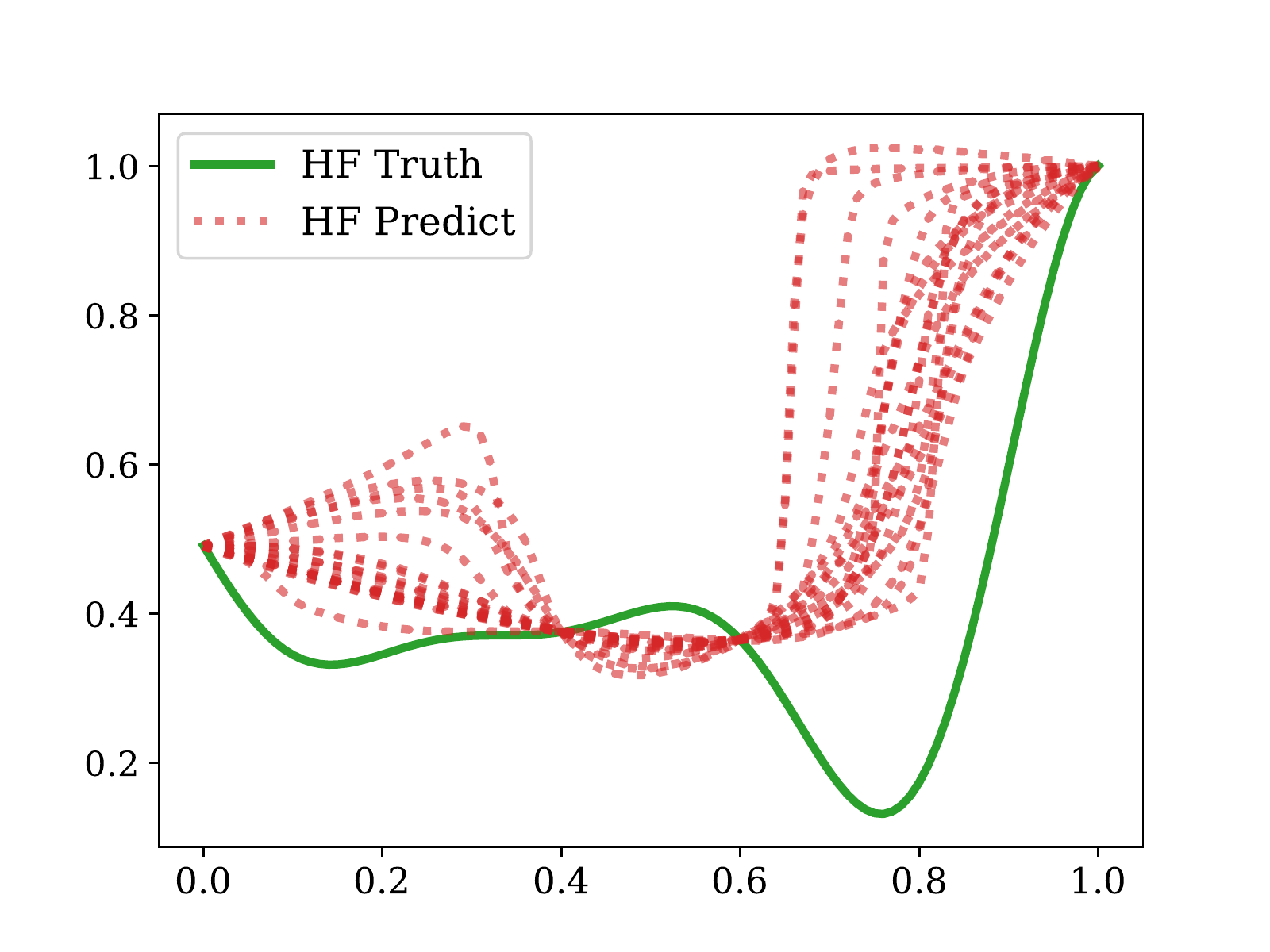} \label{LMP:fig:ex1_hf}}
\subfigure[]{\includegraphics[width=0.45\textwidth]{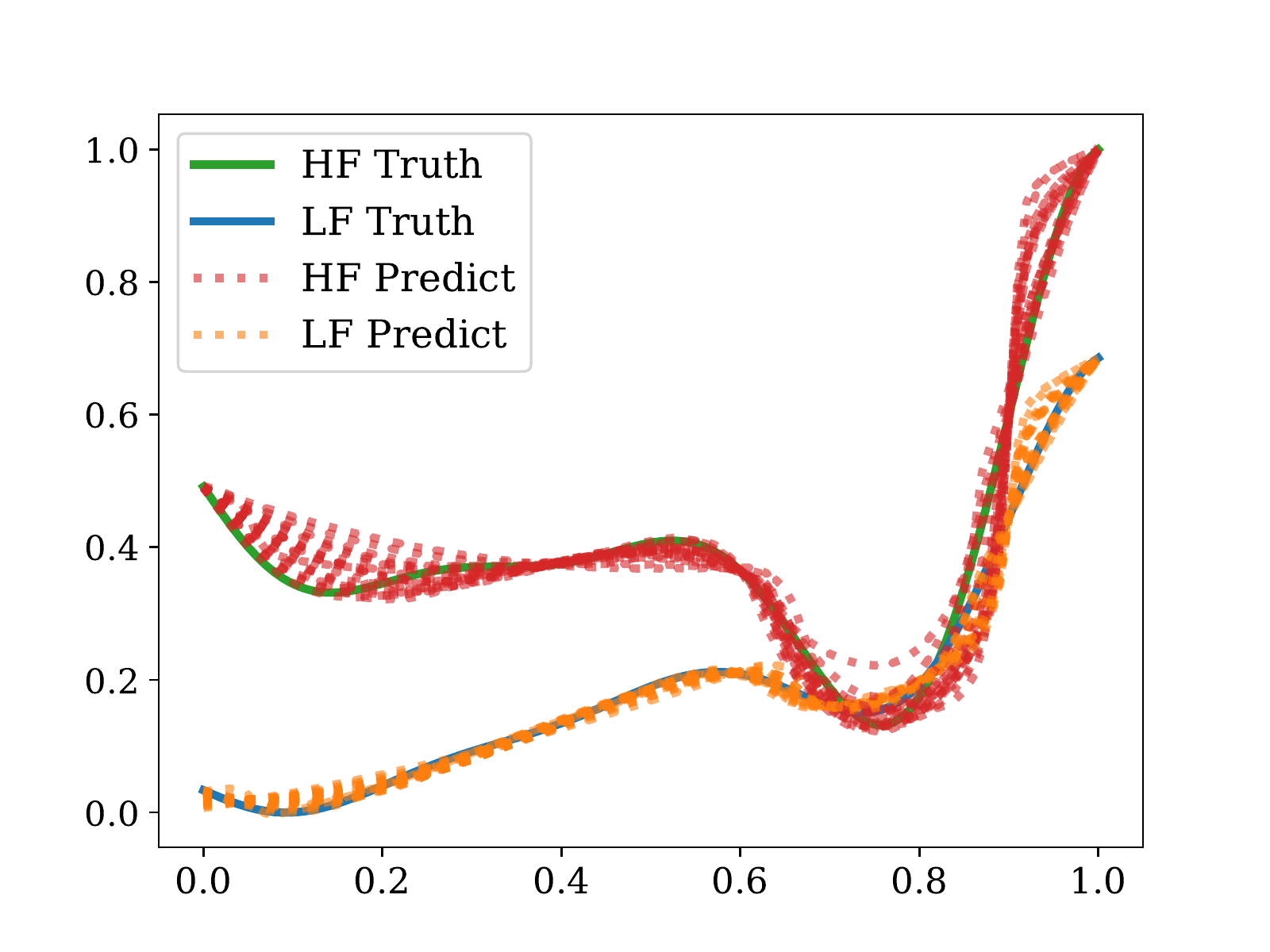} \label{LMP:fig:ex1_mf}}
\caption{Mean predictions from (a) HF only network, (b) MF network, for different initial weights. True function values are from Eqs.~\eqref{LMP:equ:example1_lf}-\eqref{LMP:equ:example1_hf}. }\label{LMP:fig:ex1_prediction}
\end{figure}

\begin{figure}[!ht]
\centering
\subfigure[]{\includegraphics[width=0.45\textwidth]{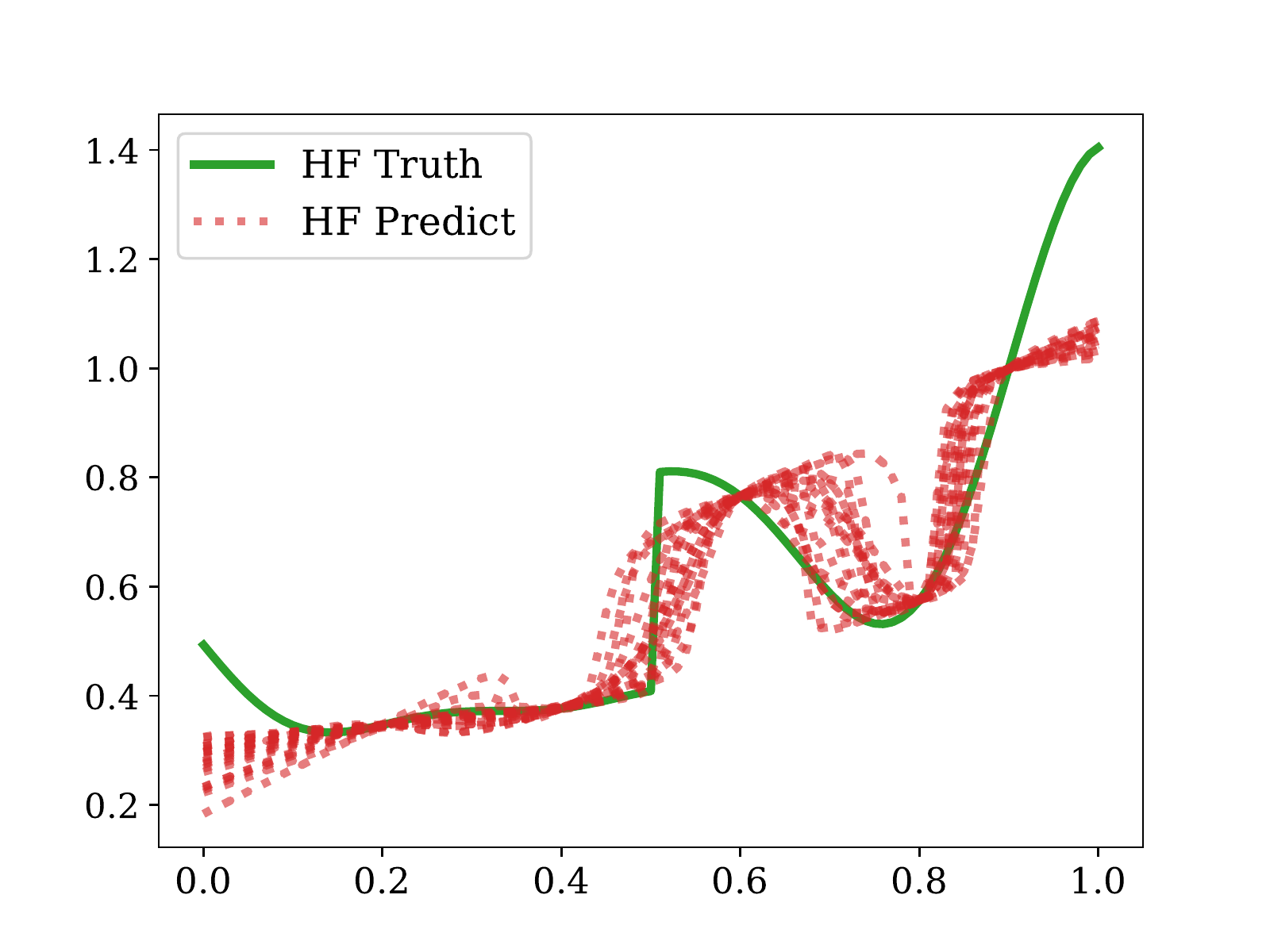} \label{LMP:fig:ex2_hf}}
\subfigure[]{\includegraphics[width=0.45\textwidth]{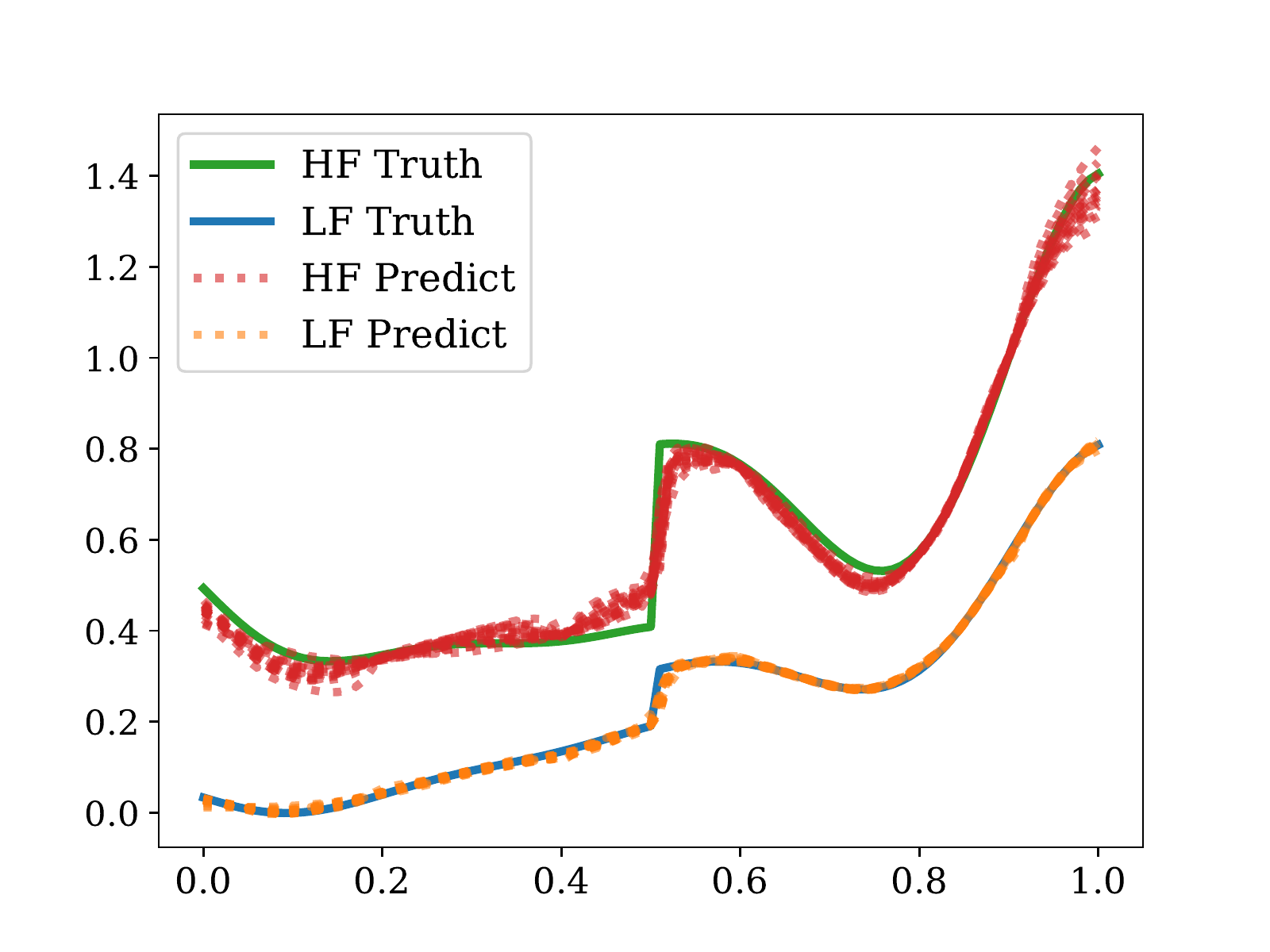} \label{LMP:fig:ex2_mf}}
\caption{Mean predictions from (a) HF only network (b) MF network trained with LF and HF data, for different initial weights. True function values are from Eqs.~\eqref{LMP:equ:example2_lf}-\eqref{LMP:equ:example2_hf}
}\label{LMP:fig:ex2_prediction}
\end{figure}

The multifidelity network is able to correctly leverage the LF data to influence the HF prediction to more closely resemble the true HF function, as shown in Fig.~\ref{LMP:fig:ex1_mf}, compared to the predictions from the network trained with HF data only in Fig.~\ref{LMP:fig:ex1_hf}.
Similarly, in Fig.~\ref{LMP:fig:ex2_prediction}, the multifidelity network is able to capture the discontinuity by extracting this information from the LF data, since this feature could not be learned from the limited HF data.
Equally accurate predictions in Fig.~\ref{LMP:fig:ex1_hf} and Fig.~\ref{LMP:fig:ex2_hf} result from different initial choices for the weights and biases (see appendix).

Including the spatial coordinate $x$ as an additional input downstream of the LF predictor was a necessary adjustment needed to separate the LF into a linear and a non linear contribution, facilitating their combination into an optimal HF predictor; in this regard, note that in Eq.~\eqref{LMP:equ:example2_hf}, $y_H(x) = \alpha y_L(x) + F_l(x)$, where $\alpha$ is a constant and $F_l(x) = -20x + 20$ is linear in $x$.

Although similar, the two problem sets for one-dimensional regression differ in one important aspect, i.e., the $x$ coordinate for the HF samples is shared across fidelities for Eq.~\eqref{LMP:equ:example1_lf} and \eqref{LMP:equ:example1_hf}, whereas these locations are different for Eqs.~\eqref{LMP:equ:example2_lf} and~\eqref{LMP:equ:example2_hf}.
In this latter case, and in the absence of sufficient regularization, spikes may appear in the multifidelity network predictions at the locations of the HF data, produced by the LF predictor without altering the loss at the LF training locations (see, e.g., Fig.~\ref{fig:1d_dropout_stats_ex2_omitdrop}).
Inclusion of multiple DropBlock layers provide sufficient regularization to prevent this behavior.
This does not happen when using the same $x$ values for the LF and HF datasets, since a spike reducing the HF loss would necessarily increase the LF loss.

Finally, although~\cite{meng2021multi} reported robust results, their network required the regularization penalty and the network size to be carefully selected to capture the true underlying HF-LF correlation. Since no validation set was included, choosing the regularization penalty and network size would require some degree of manual tuning. This operation might not be possible in a realistic application, since HF data may not be readily available. Therefore, the sensitivity of our network to the regularization penalty or other forms of regularization (e.g. DropBlock), for the example in Fig.~\ref{LMP:fig:ex2_dataset}, does not appear to be a limitation of this specific architecture.

In principle, the nonlinear convolutional sub-network in Fig.~\ref{LMP:fig:low_multifidelity_network} could also capture linear correlations, under sufficient regularization. Under limited data, a linear kernel enforces this regularization without tuning regularization penalties to ensure that the simplest relationship is captured between the LF and HF. 
Additionally, for datasets with both nonlinear and linear correlations, excessive regularization on the kernels between the LF and HF predictors would result in only capturing the linear correlation. For the given set of hyperparameters, removing the linear kernel produces significantly less accurate HF predictions outside of the training points; however, decreasing the number of kernels in the nonlinear correlation results in accurate HF predictions even in the absence of the linear kernel.

\subsection{Dense regression}\label{LMP:sec:res_hd}

\noindent We trained the multifidelity networks with implicit and explicit feedback using 32 HF and 116 LF pressure samples for each resolution (this network configuration is denoted as \emph{MF 32/116}), and compared its outputs with those from the HF 116/0 and HF 32/0 networks, respectively.

\begin{figure}[!ht]
\centering
\subfigure[]{\includegraphics[width=0.45\textwidth]{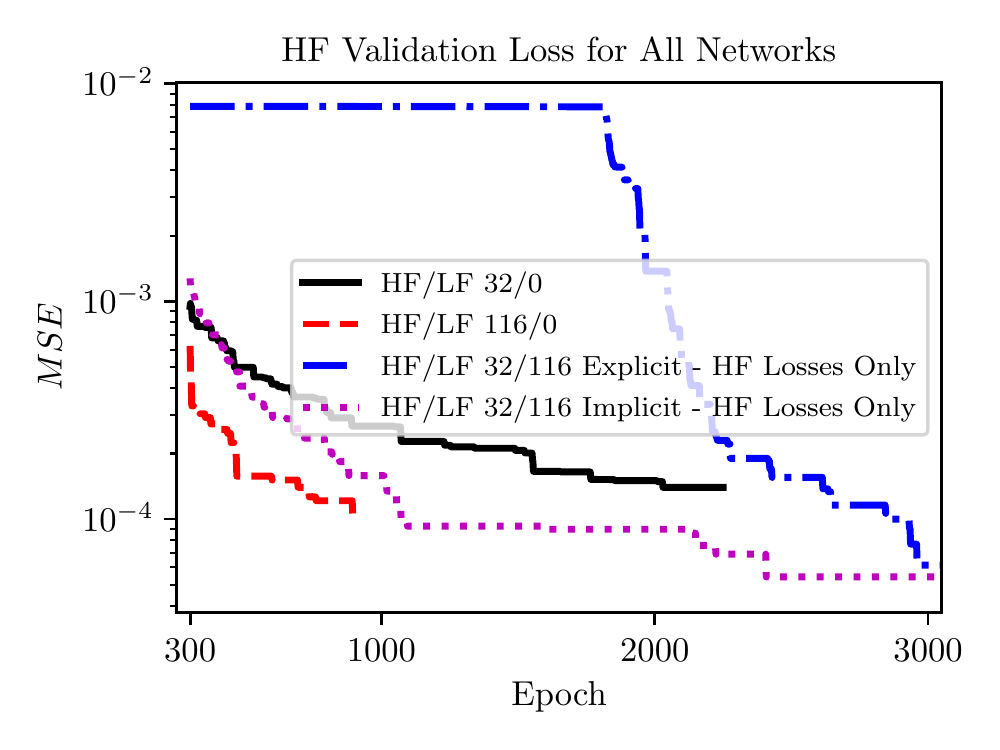} \label{LMP:fig:poiseuille_loss_HF}}
\subfigure[]{\includegraphics[width=0.45\textwidth]{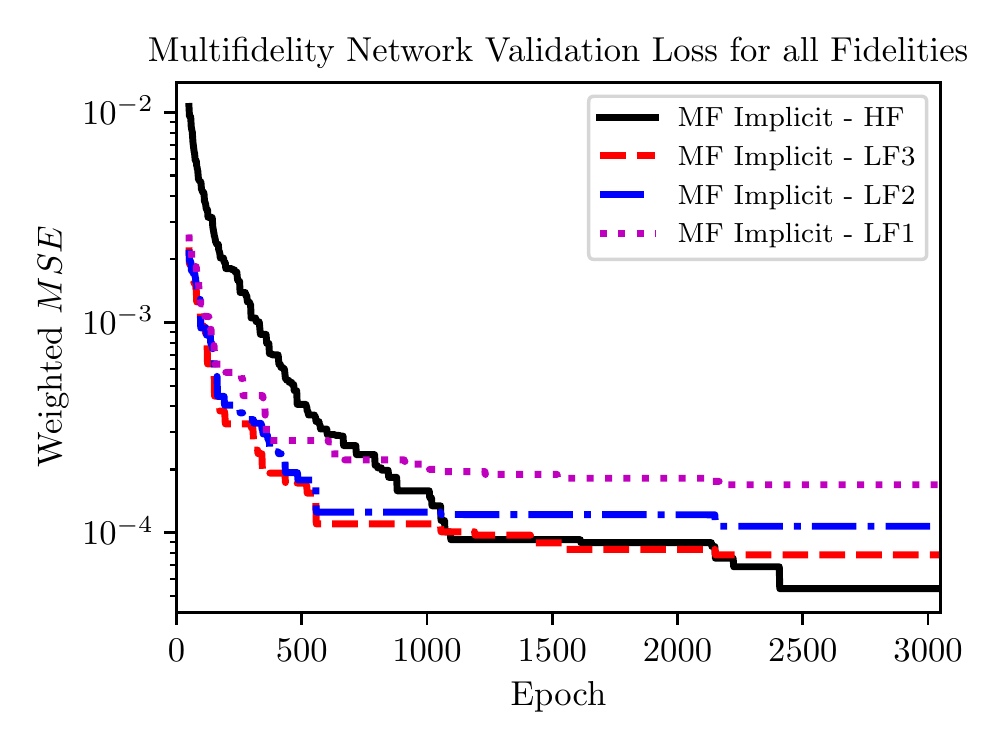} \label{LMP:fig:poiseuille_loss_MF}}
\caption{Validation loss profiles for dense pressure prediction in Poiseuille flow (a). The plot compares the losses resulting from HF only and MF training, using the best model for each category, i.e. with the highest test accuracy of those shown in Table~\ref{LMP:tab:accuracy_poiseuille}. The profiles for the weighted mean squared losses integrated over the fluid domain are shown in (b) for the best performing MF approach. All networks are trained for the same number of epochs, but only the decreasing losses are shown. Losses are plotted on a log scale.}\label{LMP:fig:poiseuille_loss}
\end{figure}

The multifidelity network with implicit feedback significantly improves the accuracy with respect to the HF 32/0 network, and its normalized accuracy is also superior to the HF 116/0 network.
In the case with multiple drop layers, the explicit multifidelity network accuracy suffers in the case of additive skip connections.
High-fidelity and multifidelity validation loss profiles are shown in Fig.~\ref{LMP:fig:poiseuille_loss}.
Fig.~\ref{LMP:fig:poiseuille_loss_HF} compares the HF contribution to the validation loss for four networks, i.e., HF 116/0, HF 32/0 and MF 32/116 with both implicit and explicit multifidelity coupling.
The plot emphasizes the acceleration in convergence produced by training networks with multifidelity data for the chosen hyperparameters in the implicit case; however, this may not be consistent across all sets of hyperparameters. Figure~\ref{LMP:fig:poiseuille_loss_MF}, on the other hand, demonstrates the ability of the network to learn multiple LF representations during training. 

\begin{figure}[!ht]
\centering
\subfigure[]{\includegraphics[width=\textwidth]{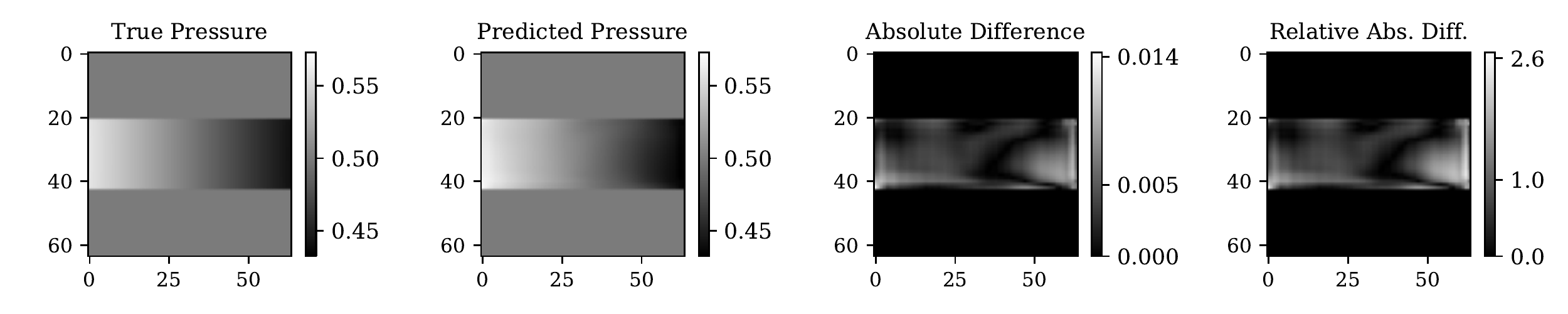}}
\subfigure[]{\includegraphics[width=\textwidth]{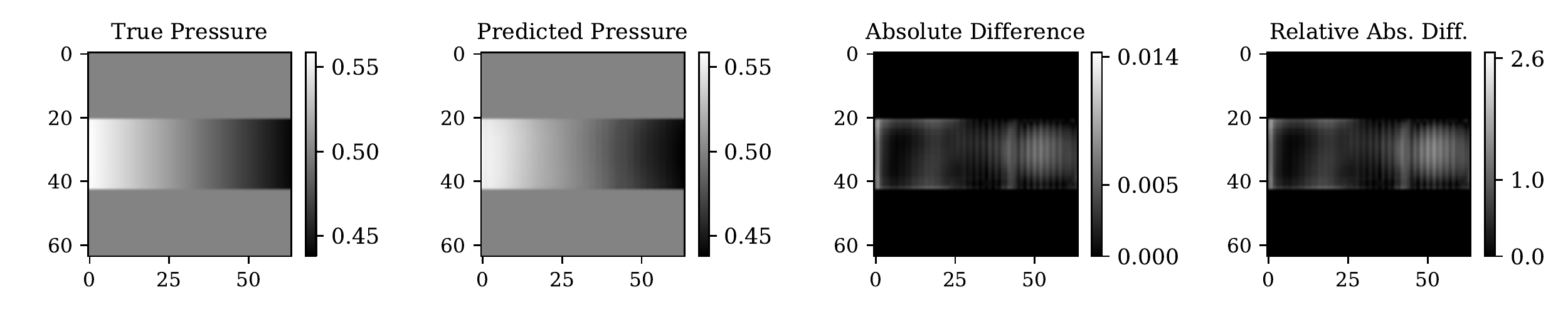}}
\subfigure[]{\includegraphics[width=\textwidth]{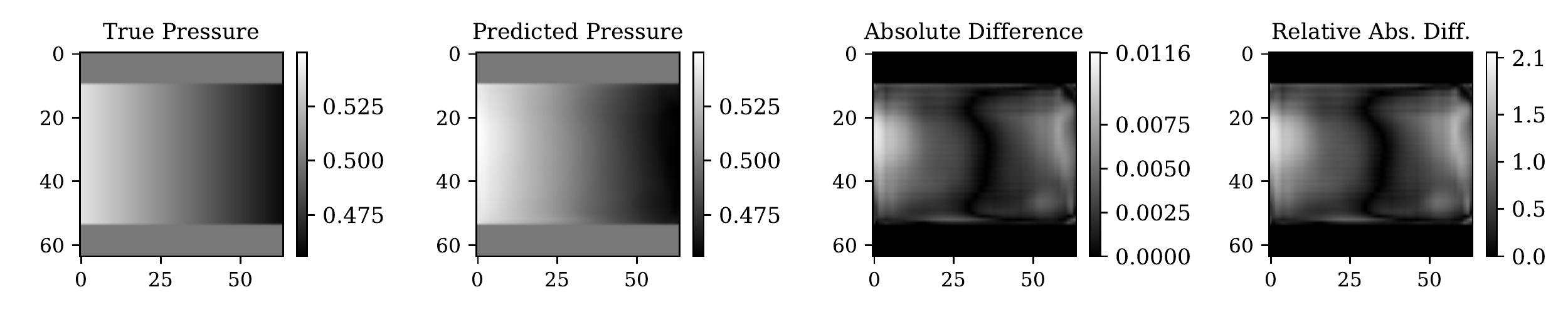}}
\subfigure[]{\includegraphics[width=\textwidth]{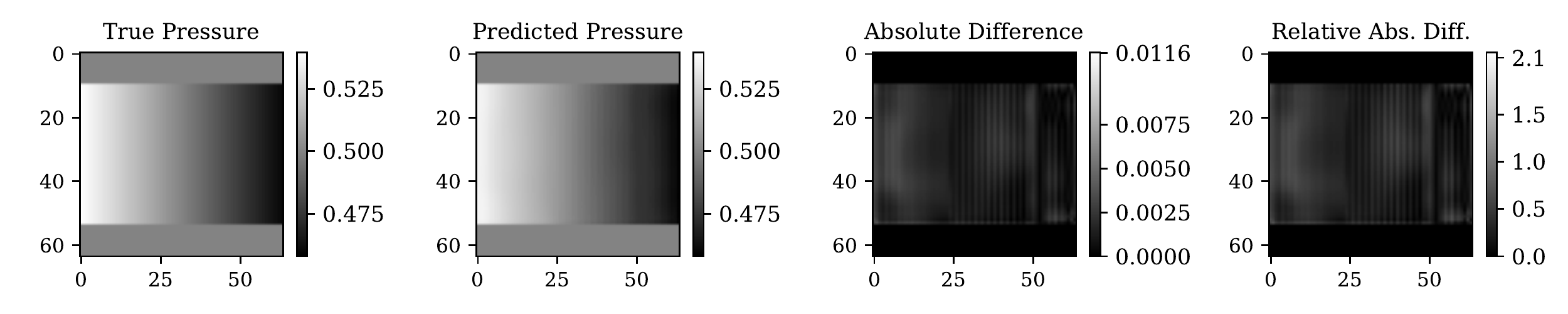}}
\caption{Dense regression network. Mean predictions from HF 32/0 on two pressure configurations from the test set (a,c). Mean predictions from MF 32/116 on the same test examples (b,d). Both network setups lead to accurate predictions with limited absolute and relative errors.}
\label{LMP:fig:result_pred}
\end{figure}

\begin{figure}[!ht]
\centering
\subfigure[Sample 1]{\includegraphics[width=0.32\textwidth]{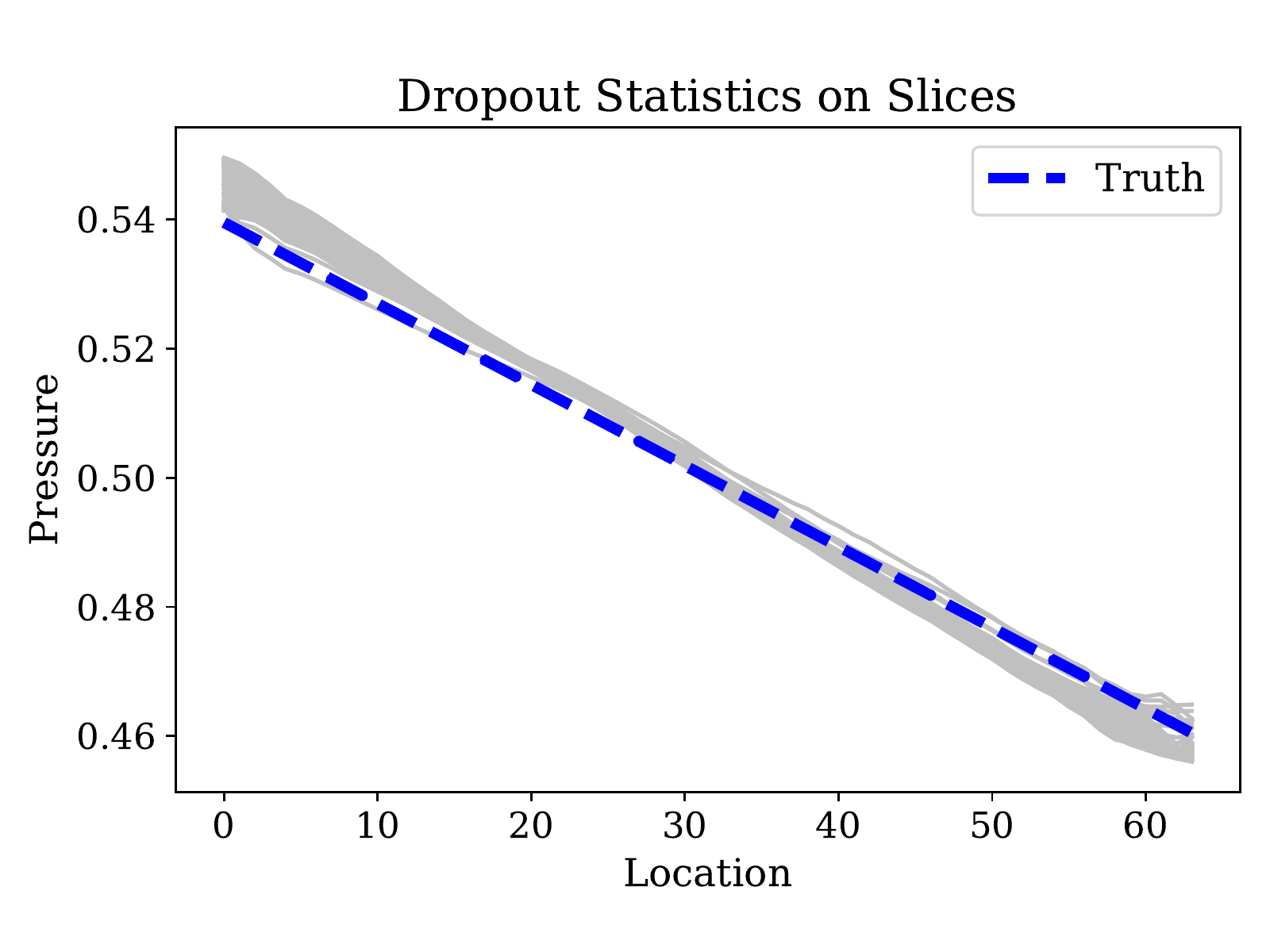}}
\subfigure[Sample 2]{\includegraphics[width=0.32\textwidth]{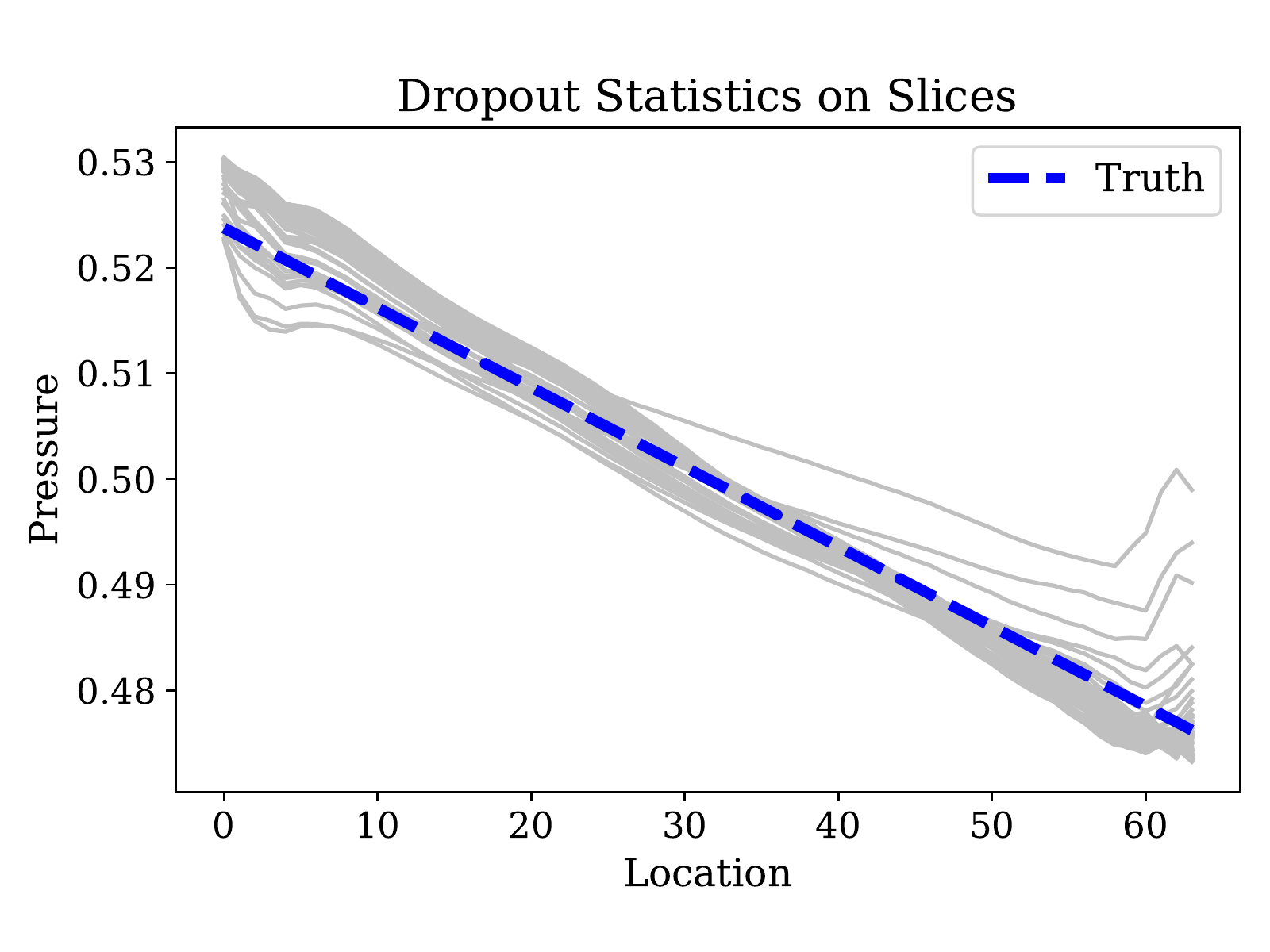}}
\subfigure[Sample 3]{\includegraphics[width=0.32\textwidth]{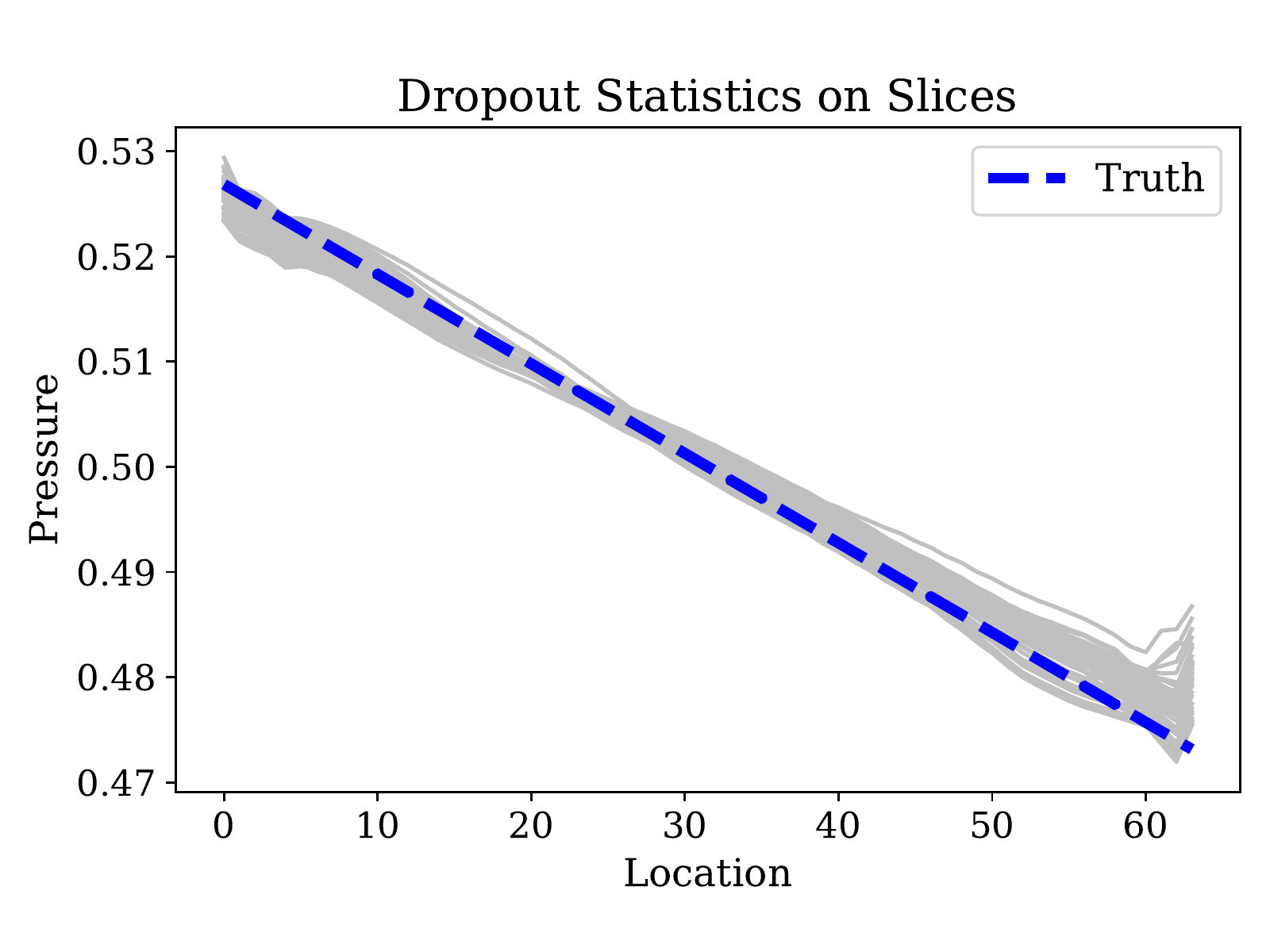}}
\subfigure[Sample 1]{\includegraphics[width=0.32\textwidth]{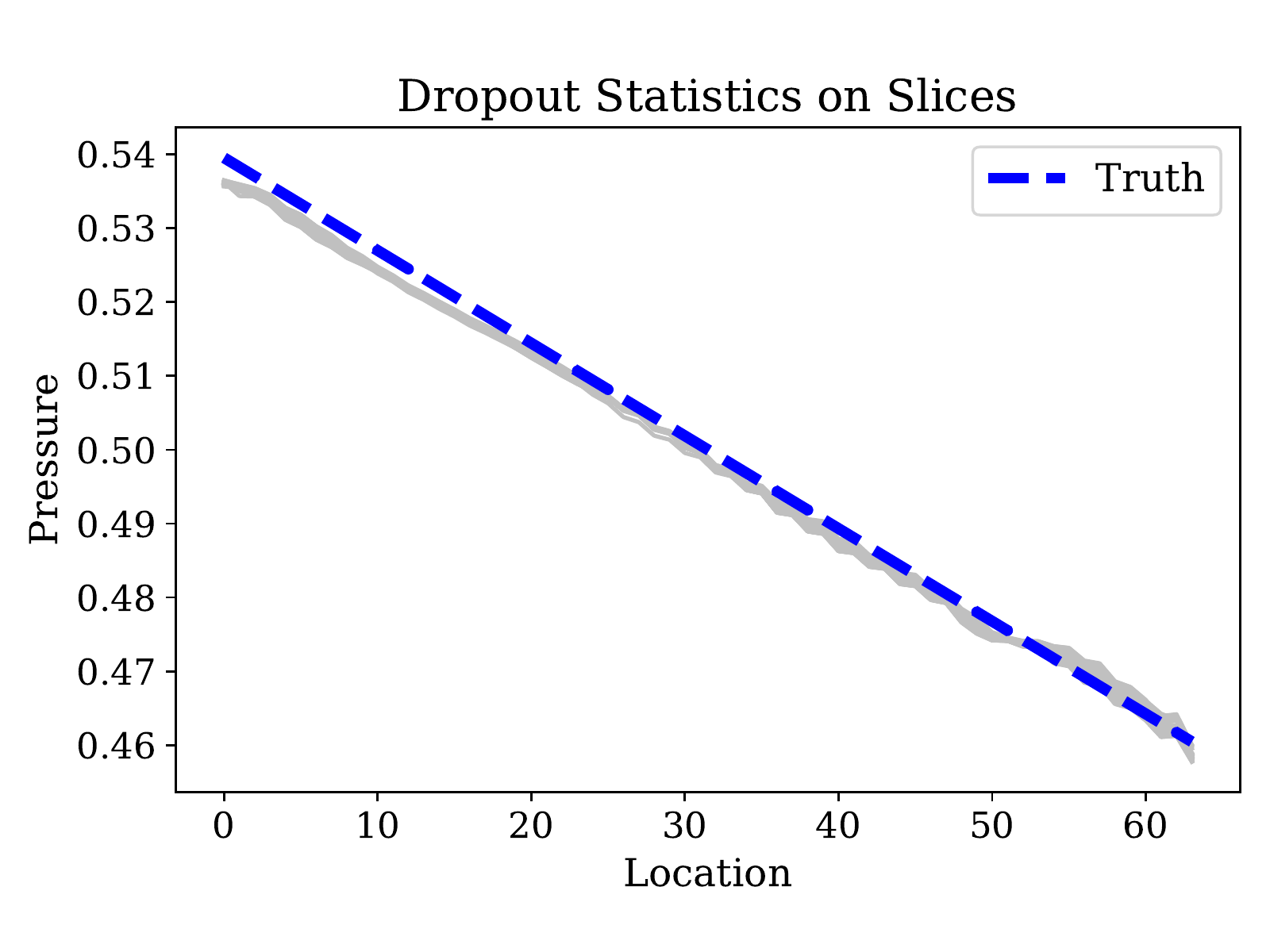}}
\subfigure[Sample 2]{\includegraphics[width=0.32\textwidth]{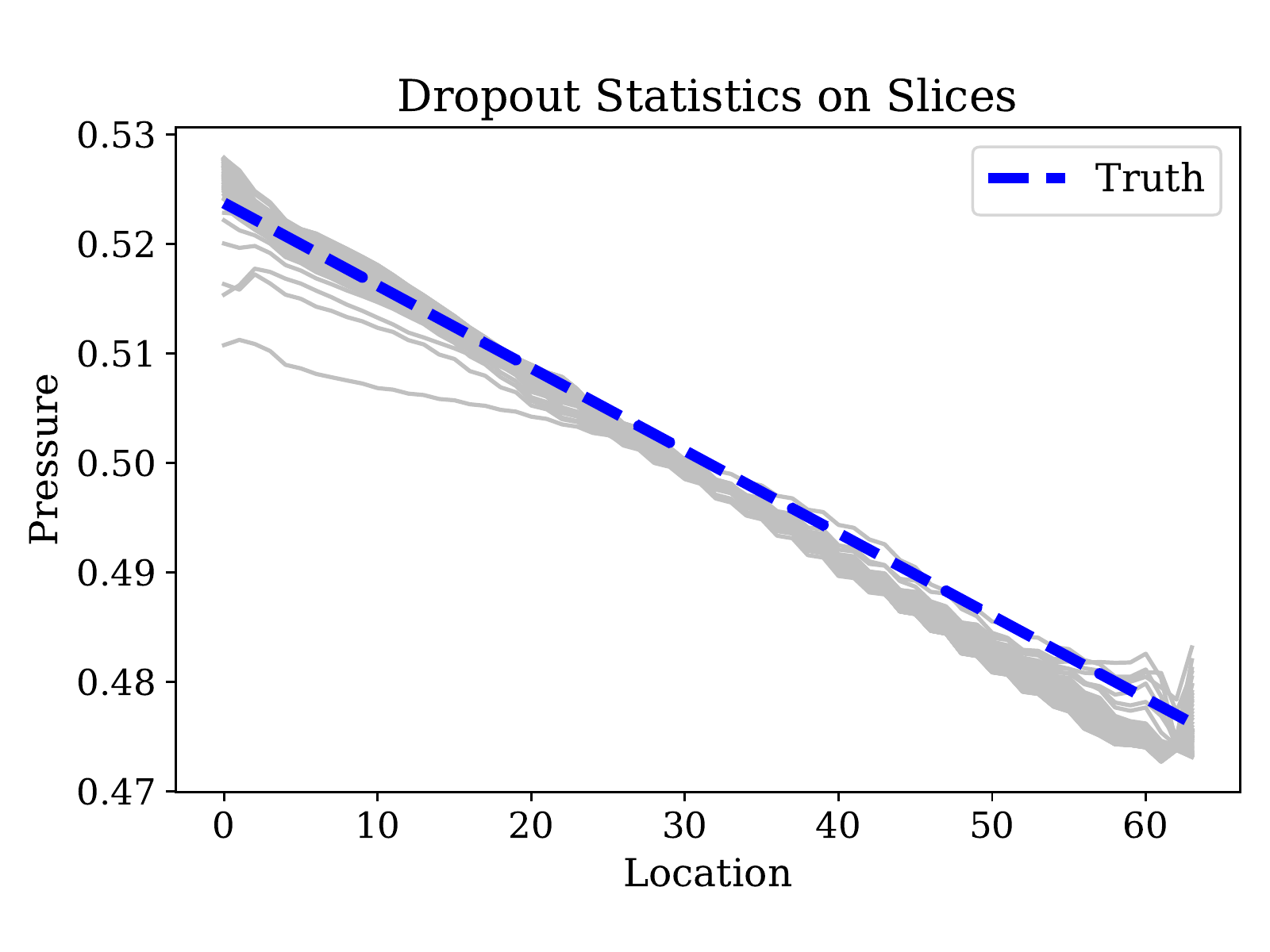}}
\subfigure[Sample 3]{\includegraphics[width=0.32\textwidth]{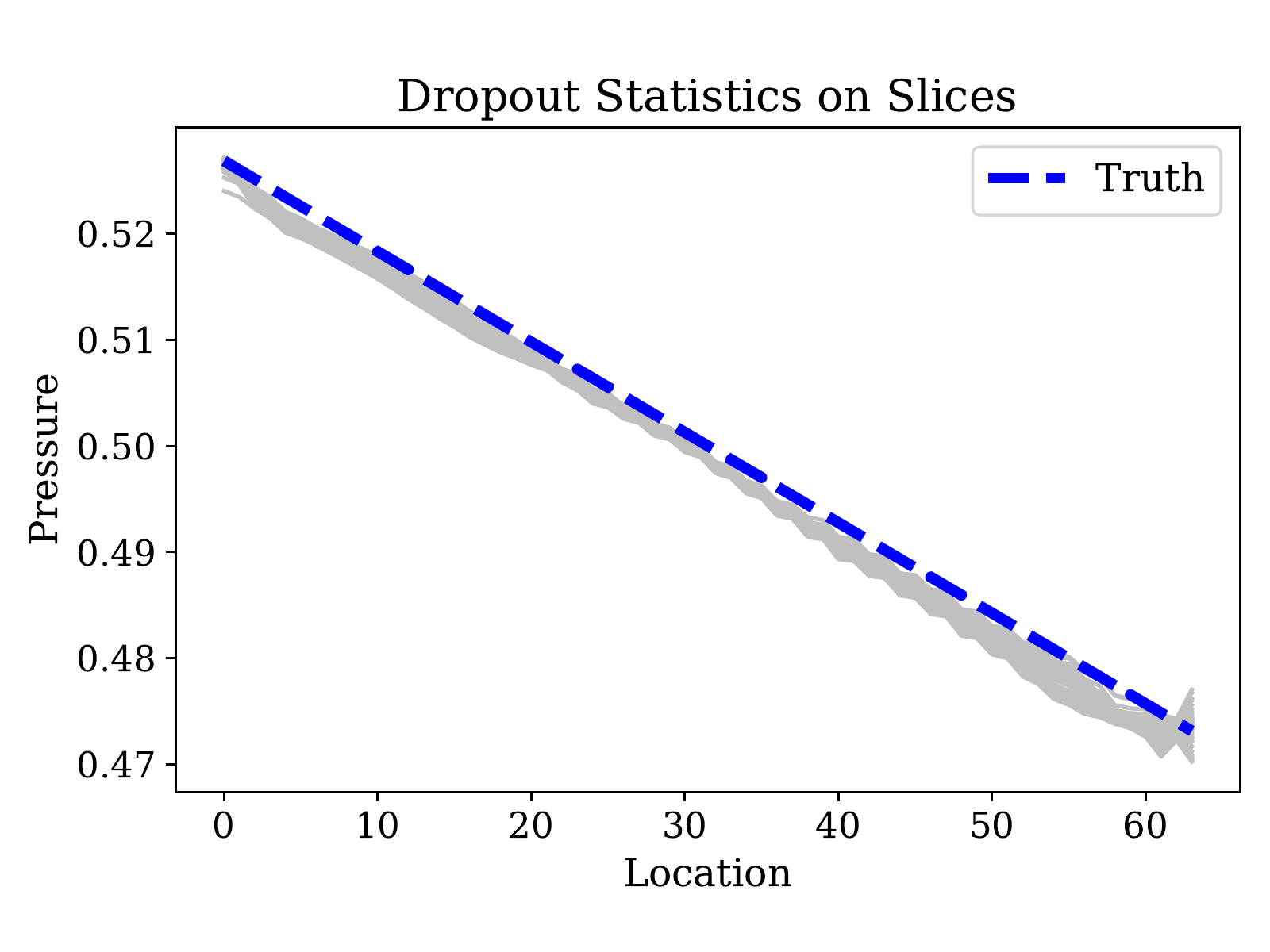}}
\caption{Dense regression network. Pressure predictions of test data on multiple slices parallel to the cylinder generator in the fluid region. Note that more slices are available for samples associated with a cylinder of larger radius $r$. (a)-(c) Mean predictions from HF 32/0 network with highest test accuracy. (d)-(f) Mean predictions from MF 32/116 network with highest test accuracy.}
\label{LMP:fig:poiseuille_predicted_slopes}
\end{figure}

The approximation accuracy of the HF 32/0 and MF 32/116 networks is compared for two pressure examples from the test set in Fig.~\ref{LMP:fig:result_pred} and for slices of the pressure field parallel to the cylinder generator in Fig.~\ref{LMP:fig:poiseuille_predicted_slopes}. 
The multifidelity network shows more consistently accurate predictions across the entire image instead of a localized region. 
The larger pressure errors noticeable near the fluid boundary appear typical of convolutional neural networks, which often report lower accuracy near the boundary (see, e.g.,~\cite{DBLP:journals/corr/abs-1805-03106}). This effect is magnified in our case as the boundary, which is associated with the (random) diameter, changes with every sample. 
The original U-Net architecture overcomes this through reflection padding on the input layer and no padding on any subsequent layers~\cite{ronneberger2015u} (whereas we zero pad each convolutional layer), although other approaches have been proposed in the convolutional network literature to overcome this limitation (see, e.g.,~\cite{DBLP:journals/corr/abs-1805-03106}).

\subsubsection{Effect of bias in low-fidelity predictor}

To explore to some extent the limitations of this network with regards to the accuracy of the low-fidelity data, we perform an additional test where the LF3 is biased. We choose the LF3 since we expect this data to have the most significant effect on the HF prediction due to its proximity in the network. Specifically, we add a constant bias $r\cdot[\max_{j,k} \text{LF}_3(j,k) - \min_{j,k} \text{LF}_3(j,k)])$ where the bias ratio is $r=0.01, 0.05$, to all of the LF3 data.
We first use the hyperparameters chosen in Table~\ref{tab:hyperparameters_encdec}. As shown in Table~\ref{tab:poiseuille_accuracy_bias}, most of the accuracy values are similar to those found in the absence of biased low-fidelity data (i.e. Table~\ref{LMP:tab:accuracy_poiseuille}), except a single outlier for the MF additive explicit feedback. However, after the selection of more appropriate hyperparameters, the accuracy for this test case is again similar to the case without bias (see updated accuracy in parenthesis). 
This seems to suggest the ability of the network to counteract excessive bias or inaccuracy in the LF predictors, but further analysis is required.
\begin{table}[ht!]
\centering
\begin{tabular}{ccccccc}
\hline
{\bf Skip Conn.} & {\bf Network Type} & {\bf MF Feedback} & {\bf HF/LF} & {\bf $R^{2}$} & {\bf Normalized $R^{2}$}\\
\toprule
Concat & MF & Explicit & 32/116 & 0.9326  & 3.250e-06 \\
Add & MF & Explicit & 32/116 & 0.9372  & 3.266e-06  \\
\midrule
Concat & MF & Implicit & 32/116 & {\bf 0.9672} & 3.370e-06 \\
Add & MF & Implicit & 32/116 & 0.9541  & 3.325e-06 \\
\midrule
Concat & HF & - & 32/0 & 0.9284 & {\bf 7.083e-06} \\
Add & HF & - & 32/0 & 0.9047 & 6.902e-06 \\
\midrule
Concat & HF & - & 116/0 & 0.9408 & 1.980e-06 \\
Add  & HF & - & 116/0 & 0.9257 & 1.948e-06 \\
\bottomrule
\end{tabular}
\caption{Comparison of HF and MF network performance for dense regression. The normalized accuracy is $R^2/C$ where $C$ is the cost (see Section~\ref{sec:accuracy}). The terms \emph{Concat} and \emph{Add} refer to how the information from a skip connection is assembled into the decoder.}
\label{LMP:tab:accuracy_poiseuille}
\end{table}

\begin{table}[]
\begin{tabular}{cccccc}
\hline
{\bf Bias ratio} & {\bf Skip Conn.} & {\bf Network Type} & {\bf MF Feedback} & {\bf HF/LF} & {\bf $R^{2}$} \\
\toprule
0.01 & Concat & MF & Explicit & 32/116 &  0.944853 \\
0.05 & Concat & MF & Explicit & 32/116 &  0.933174 \\

0.01 & Add & MF & Explicit & 32/116 & 0.198742 (0.935613*) \\
0.05 & Add & MF & Explicit & 32/116 & 0.92367 \\
\midrule

0.01 & Concat & MF & Implicit & 32/116 & 0.95079 \\
0.05 & Concat & MF & Implicit & 32/116 & {\bf 0.960783} \\

0.01 & Add & MF & Implicit & 32/116 & 0.942729 \\
0.05 & Add & MF & Implicit & 32/116 & 0.949461 \\
\bottomrule
\end{tabular}
    \caption{Accuracy of multifidelity networks in Table~\ref{LMP:tab:accuracy_poiseuille}, when trained with biased LF3. (*) Validation accuracy obtained by re-optimizing the network hyperparameters.}\label{tab:poiseuille_accuracy_bias}
\end{table}

\subsection{Low- to high-dimensional dense regression}
\label{LMP:sec:res_ld_hd}

\begin{figure}[!ht]
\centering
\subfigure[]{\includegraphics[width=0.9\textwidth]{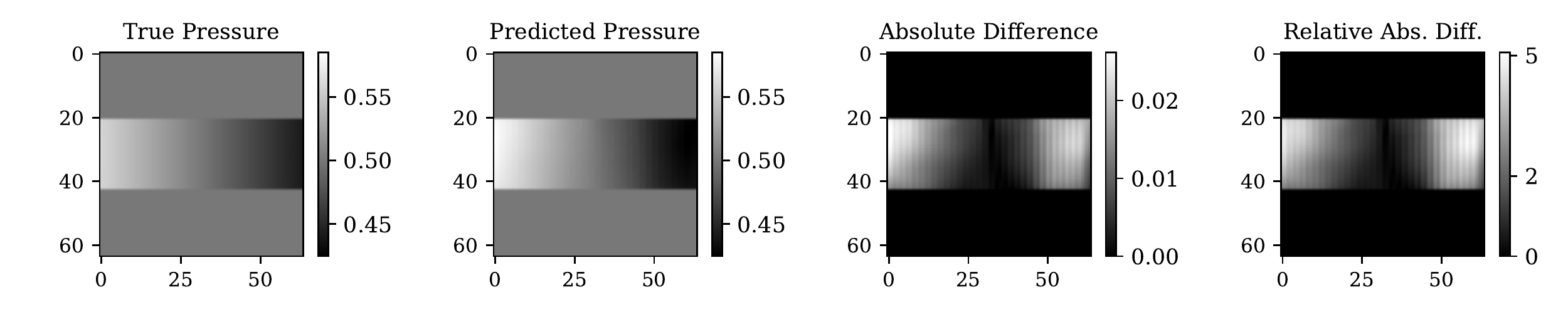} }
\subfigure[]{\includegraphics[width=0.9\textwidth]{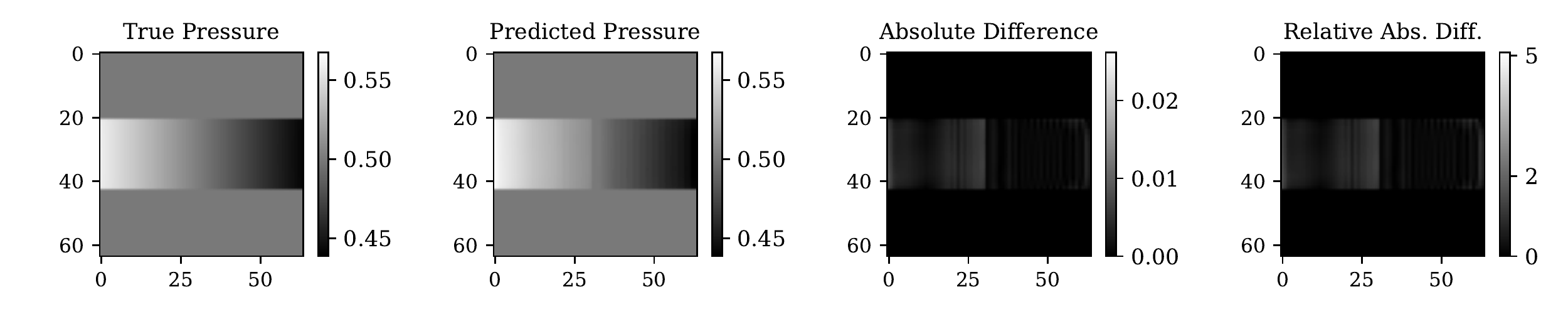} }
\subfigure[]{\includegraphics[width=0.9\textwidth]{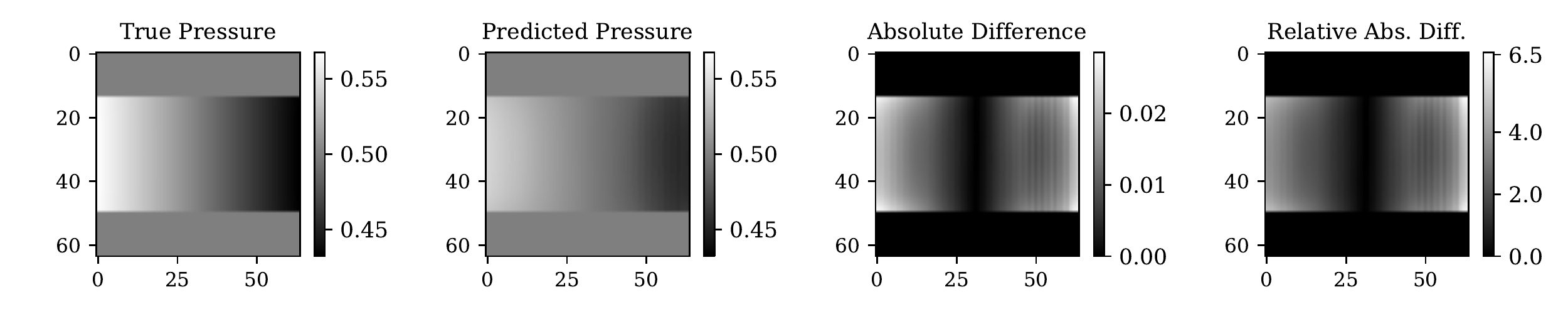} \label{fig:predictions_poiseuille_params_largeradius1} }
\subfigure[]{\includegraphics[width=0.9\textwidth]{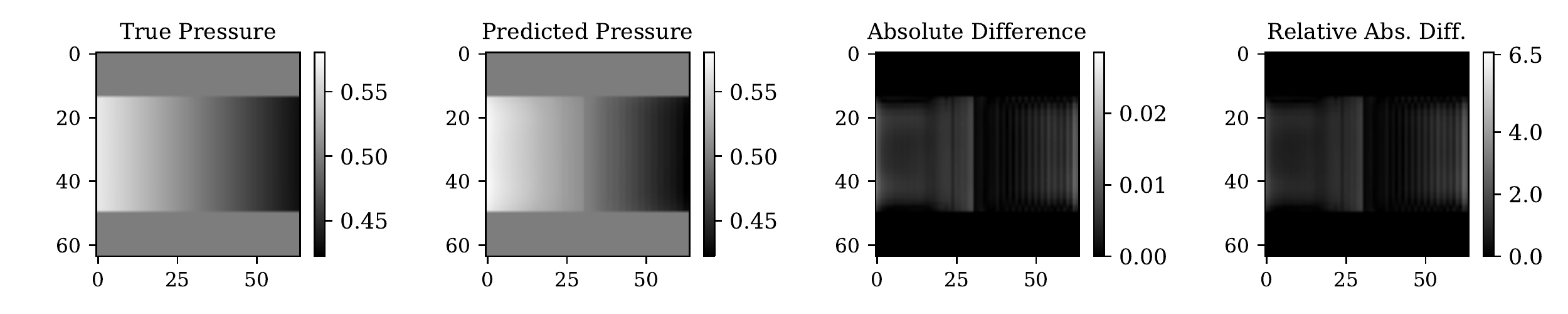} \label{fig:predictions_poiseuille_params_largeradius2}}
\caption{Predictions from the decoder network for the low- to high-dimensional regression test case. (a,c)  Mean prediction from HF 32/0 network. (b,d) Mean prediction from MF 32/116 network.}
\label{fig:predictions_poiseuille_params}
\end{figure}

\noindent The results from the low- to high-dimensional decoder architecture reported in Table~\ref{tab:accuracy_poiseuille_params} show competitive test set accuracy for the MF 32/116 network with respect to the HF 116/0 network. 
In addition, both MF networks perform better than the HF 32/0 network, as shown in Fig.~\ref{fig:predictions_poiseuille_params}.

\begin{table}[ht!]
\centering
\begin{tabular}{cccccc}
\toprule
{\bf Network Type} & {\bf MF Feedback} & {\bf HF/LF} & {\bf $R^{2}$} & {\bf Normalized $R^{2}$}\\
\midrule
MF & Explicit & 32/116  & {\bf 0.9330}  & 3.251e-06 \\
MF & Implicit & 32/116  & 0.9054 & 3.155e-06 \\
\midrule
HF & - & 32/0  & 0.8369 & {\bf 6.385e-06} \\
HF & - & 116/0  & 0.8907  &  1.875e-06 \\
\bottomrule
\end{tabular}
\caption{Low- to high-dimensional regression comparison of HF and MF network performance. Accuracy is computed similar to Table~\ref{LMP:tab:accuracy_poiseuille}. }
\label{tab:accuracy_poiseuille_params}
\end{table}

\subsection{Uncertainty quantification of network predictions}\label{sec:res_uq}

\noindent In this section, we analyze the uncertainty estimates from MC DropBlock, or, in other words, we quantify the variability from an ensemble of predictions obtained by feeding the same input to the network $N_{\text{UQ}}= 1,000$ times.
 
\subsubsection{One-dimensional regression}
\begin{figure}[!ht]
\centering
\subfigure[]{\includegraphics[width=0.45\textwidth]{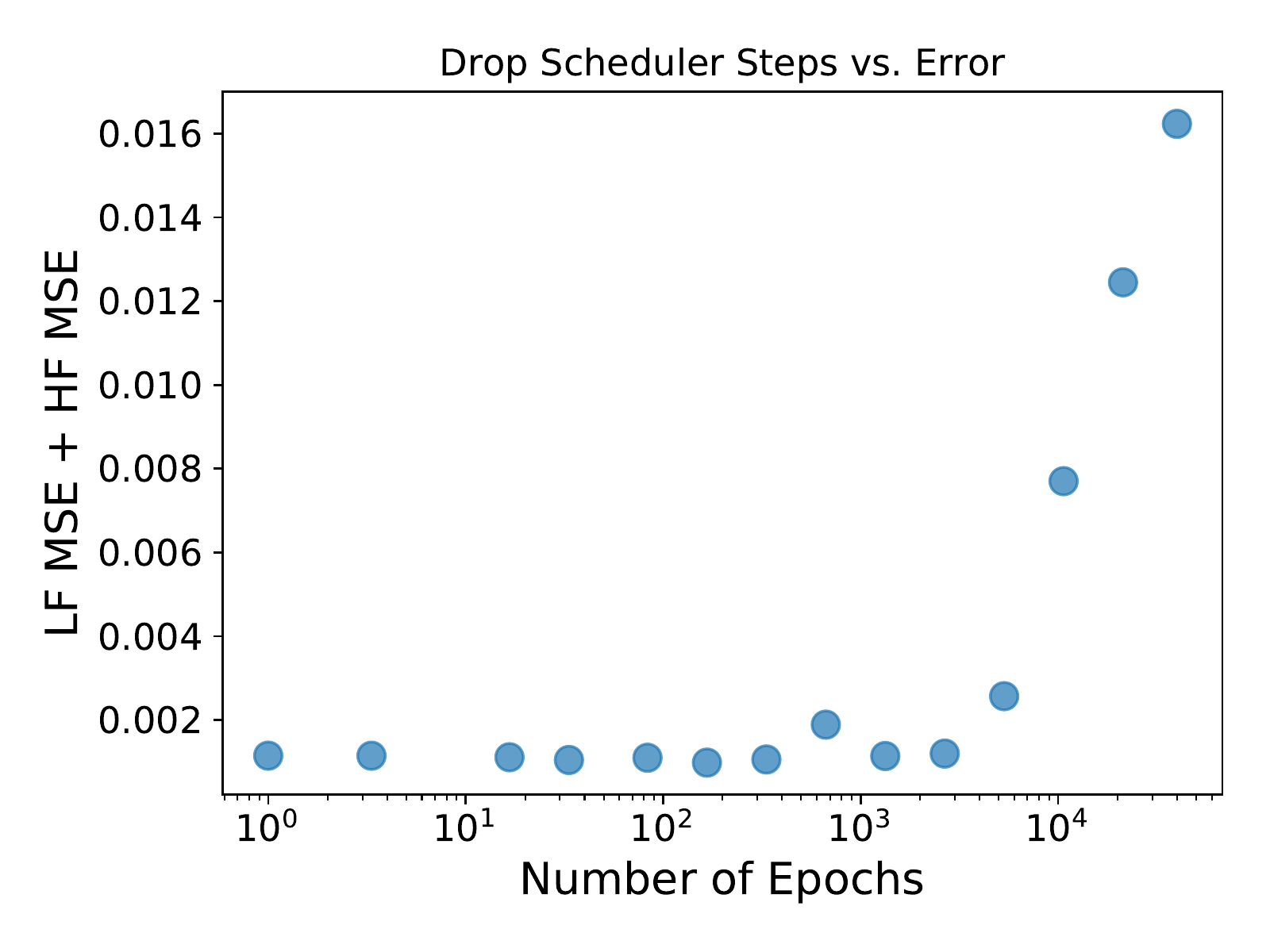} \label{fig:ex1_dropscheduler}}
\subfigure[]{\includegraphics[width=0.45\textwidth]{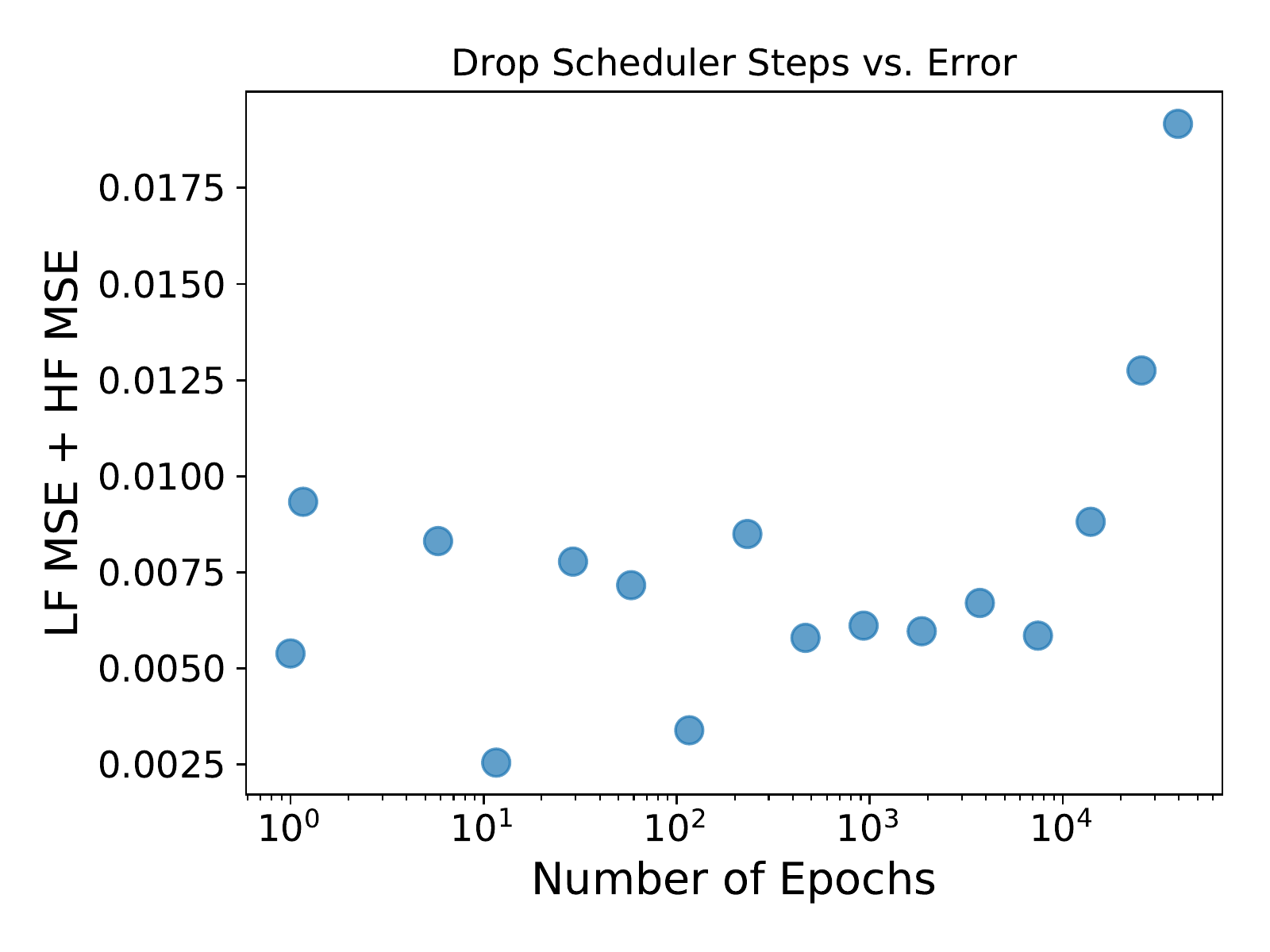} \label{fig:ex2_dropscheduler}}
\caption{One-dimensional regression errors produced by varying the drop probability scheduler ramp. Mean squared error (MSE) is computed over all dropout predictions for the LF training points and HF training points separately, and summed.
The x axis refers to the number of epochs required to reach $p=0.1$ using a linear drop probability scheduler.
(a) Eq.~\eqref{LMP:equ:example1_lf}-\eqref{LMP:equ:example1_hf} with dropout layer after every convolution layer, \textit{except} the first two convolution layers, preceding the LF prediction and (b) Eq.~\eqref{LMP:equ:example2_lf}-\eqref{LMP:equ:example2_hf} with dropout layer after every convolution layer, \textit{except} the first four convolution layers, preceding the LF prediction. A tanh activation is used in all these tests.}
\label{fig:drop_scheduler_comparison}
\end{figure}

\begin{figure}[!ht]
\centering
\subfigure[]{\includegraphics[width=0.45\textwidth]{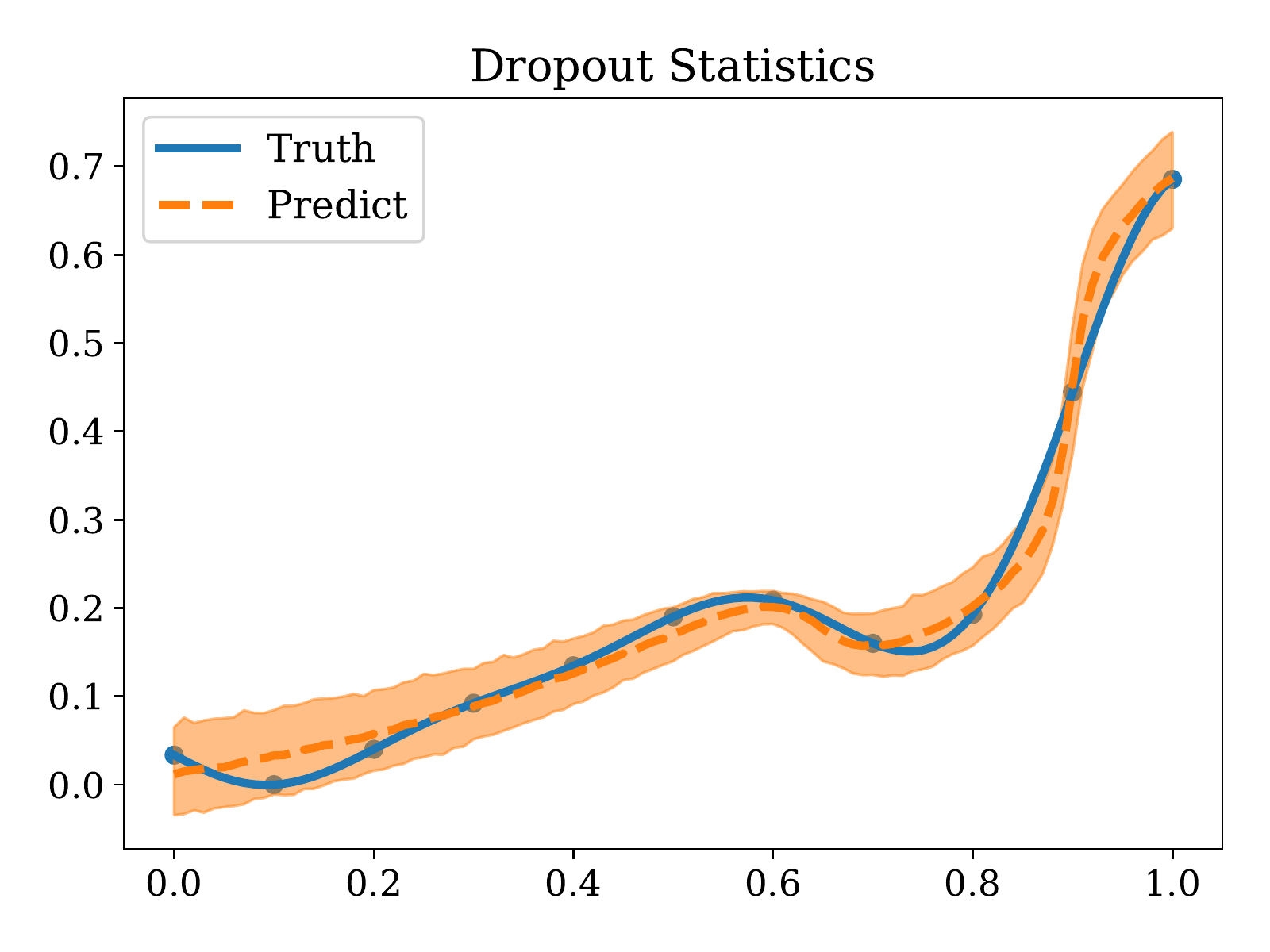} \label{fig:ex1_dropout_lf}}
\subfigure[]{\includegraphics[width=0.45\textwidth]{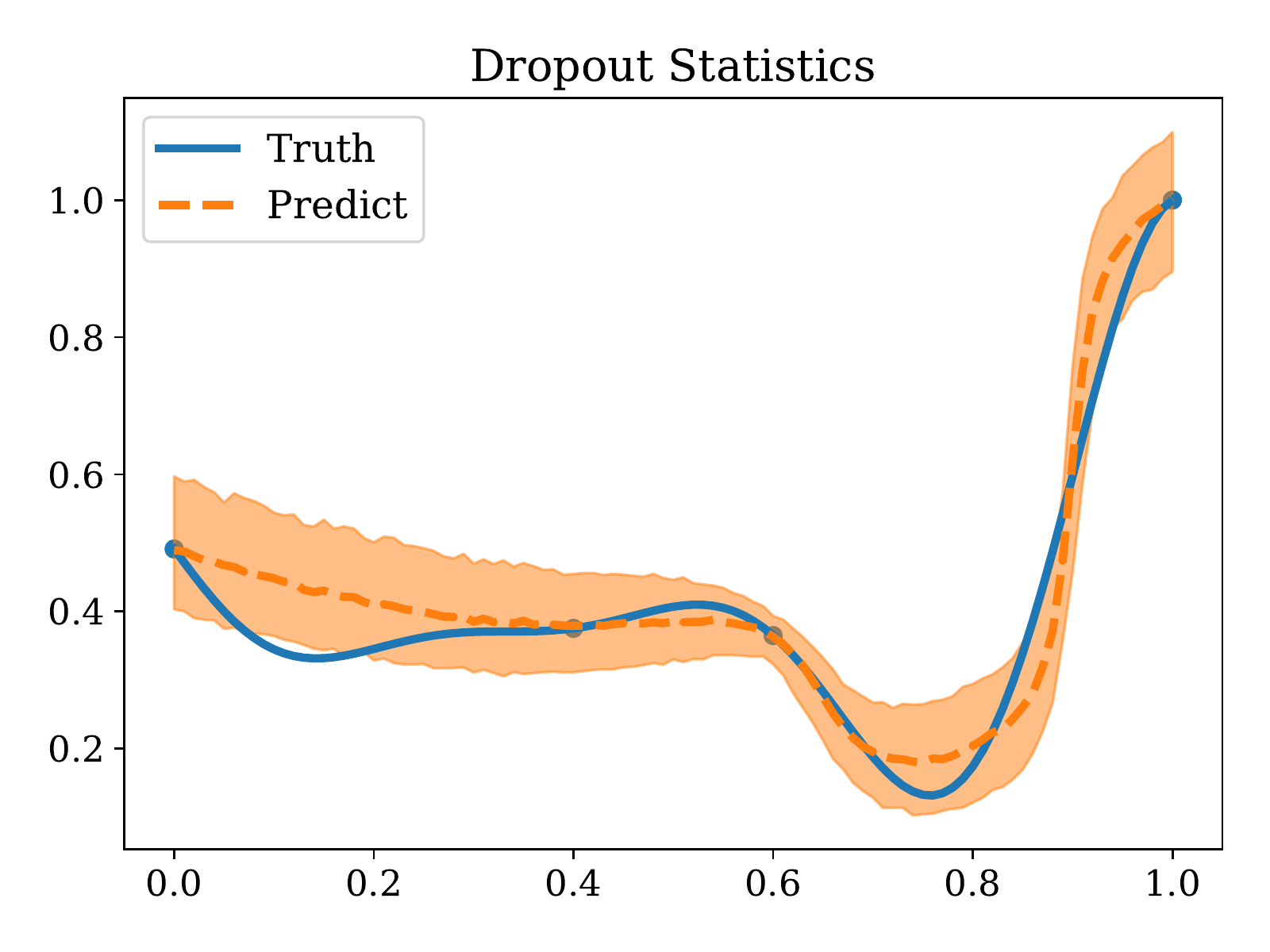} \label{fig:ex1_dropout_hf}}
\caption{Dropout layer after every convolution layer preceding the LF prediction. Mean prediction and estimated 5\%-95\% percentiles for (a) Eq.~\eqref{LMP:equ:example1_lf} and (b) Eq.~\eqref{LMP:equ:example1_hf}, from a network with multiple DropBlock realizations. In these tests we used a tanh activation function.}\label{fig:1d_dropout_stats_ex1}
\end{figure}

\begin{figure}[!ht]
\centering
\subfigure[]{\includegraphics[width=0.45\textwidth]{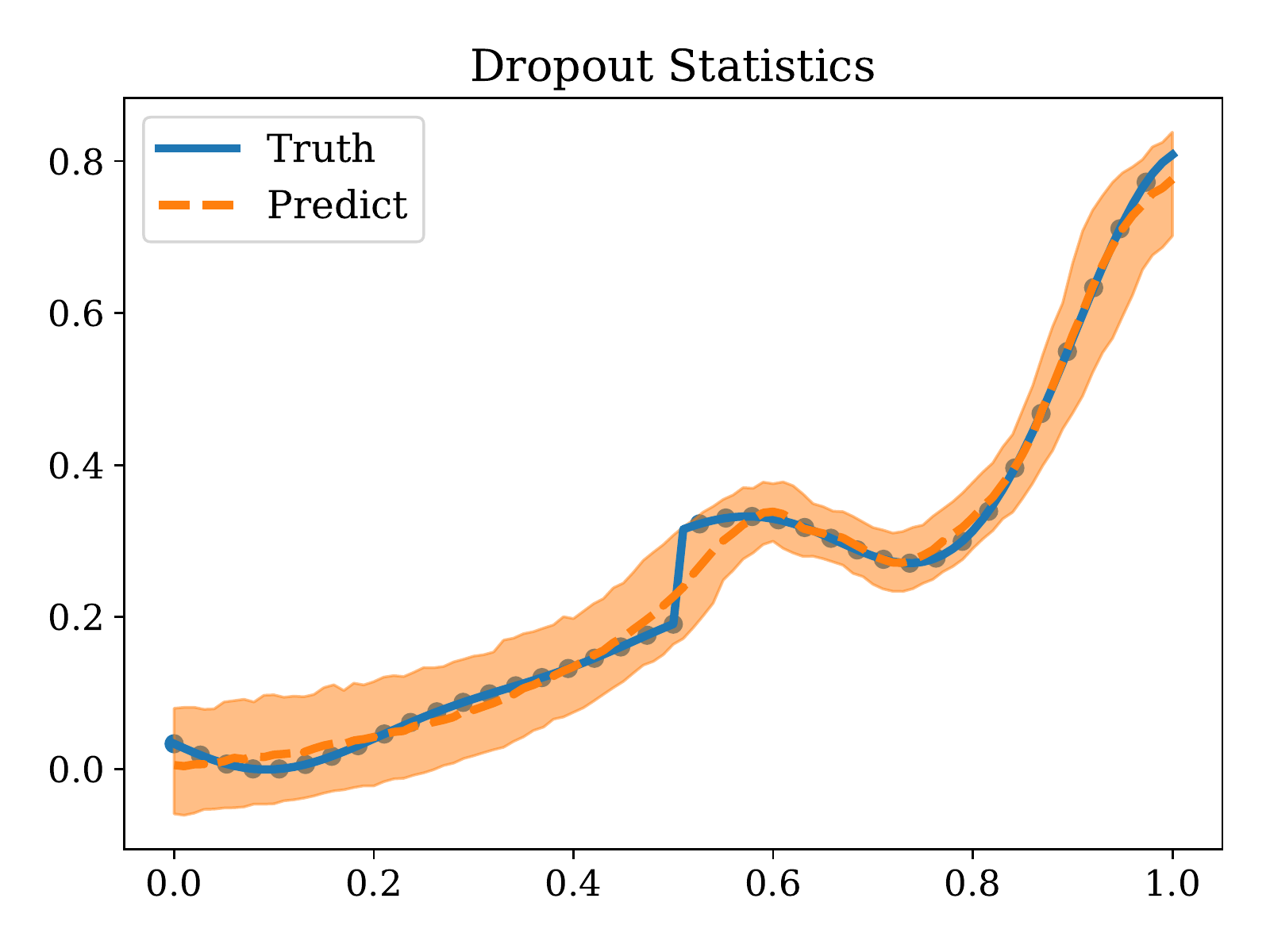} \label{fig:ex2_dropout_lf}}
\subfigure[]{\includegraphics[width=0.45\textwidth]{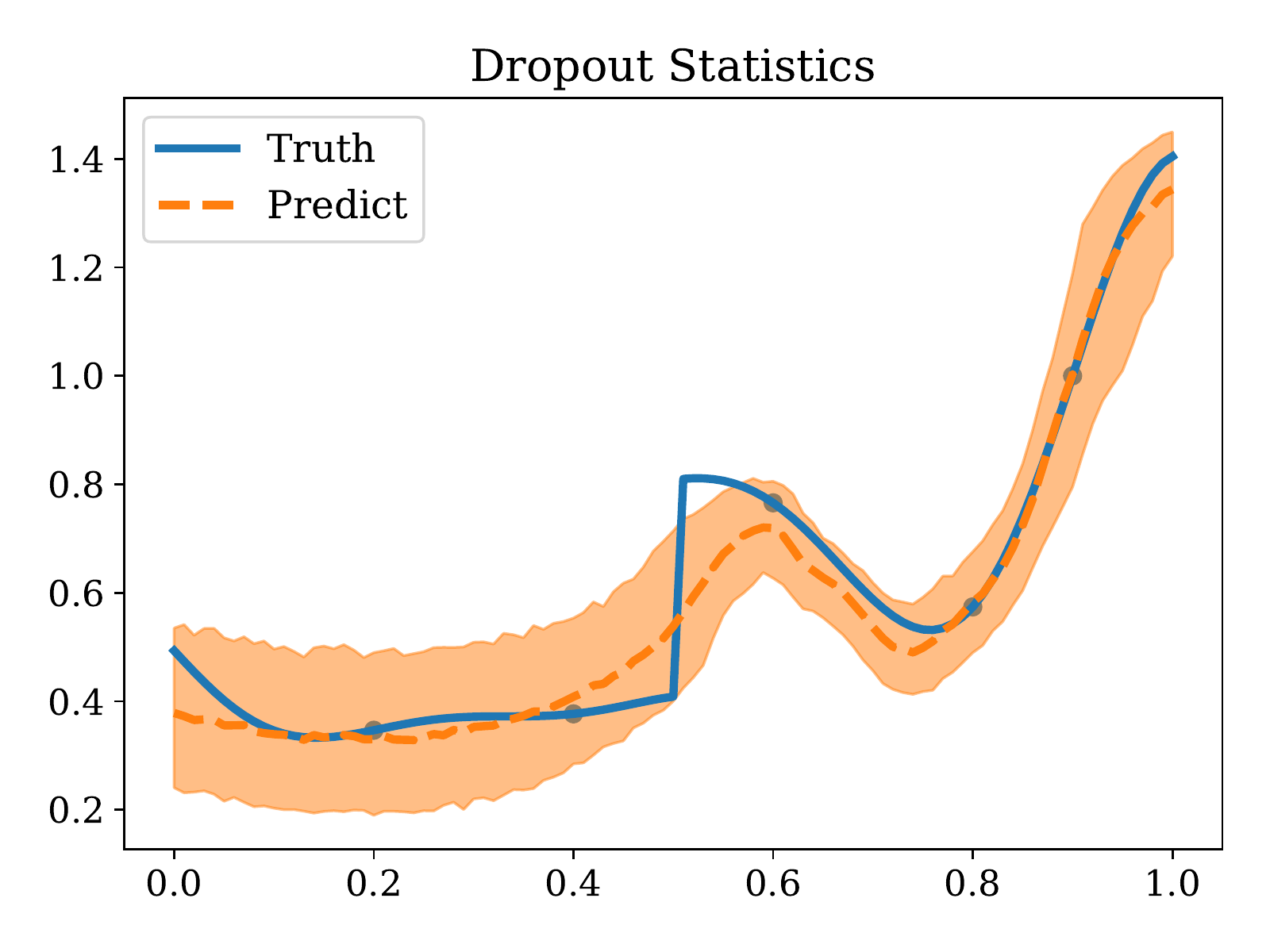} \label{fig:ex2_dropout_hf}}
\caption{Dropout layer after every convolution layer preceding the LF prediction. Mean prediction and estimated 5\%-95\% percentiles for (a) Eq.~\eqref{LMP:equ:example2_lf} and (b) Eq.~\eqref{LMP:equ:example2_hf}, from a network with multiple DropBlock realizations. In these tests we used a tanh activation function.}\label{fig:1d_dropout_stats_ex2}
\end{figure}

\noindent  We experimented with different drop probability schedulers which have been shown to produce improved testing accuracy, including an exponential decay~\cite{morerio8237645cvdropout} and a delayed start to a linear scheduler. No scheduler was found to generate accurate mean predictions and, at the same time, uncertainty intervals which bounded both the true LF and HF functions.
Exponential schedulers were found to produce less accurate dropout realizations and wider uncertainty intervals. 
%
Additionally, no significant difference was observed by changing the speed at which the linear drop probability scheduler ramps up to the desired $p$ (see Fig.~\ref{fig:drop_scheduler_comparison}).
Therefore, for the results reported in Figs.~\ref{fig:1d_dropout_stats_ex1} and \ref{fig:1d_dropout_stats_ex2}, no drop probability scheduler was applied. 

\begin{figure}[!ht]
\centering
\subfigure[]{\includegraphics[width=0.45\textwidth]{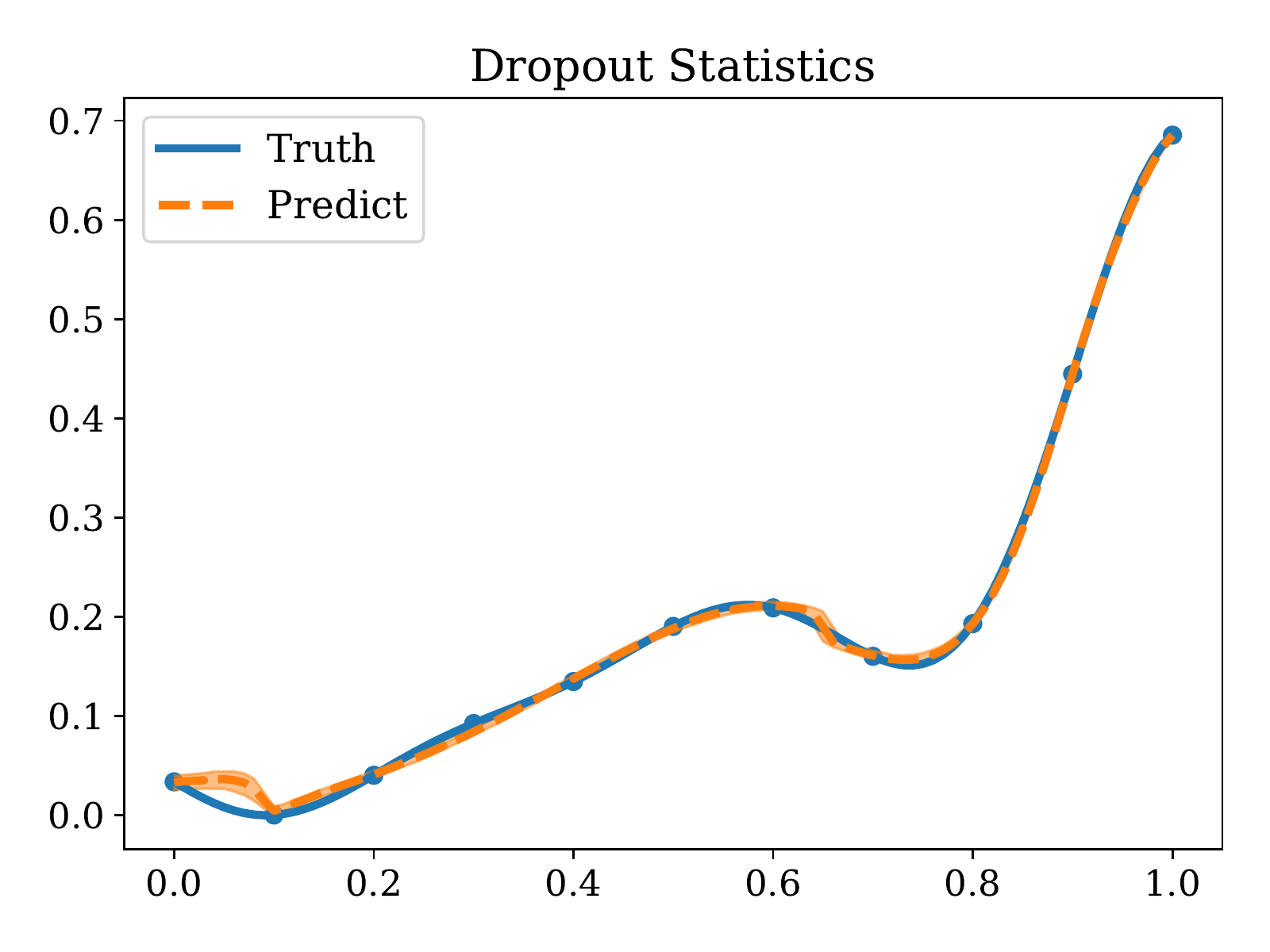} \label{fig:ex1_dropout_lf_omitdrop}}
\subfigure[]{\includegraphics[width=0.45\textwidth]{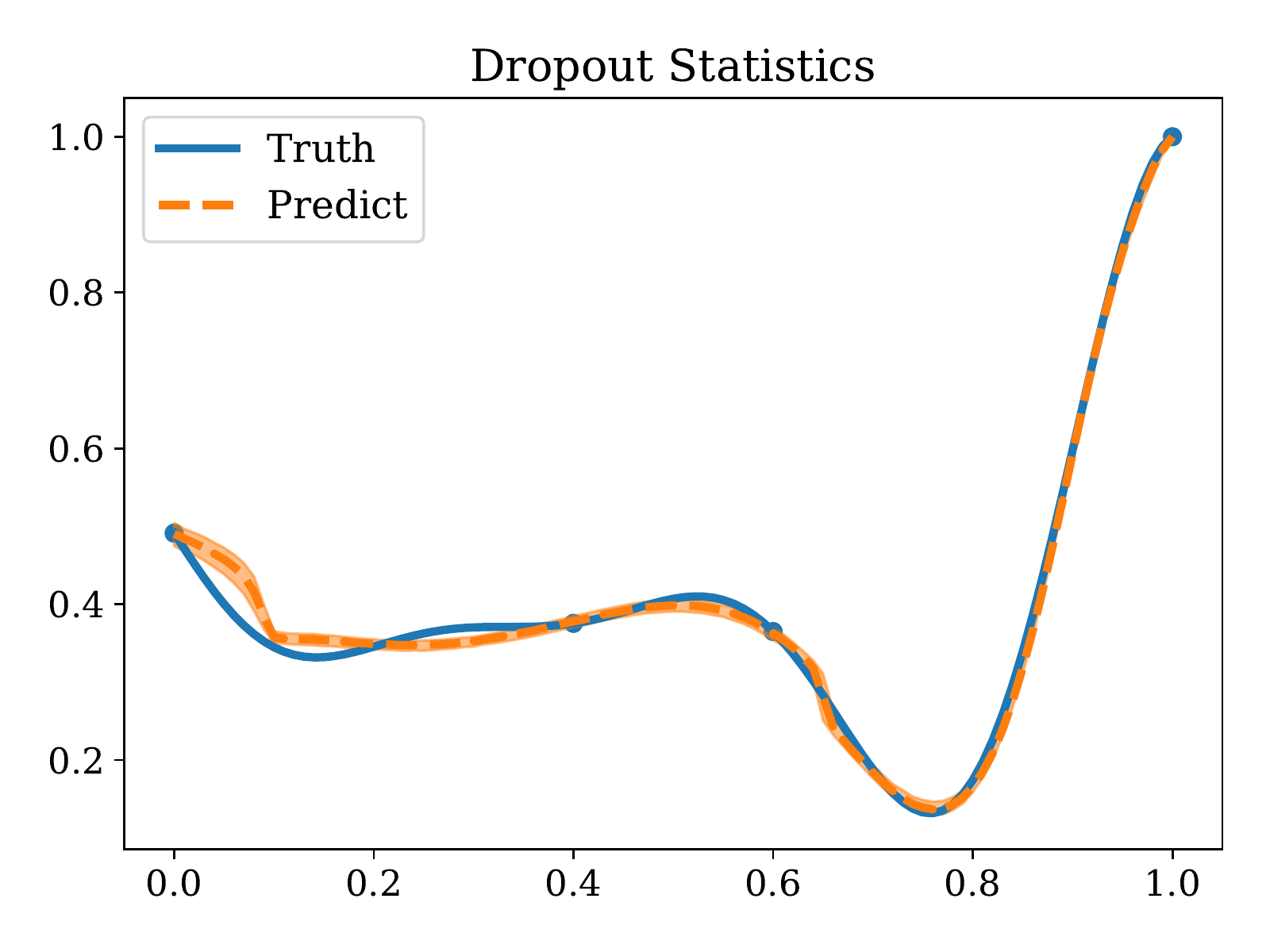} \label{fig:ex1_dropout_hf_omitdrop}}
\caption{Dropout layer after every convolution layer, \textit{except} the first two convolution layers, preceding the LF prediction. Mean prediction and estimated 5\%-95\% percentiles for (a) Eq.~\eqref{LMP:equ:example1_lf} and (b) Eq.~\eqref{LMP:equ:example1_hf}, from a network with multiple DropBlock realizations. In these tests we used a tanh activation function.}\label{fig:1d_dropout_stats_ex1_omitdrop}
\end{figure}

\begin{figure}[!ht]
\centering
\subfigure[]{\includegraphics[width=0.45\textwidth]{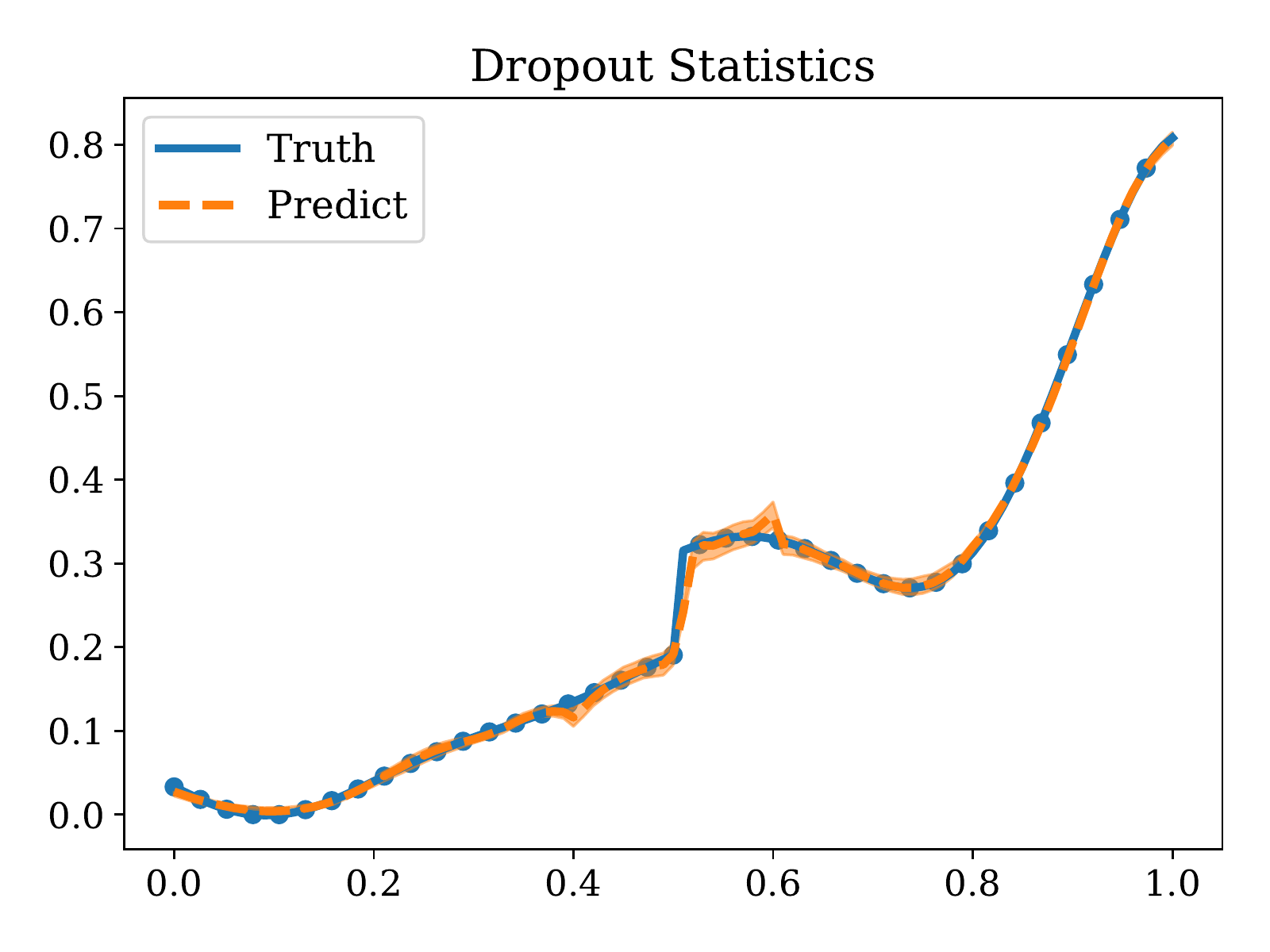} \label{fig:ex2_dropout_lf_omitdrop}}
\subfigure[]{\includegraphics[width=0.45\textwidth]{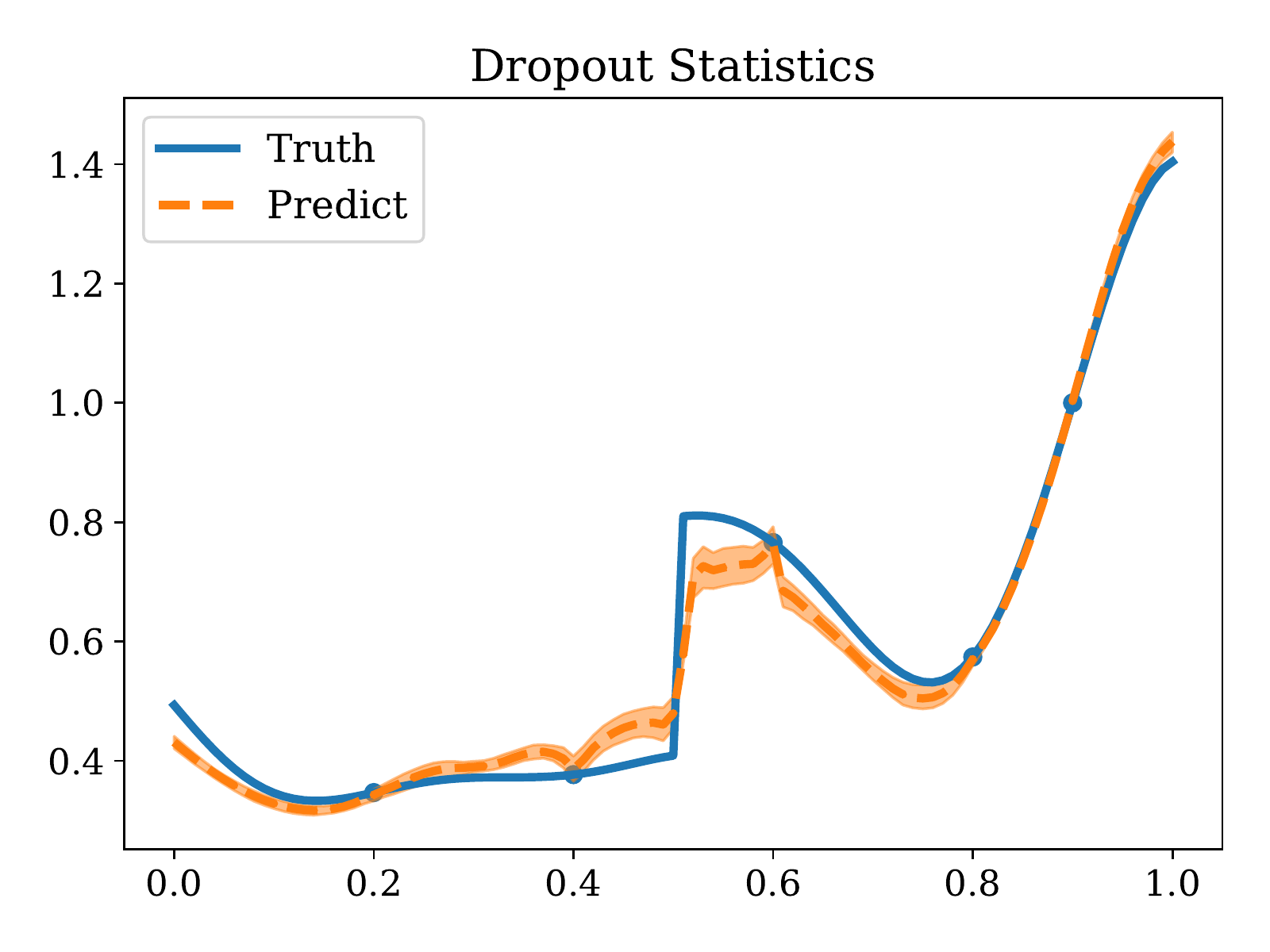} \label{fig:ex2_dropout_hf_omitdrop}}
\caption{Dropout layer after every convolution layer, \textit{except} the first four convolution layers, preceding the LF prediction. Mean prediction and estimated 5\%-95\% percentiles for (a) Eq.~\eqref{LMP:equ:example2_lf} and (b) Eq.~\eqref{LMP:equ:example2_hf}, from a network with multiple DropBlock realizations. In these tests we used a tanh activation function.}\label{fig:1d_dropout_stats_ex2_omitdrop}
\end{figure}

Large uncertainty intervals resulted by adding a DropBlock layer after each convolutional layer, as shown in Fig.~\ref{fig:1d_dropout_stats_ex1} and Fig.~\ref{fig:1d_dropout_stats_ex2}. However, this generated inaccurate individual realizations, due to excessive regularization.
We then investigated the possibility to selectively remove DropBlock layers in order to minimize the mean square error over all DropBlock realizations for all training examples, with results shown in Fig.~\ref{fig:1d_dropout_stats_ex1_omitdrop} and Fig.~\ref{fig:1d_dropout_stats_ex2_omitdrop}.
As expected, the uncertainty intervals are extremely narrow at the training points. 
In between the training points, however, we see varying levels of uncertainty. Although the uncertainty does not always capture the underlying true response, it provides an indication of where data should be selected to improve the network accuracy. 
For example, in Figs.~\ref{fig:ex1_dropout_lf_omitdrop} and \ref{fig:ex1_dropout_hf_omitdrop}, we see the largest uncertainty near $x=0.05$, which is also where we see the largest error in the predictions. 
Similarly, in Fig.~\ref{fig:ex2_dropout_hf_omitdrop}, we see large uncertainty near $x=0.45$, $x=0.55$ surrounding the discontinuity.

\begin{figure}
\centering
\subfigure[]{\includegraphics[width=0.45\textwidth]{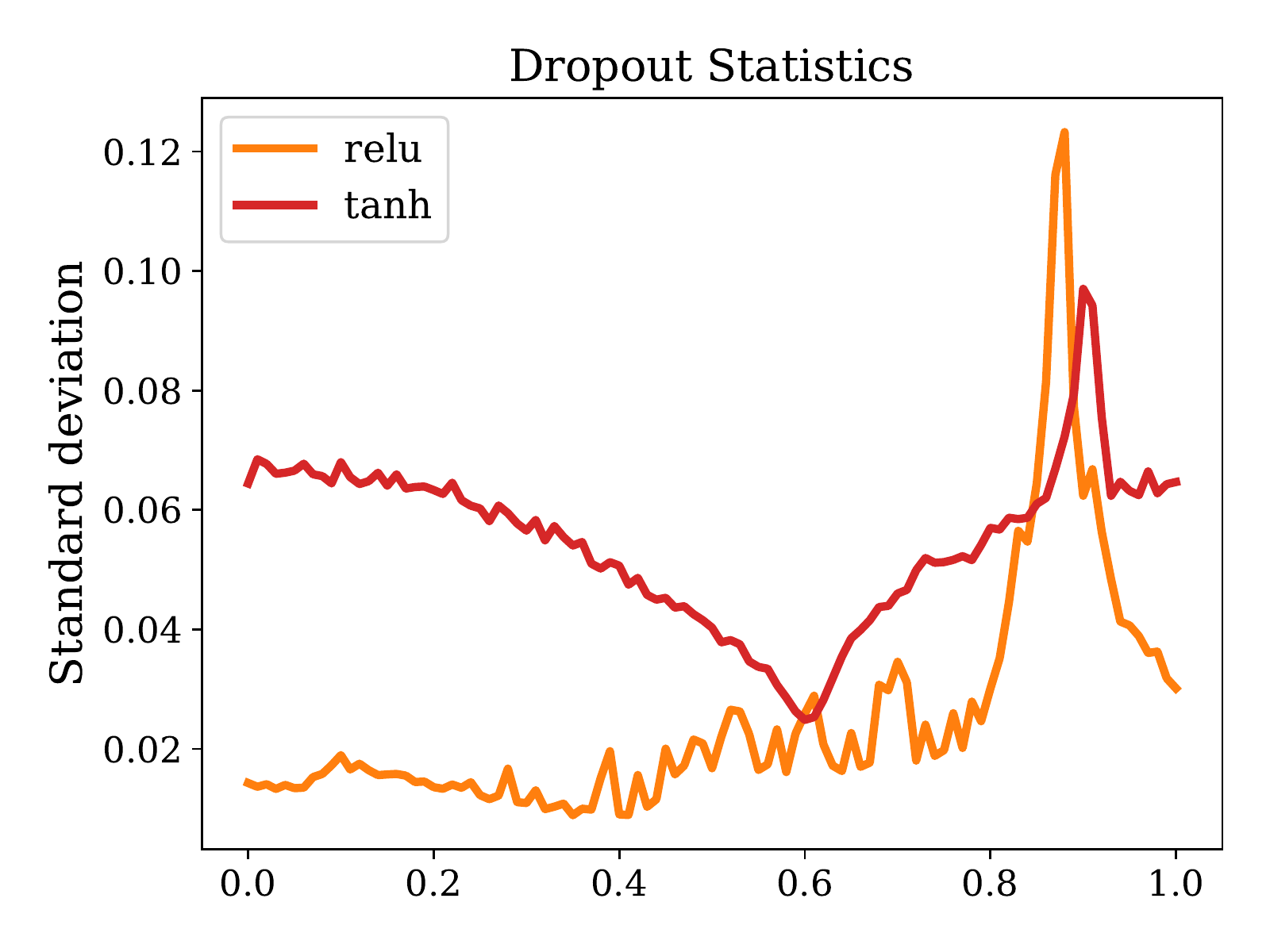}}
\subfigure[]{\includegraphics[width=0.45\textwidth]{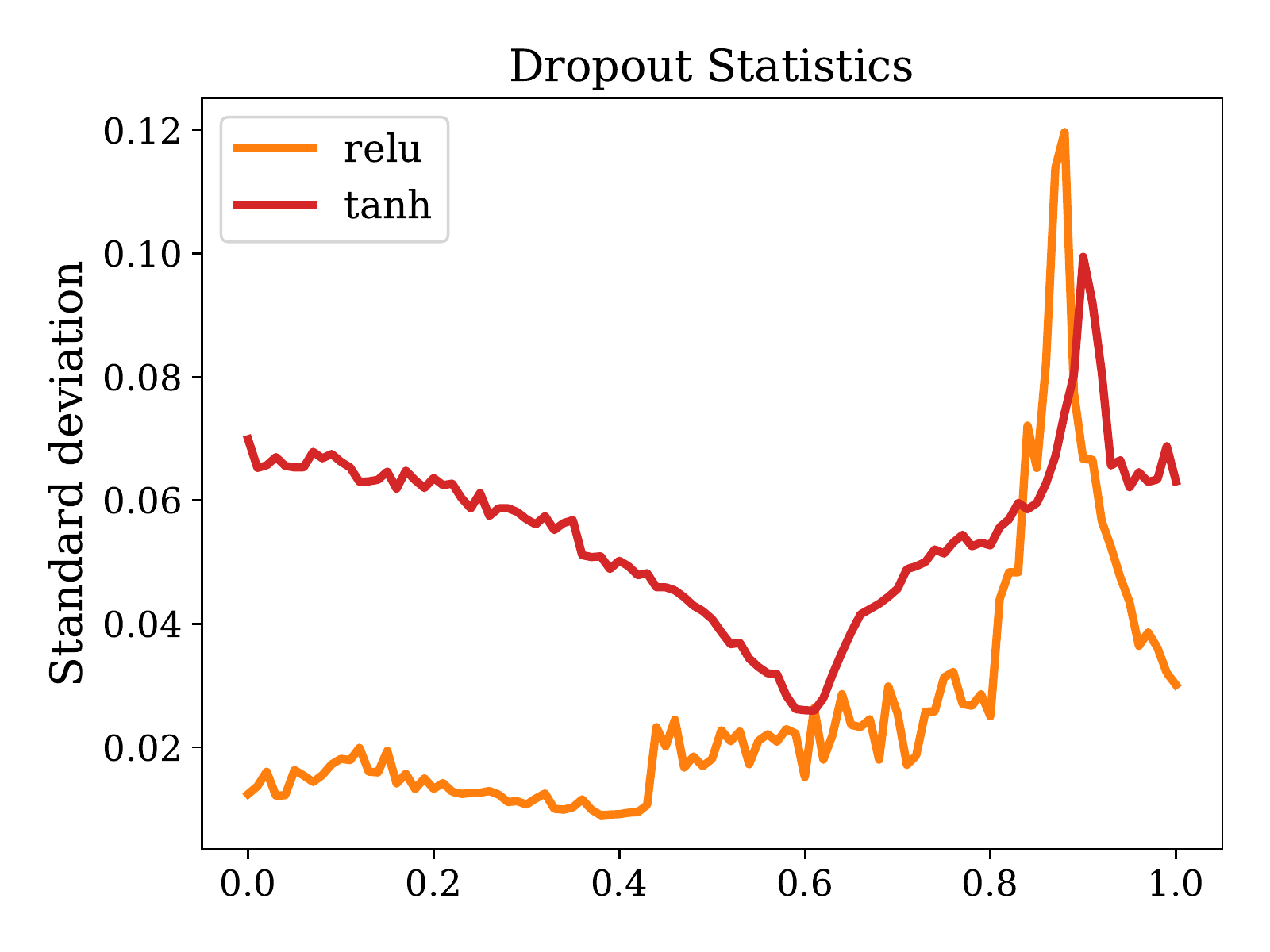}}
\caption{Tanh vs. ReLU activation functions comparison of standard deviation from multiple DropBlock realizations. Dropout layer after \textit{every} convolution layer preceding the LF prediction. (a) LF Eq.~\eqref{LMP:equ:example1_lf} and (b) HF Eq.~\eqref{LMP:equ:example1_hf}.} \label{fig:1d_stdev_tanhvsrelu_ex1}
\end{figure}

\begin{figure}
\centering
\subfigure[]{\includegraphics[width=0.45\textwidth]{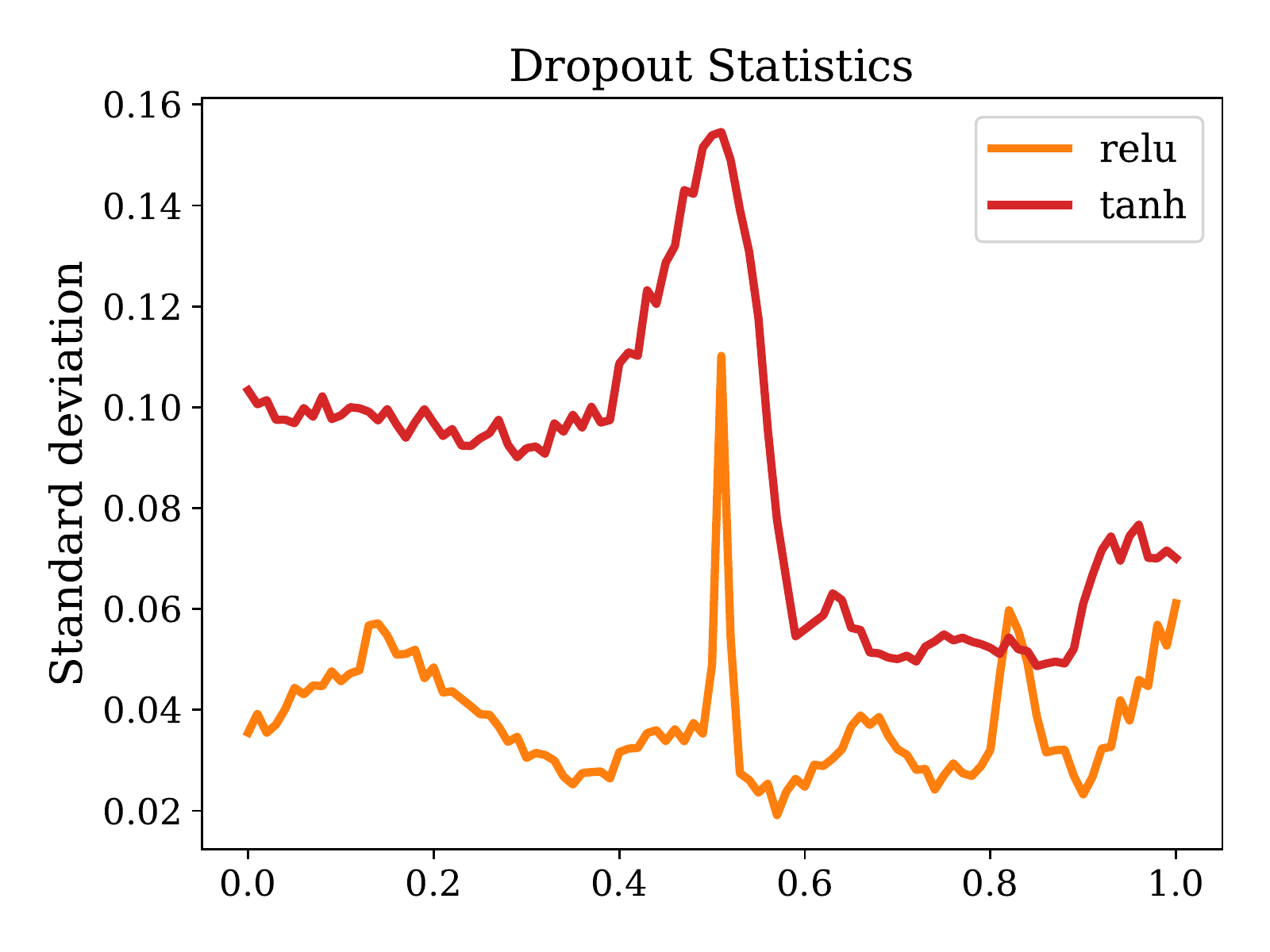}}
\subfigure[]{\includegraphics[width=0.45\textwidth]{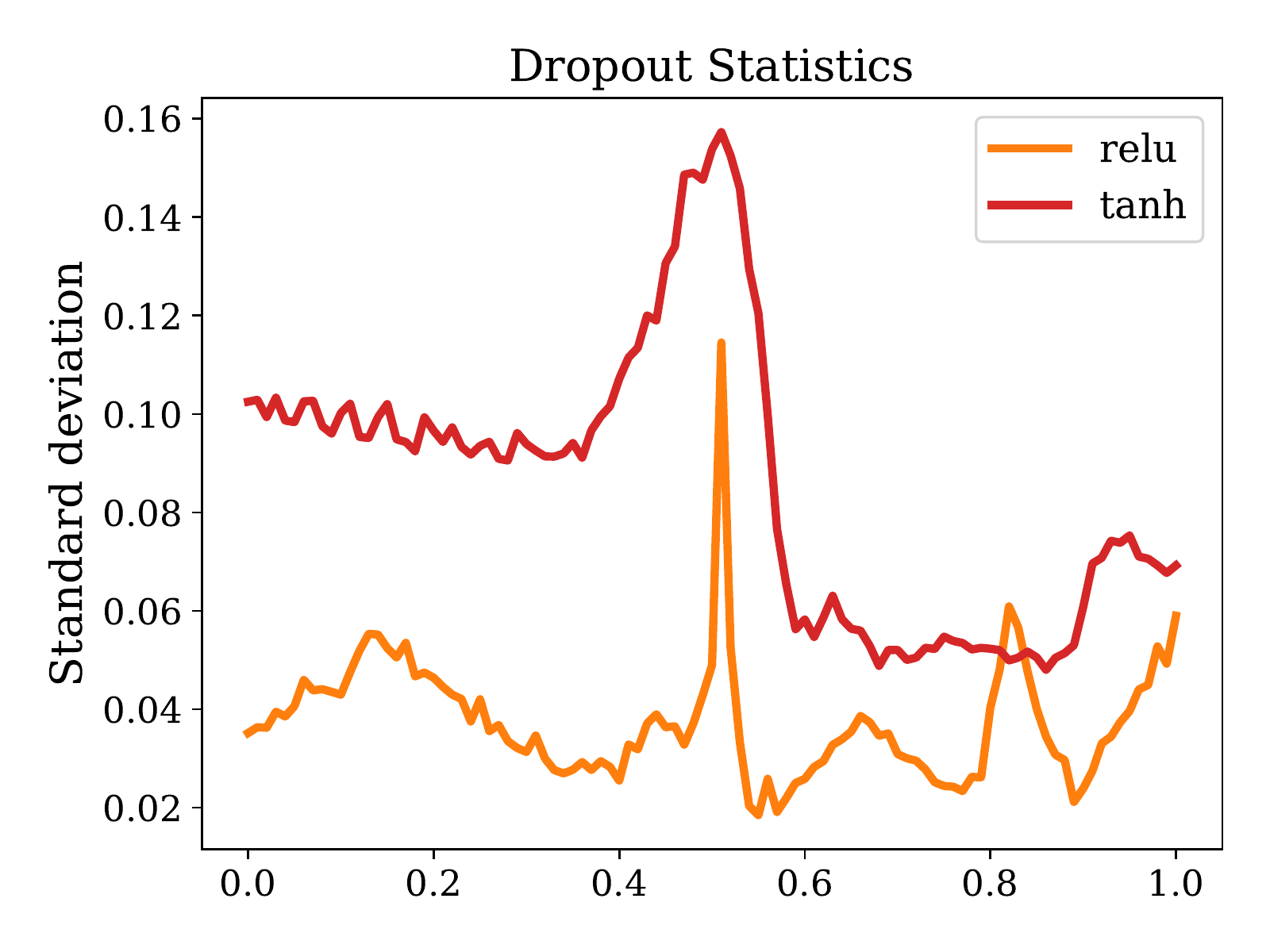}}
\caption{Tanh vs. ReLU activation functions comparison of standard deviation from multiple DropBlock realizations. Dropout layer after \textit{every} convolution layer preceding the LF prediction. (a) LF Eq.~\eqref{LMP:equ:example2_lf} and (b) HF Eq.~\eqref{LMP:equ:example2_hf}.}\label{fig:1d_stdev_tanhvsrelu_ex2}
\end{figure}

Uncertainty bounds predicted using ReLU or tanh activation functions appear similar when multiple Dropblocks are introduced in the network, and maximum uncertainty is achieved at the same values of $x$, as shown in Fig.~\ref{fig:1d_stdev_tanhvsrelu_ex1} and \ref{fig:1d_stdev_tanhvsrelu_ex2}.
Slightly smaller standard deviations are produced by ReLU due to its greater flexibility (due to scale invariance and lack of saturation) and since single DropBlock realizations are included in the loss function.
Moreover, we observe almost identical patterns for the standard deviation in the LF and HF functions, as expected. Since all DropBlock layers are located before the LF prediction, HF prediction uncertainty is propagated identical from the LF representation.
ReLU tends to spike to a larger value within a small interval, which may relate to its unbounded property.

Addition of LF training data leads to a reduction in the estimated uncertainty, but preliminary tests suggest a re-calibration of the hyperparameters to be essential for this to occur.

\subsubsection{Dense regression} \label{high_dim_uq}

\begin{figure}[!ht]
\centering
\subfigure[Sample 1]{\includegraphics[width=0.32\textwidth]{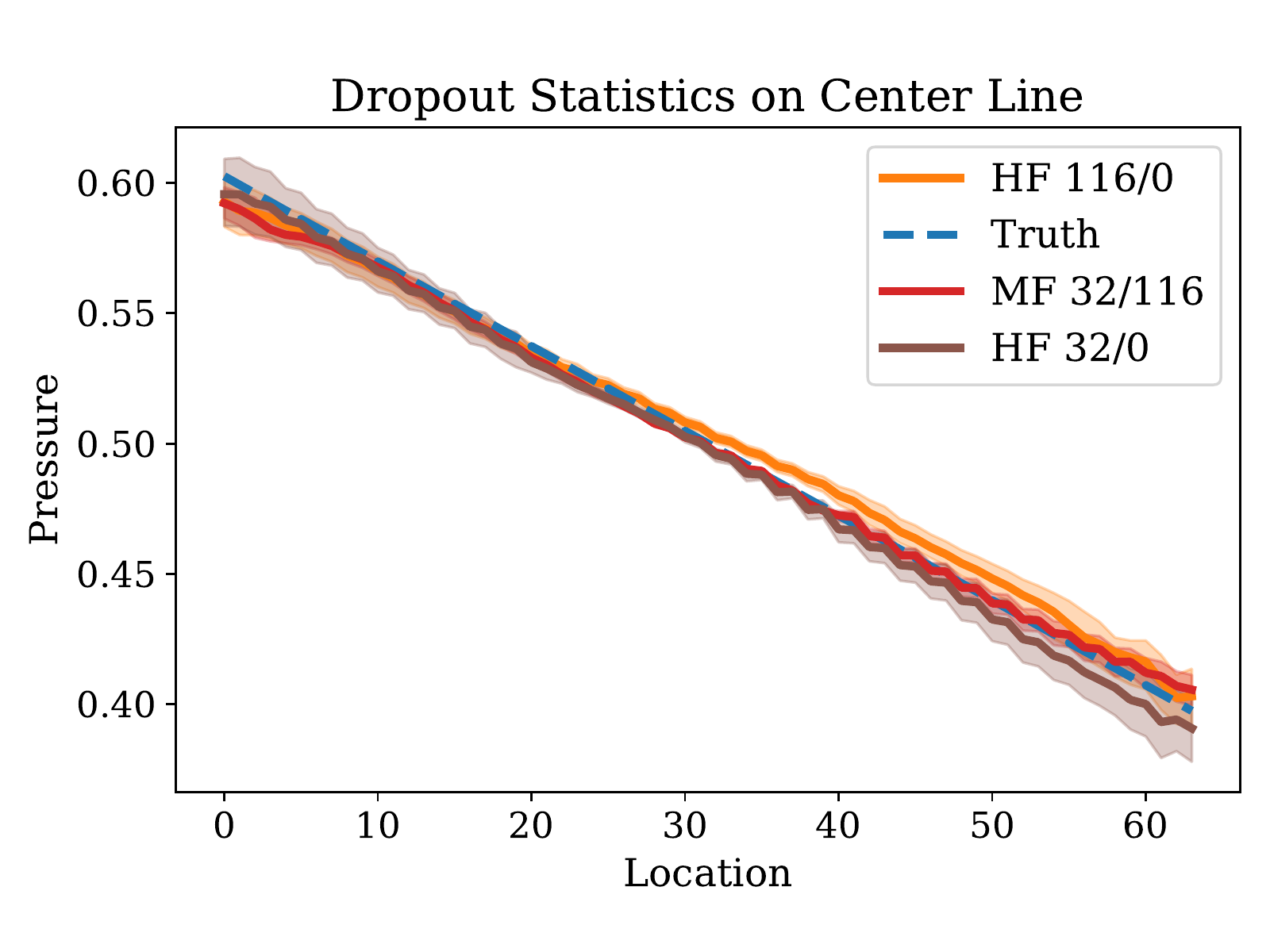}}
\subfigure[Sample 2]{\includegraphics[width=0.32\textwidth]{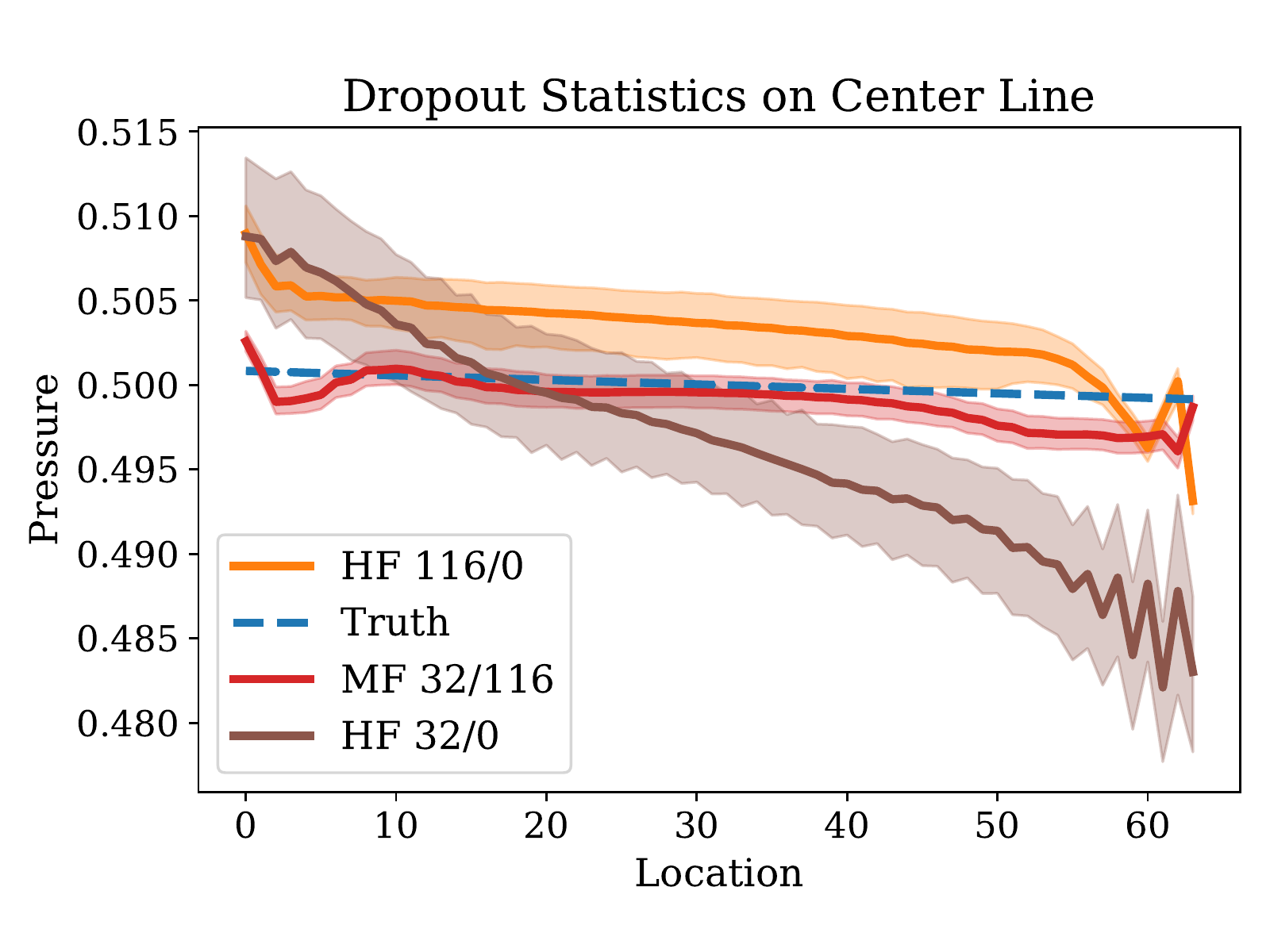} \label{LMP:fig:poiseuille_n12_drop}}
\subfigure[Sample 3]{\includegraphics[width=0.32\textwidth]{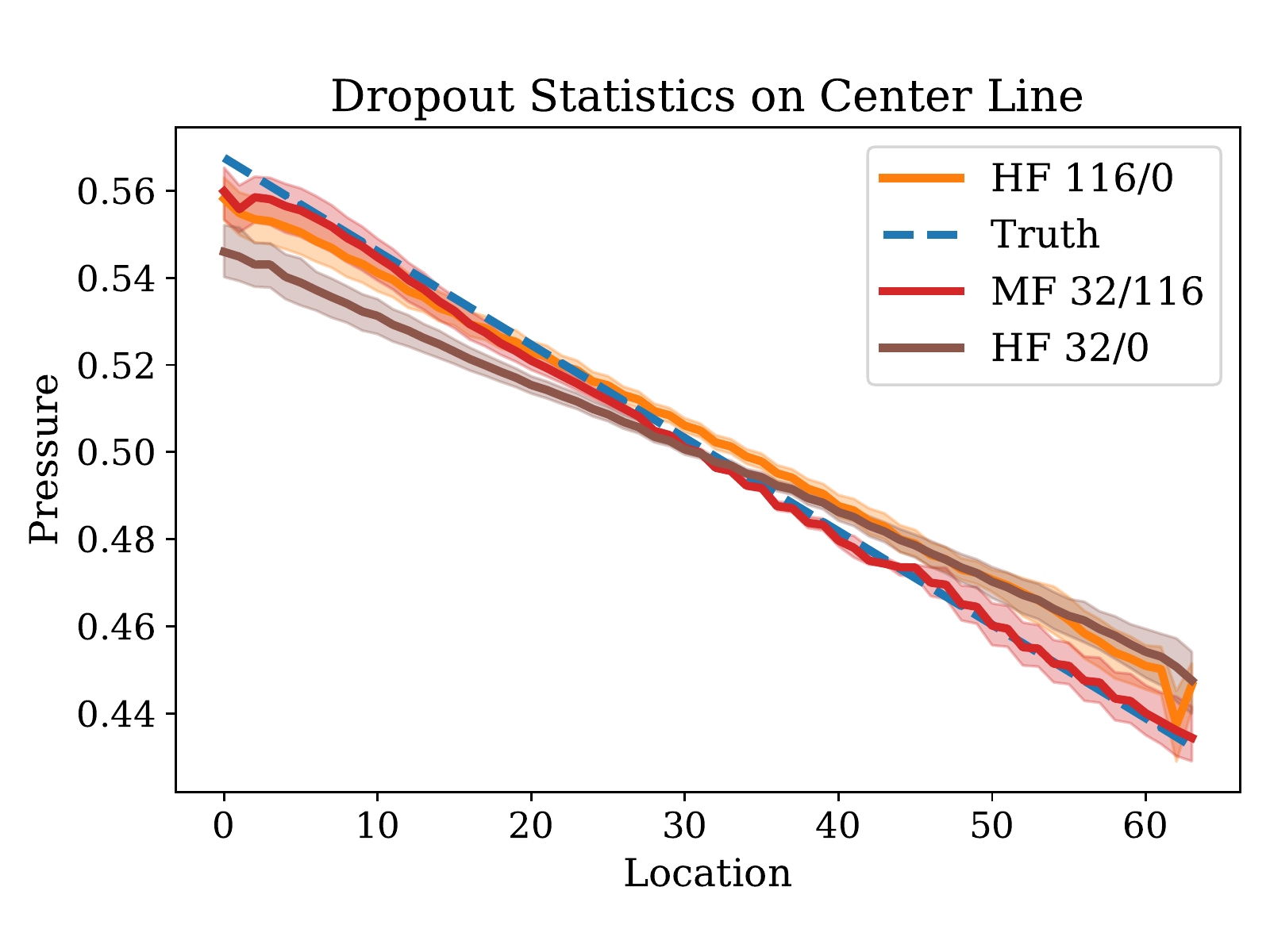} \label{LMP:fig:poiseuille_n29_drop}}
\caption{Dense regression network. Mean prediction and 5\%-95\% percentiles from an ensemble of 1000 DropBlock pressure realizations for three test data examples sliced along the cylinder centerline.}\label{LMP:fig:poiseuille_dropout_stats_centerline}
\end{figure}

\noindent Uncertainty estimates for the pressure along the centerline of the fluid region from $N_{\text{UQ}}$ realizations are shown in Fig.~\ref{LMP:fig:poiseuille_dropout_stats_centerline}. 
As is most evident in Fig.~\ref{LMP:fig:poiseuille_n12_drop}, we see that the HF 32/0 produces less accurate predictions and wider uncertainty intervals.

\begin{figure}[!ht]
\centering
\subfigure[]{\includegraphics[width=0.4\textwidth]{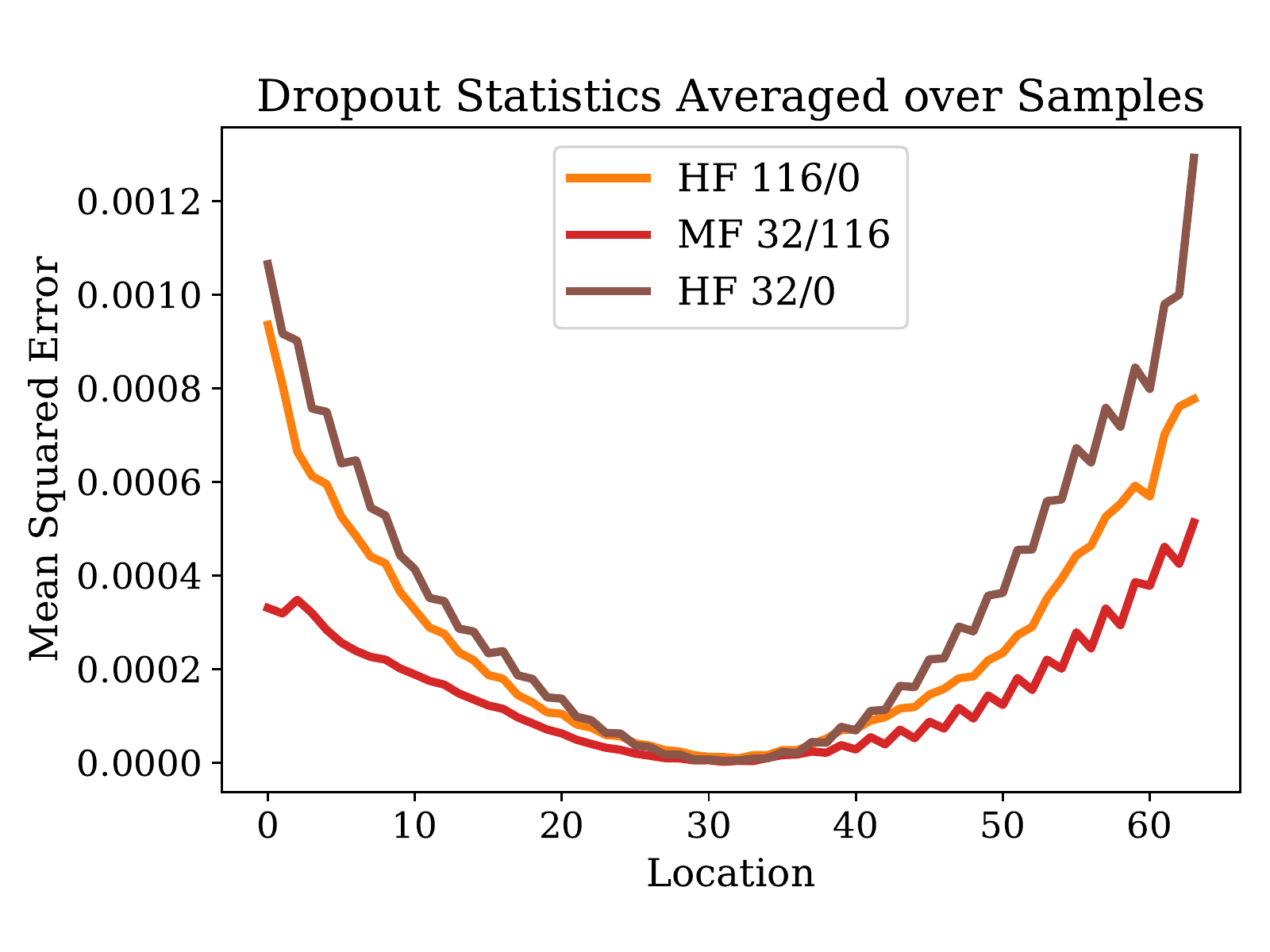}}
\subfigure[]{\includegraphics[width=0.4\textwidth]{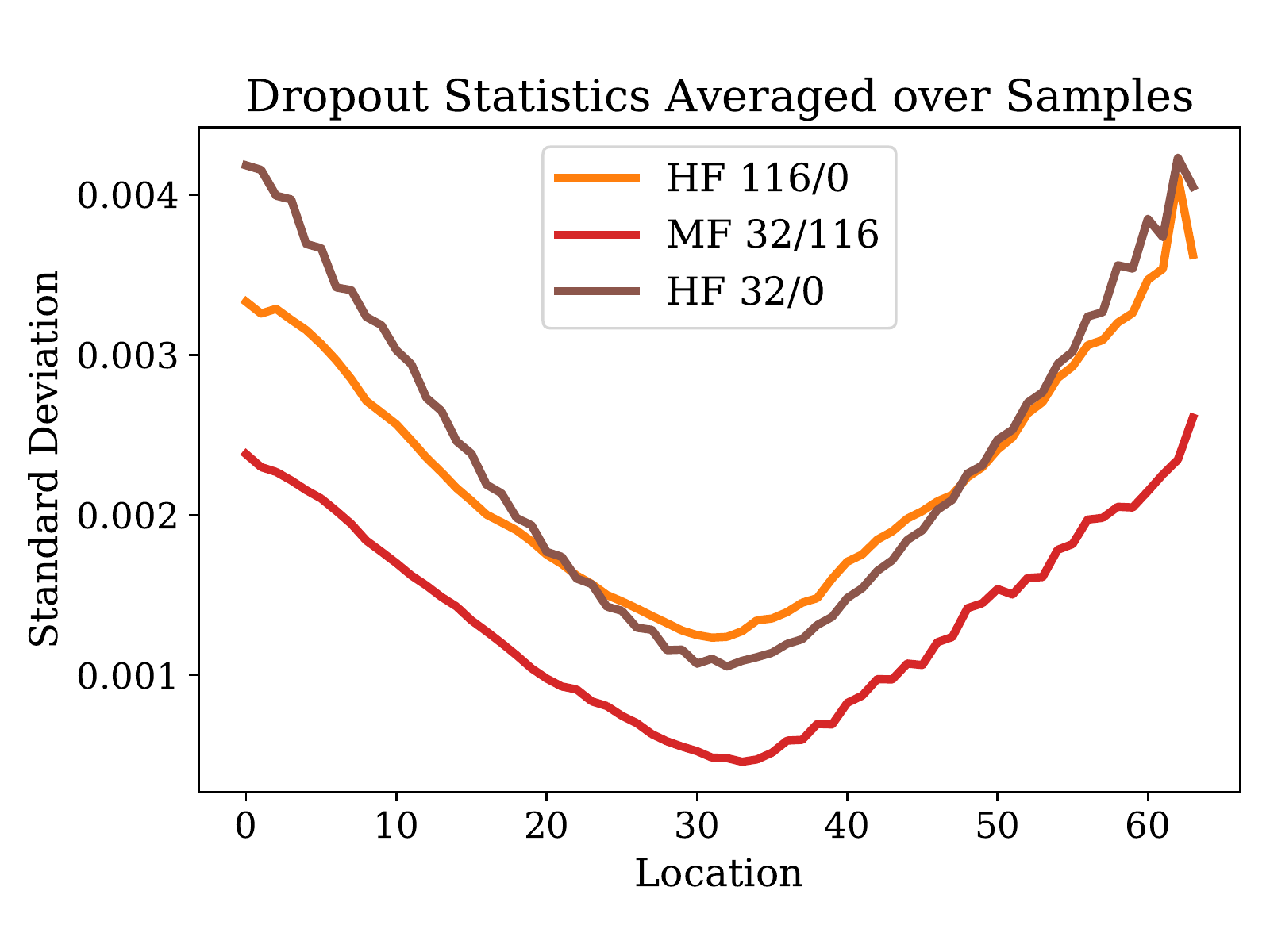}}
\caption{Dense regression network. Mean square error and standard deviation resulting from 1000 network evaluations with 10 DropBlock layers in each network. Quantities are calculated along the axis of the cylindrical fluid domain. The optimal network is selected using the minimum validation loss calculated by activating all DropBlock layers.} \label{fig:poiseuille_dropout_stats_network_comparison}
\end{figure}

In Fig.~\ref{fig:poiseuille_dropout_stats_network_comparison} we plot the mean square error (MSE) and standard deviation of $N_{\text{UQ}}$ MC-DropBlock realizations for every example in the test set, and plot them versus the corresponding location along the centerline of the cylindrical fluid domain.
Both the variance and the MSE appear parabolic, consistent with the pressure results being approximately 0.5 for all samples at the center of the fluid domain, due to the way samples are normalized. Also, error and uncertainty increase away from the center due to the slope variability in the true pressure profiles.
MF networks are observed to produce lower errors and limited uncertainty with respect to HF networks. The HF 32/0 is characterized by both the highest variance and highest error, which is consistent with the limited amount of data available during training.

\begin{figure}[!ht]
\centering
\subfigure[HF 32/0]{\includegraphics[width=0.3\textwidth]{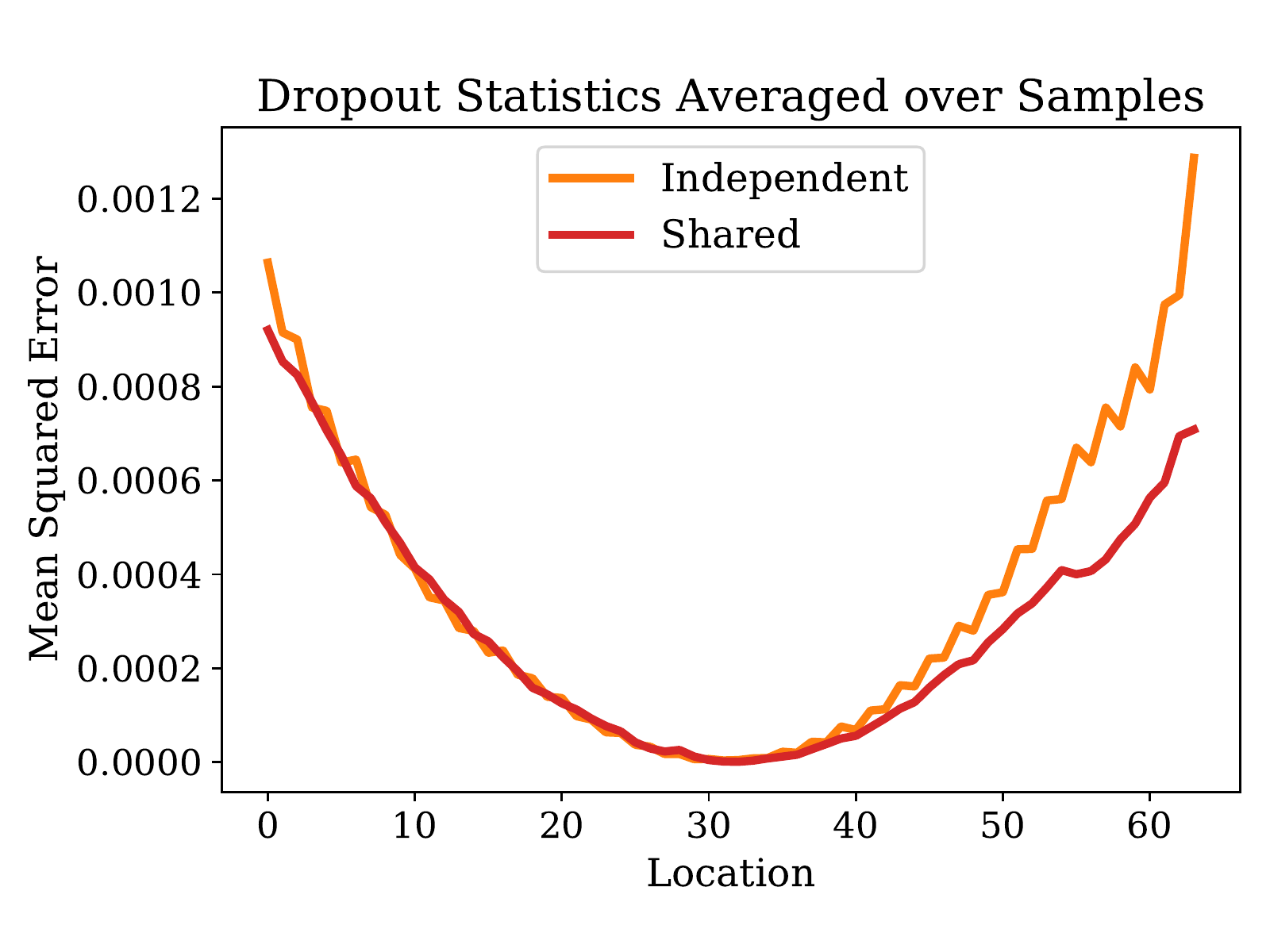}}
\subfigure[MF 32/116]{\includegraphics[width=0.3\textwidth]{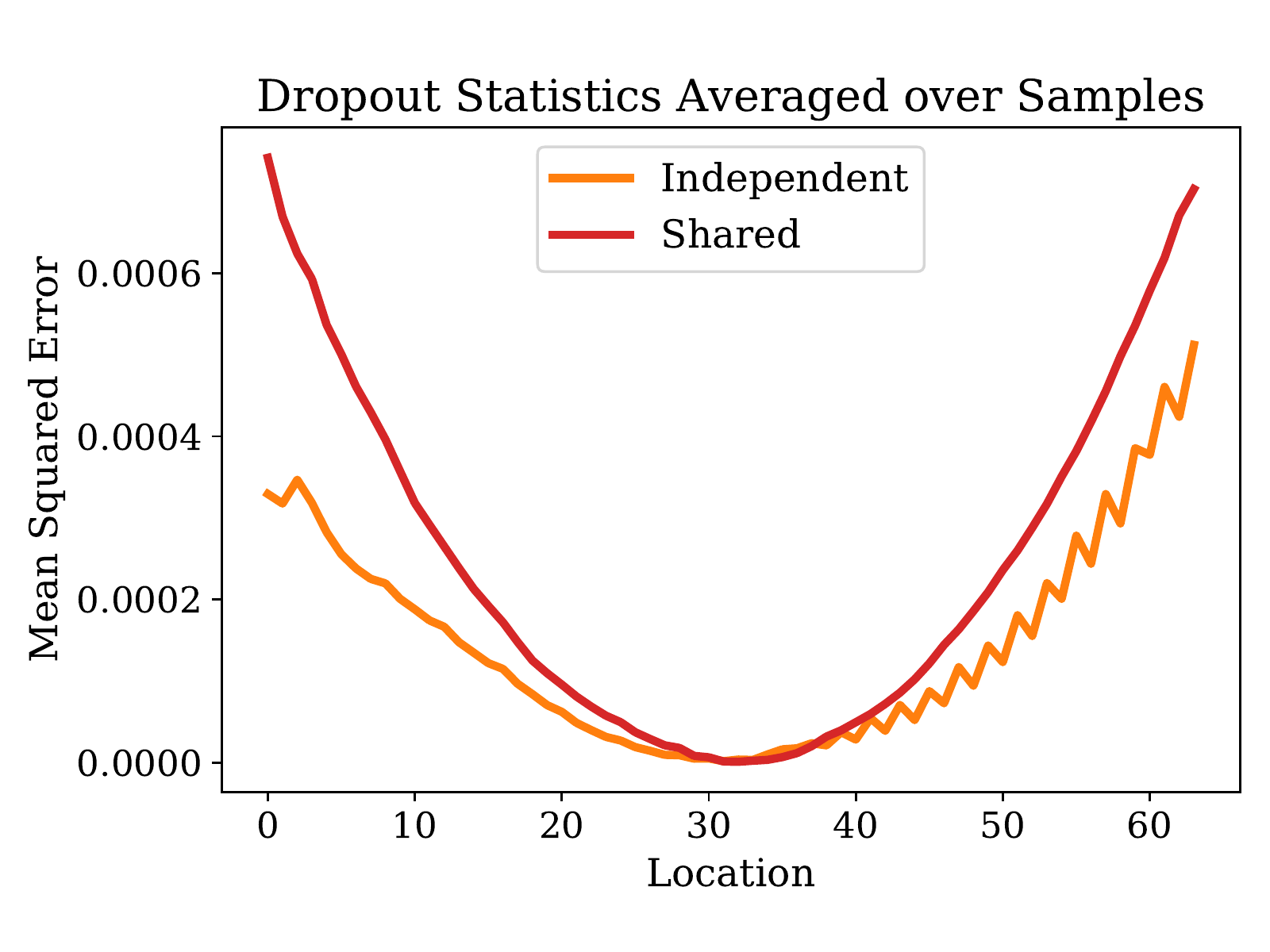}}
\subfigure[HF 116/0]{\includegraphics[width=0.3\textwidth]{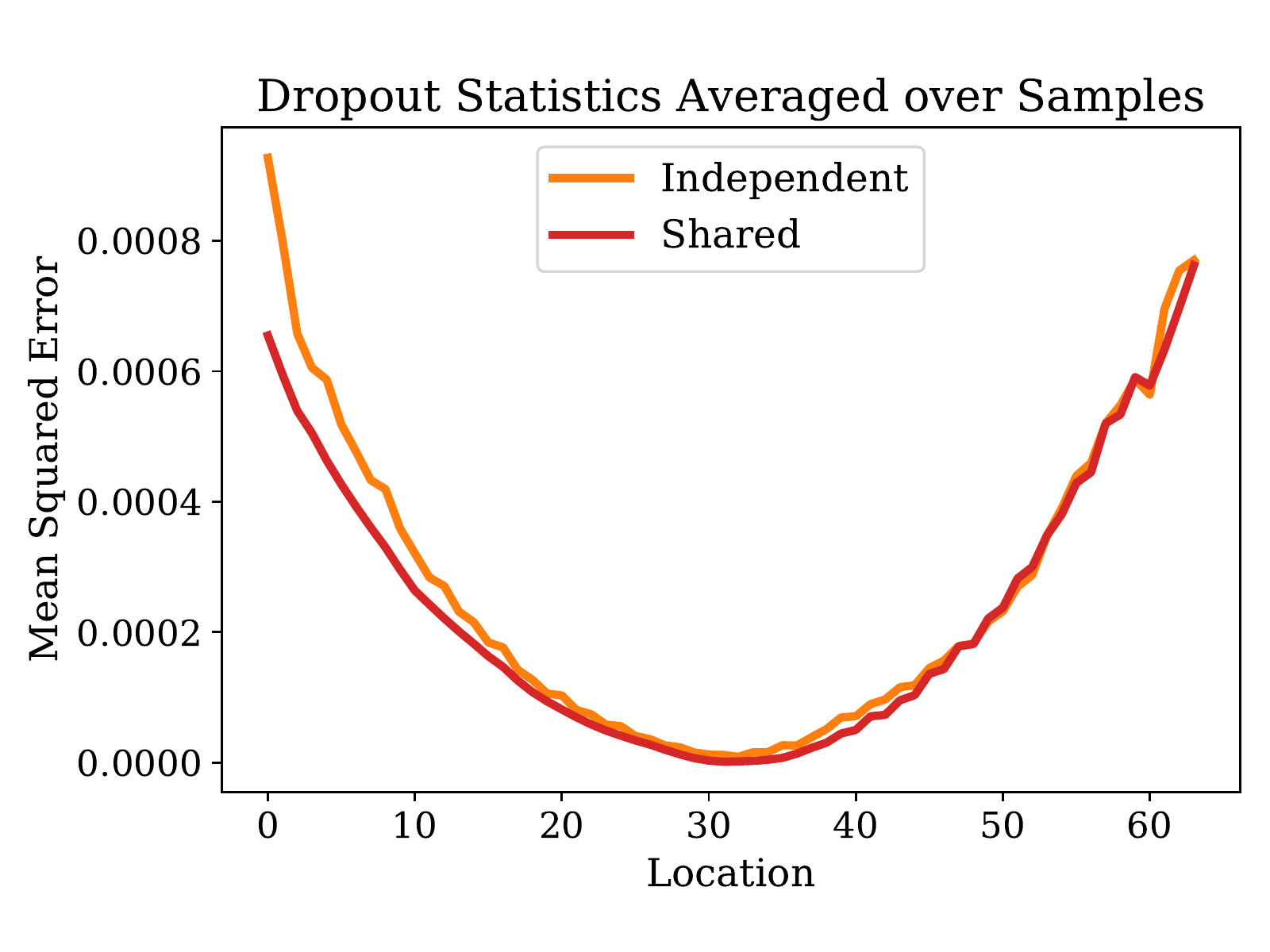}}
\subfigure[HF 32/0]{\includegraphics[width=0.3\textwidth]{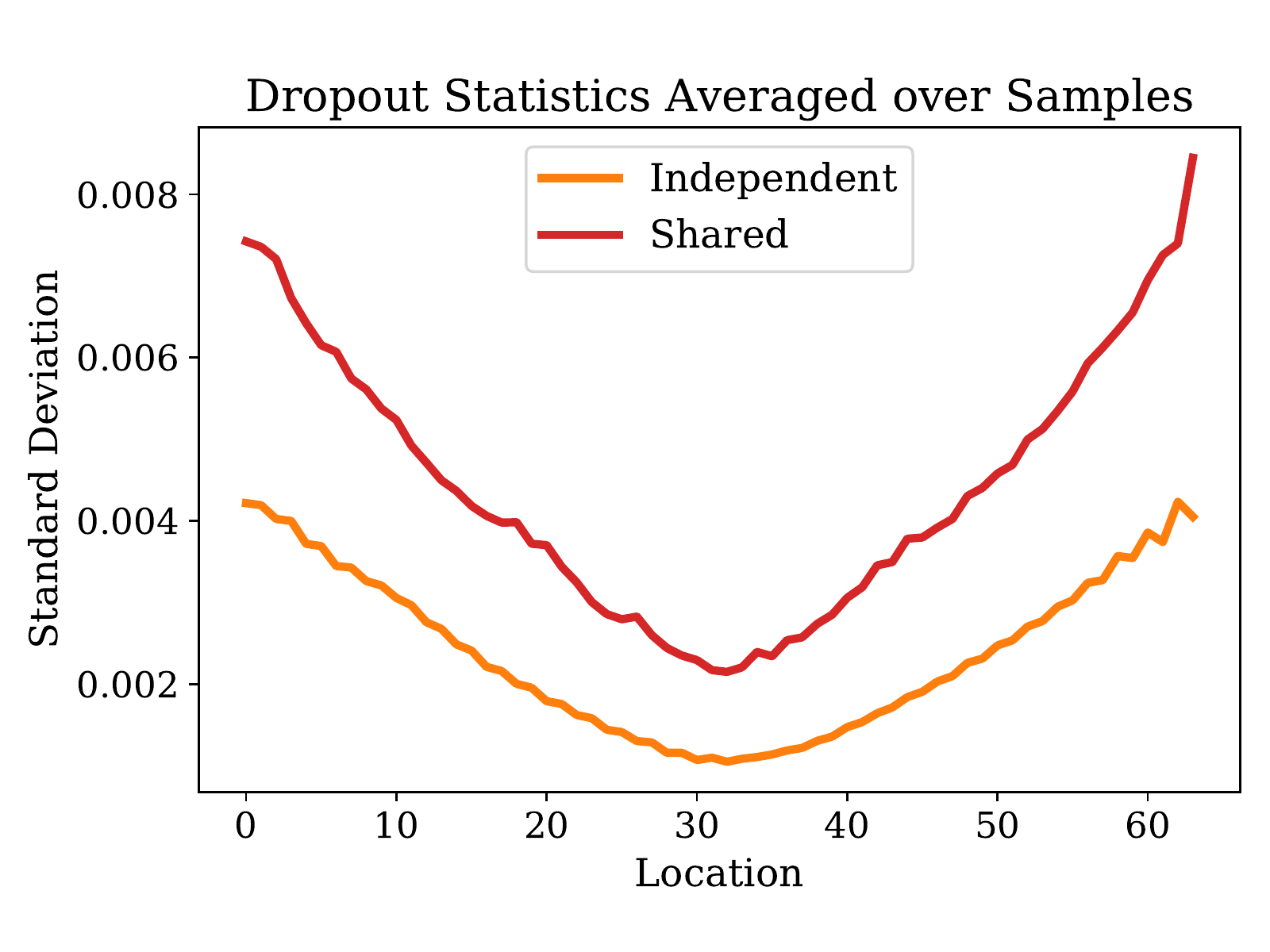}}
\subfigure[MF 32/116]{\includegraphics[width=0.3\textwidth]{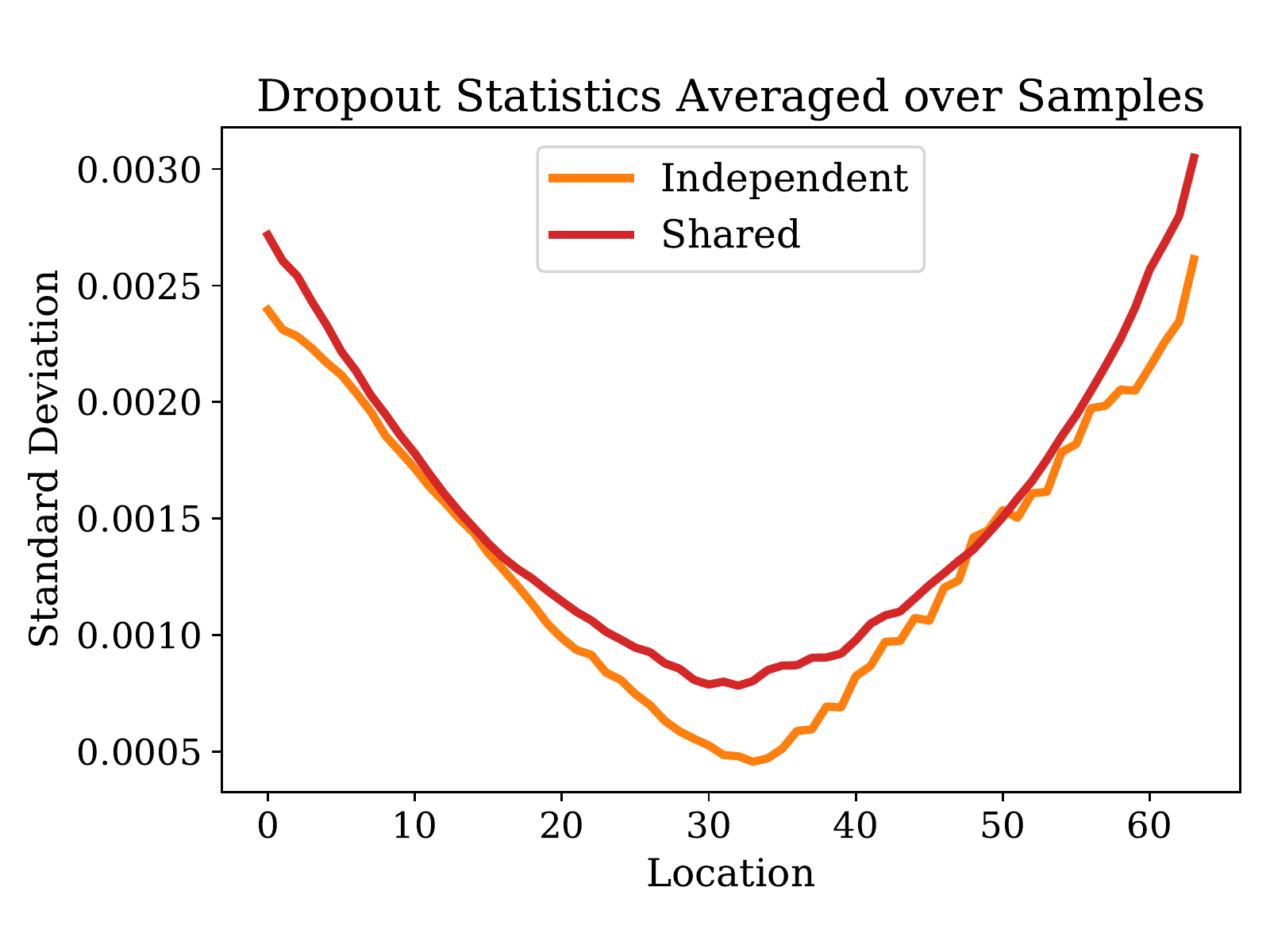}}
\subfigure[HF 116/0]{\includegraphics[width=0.3\textwidth]{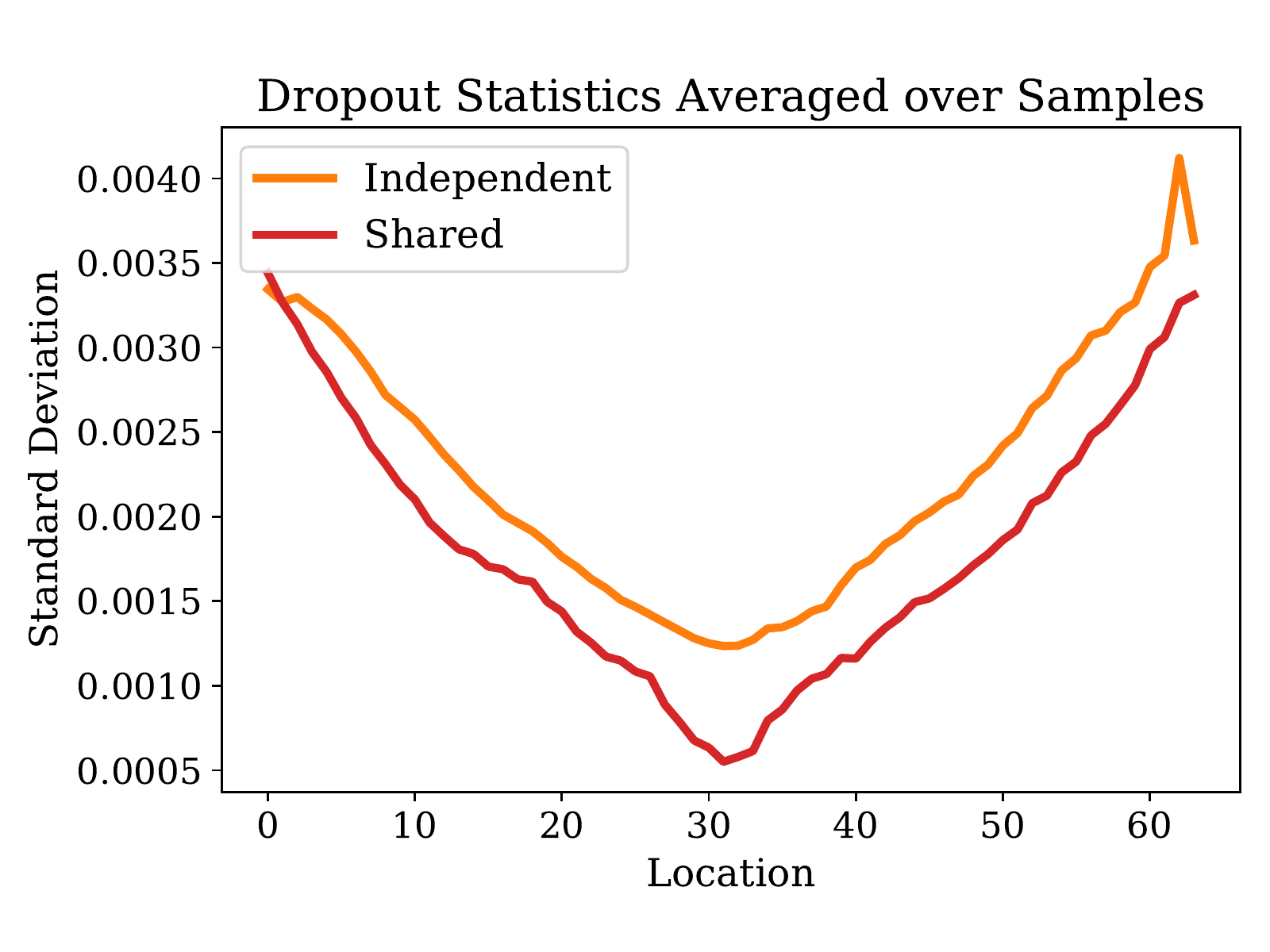}}
\caption{Dense regression network. Mean squared errors (top row) and standard deviations (bottom row) from 1000 DropBlock realizations along the axis of the cylindrical fluid domain for networks containing 10 DropBlock layers, The optimal network is selected using the minimum validation loss calculated by activating all DropBlock layers.}\label{fig:poiseuille_dropout_stats_all_samples_sharedvsind}
\end{figure}

We also compare the results from DropBlock masks being shared or independent across feature channels.
The study in~\cite{ghiasi2018dropblock} concludes that independent DropBlock masks work better, but it only analyzes the resulting network accuracy and not prediction uncertainty. Our results in Fig.~\ref{fig:poiseuille_dropout_stats_all_samples_sharedvsind} show comparable accuracy for shared and independent masks.

\subsubsection{Low- to high-dimensional regression}
\begin{figure}[!ht]
\centering
\subfigure[]{\includegraphics[width=0.4\textwidth]{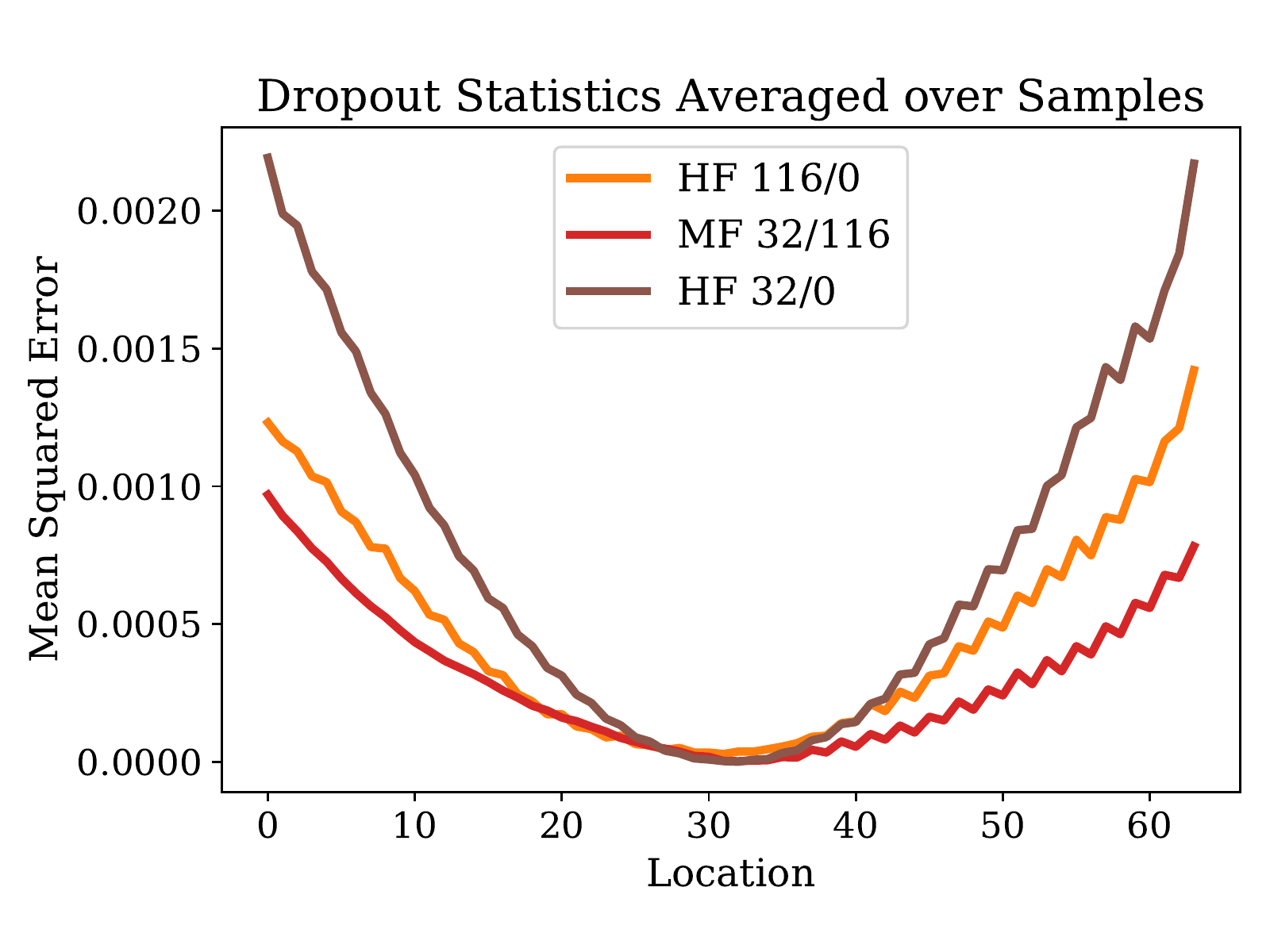}}
\subfigure[]{\includegraphics[width=0.4\textwidth]{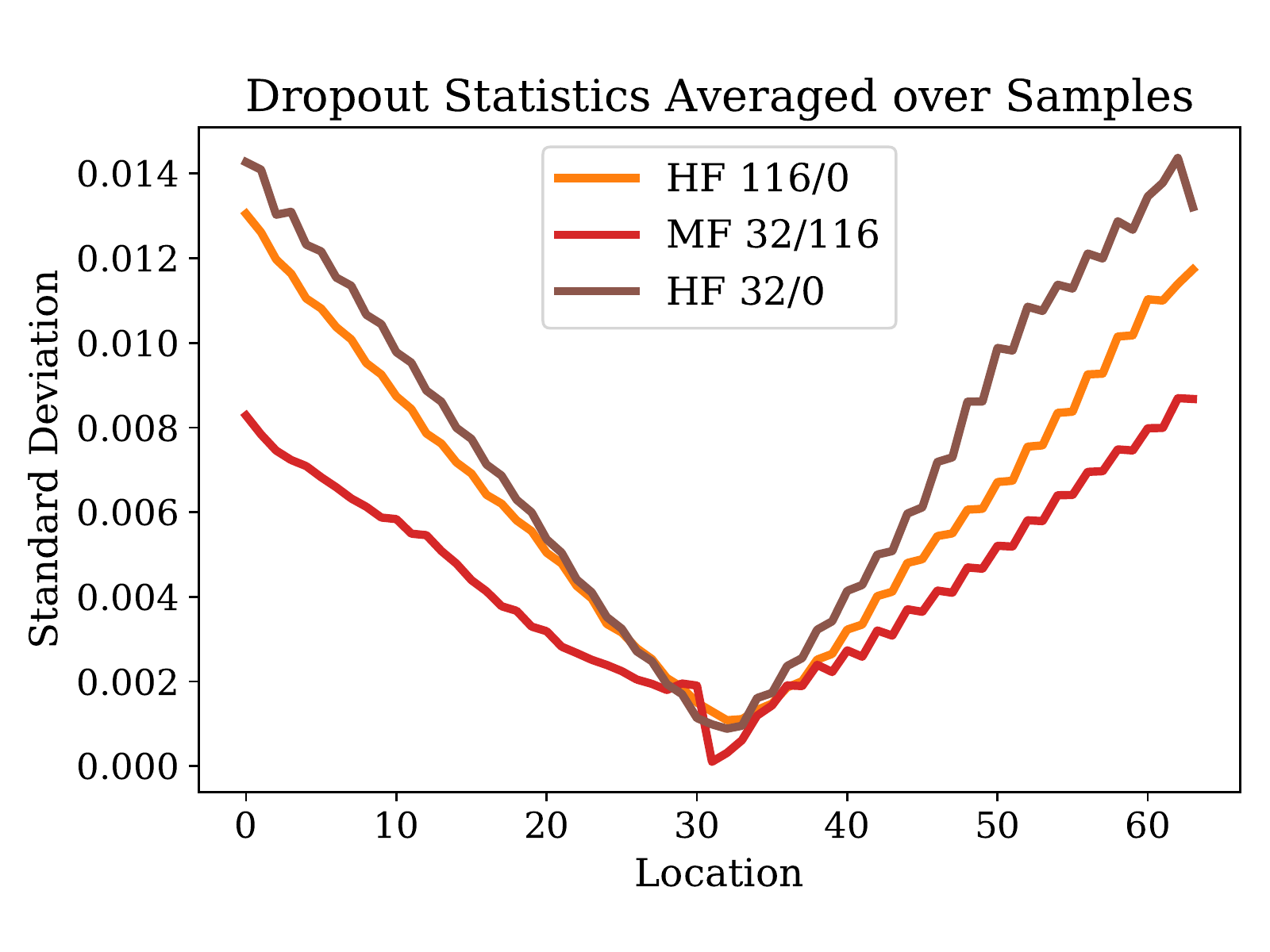}}
\caption{Low- to high-dimensional case: Mean square error and standard deviation resulting from 1000 network evaluations with 6 DropBlock layers in each network. Quantities are calculated along the axis of the cylindrical fluid domain. The optimal network is selected using the minimum validation loss calculated by activating all DropBlock layers.} \label{fig:poiseuille_params_dropout_stats_network_comparison}
\end{figure}

\noindent In Fig.~\ref{fig:poiseuille_params_dropout_stats_network_comparison} we plot the mean square error (MSE) and standard deviation of the MC-DropBlock realizations across test samples versus their location along the axis of the cylindrical fluid domain. Similarly to Section~\ref{high_dim_uq}, the MF network produces results as accurate as the HF 116/0 network, while also producing lower variance than the HF networks. Some of this difference could be attributed to different drop probabilities, since higher drop probability tended to produce higher variance in our experiments. Both the variance and MSE appear parabolic, consistent with the conclusions presented in Section~\ref{high_dim_uq}.

\section{Conclusions and future work} \label{LMP:sec:conclusionsfuture}

\noindent In this work, we focus on convolutional neural networks, specifically architectures resulting from an assembly of encoders, decoders and skip connections. 
Such architectures have the flexibility to predict the results from HF physics-based solvers having either high-dimensional inputs, high-dimensional outputs or both, using only a fraction of the weights that would be required by fully connected networks.
If trained from a few expensive HF and many inexpensive LF examples, they also exhibit a comparable performance to fully-connected multifidelity networks for one-dimensional function approximation and produce a consistently accurate performance for high-dimensional inputs/outputs.

In this context, using two test cases in one-dimensional functional approximation and the solution of the pressure Poisson equation, respectively, we show that multifidelity networks produce, at a reduced cost, a validation accuracy comparable to that of networks trained from a much larger number of high-fidelity realizations.
Use of datasets containing examples from multiple fidelities also accelerates training, leading to significant loss reductions early on during the training process.

We also focus on quantifying the variability in the network predictions using DropBlocks. 
Using DropBlock layers not only during training, but also during testing and validation (and also investigating the impact of their locations and count) improves the accuracy of each MC-DropBlock realization, leading to more robust uncertainty estimates and reduced variability.
We also investigated how the location and the number of DropBlock layers affect prediction uncertainty. Adding multiple DropBlocks after each convolutional layer, while still providing a shared parameterization across all fidelities, appear to maximize the uncertainty due to variability in the network architecture. However, this was found to produce excessive regularization and reduced accuracy for the decoder-encoder network used for one-dimensional regression.
DropBlock masks that are independently applied or shared across feature channels were found to produce similar accuracy and uncertainty estimates, as well as the use of ReLU or tanh activation functions.

This proposed approach is not without limitations. First, all the results presented in this paper are based on optimal hyperparameters selected through a systematic grid search. Improved results could in principle be obtained from continuous parameter optimization.
Also, inclusion of a large number of LF models, possibly non ordered hierarchically, in the proposed network would require some architectural changes. This could be realized by increasing the depth (number of downsampling blocks) of the network, assuming the input resolution is greater than $2^{\text{depth}}$. In addition, the order of the LF predictors could reflect other aspects besides resolution, such as which LF has the highest correlation with the HF or the highest accuracy.

Future work will be devoted to investigating the performance of additional network layouts, testing new encoder-decoder configurations and on comparing the variance reduction of MC-DropBlock with multifidelity Monte Carlo estimators like the ones covered in~\cite{Gorodetsky2020}. 

\appendix

\section{Network implementation details}

\subsection{One-dimensional regression}

\noindent The architectural details for this network are reported in Table~\ref{tab:oned_convolution_kernels}.
\begin{table}[!h]
\centering
\begin{tabular}{cccc}
\toprule
{\bf Component} & {\bf Layers} & {\bf Kernels per convolution layer} & {\bf Kernel}\\
\midrule
{\bf Encoder} & 2 CBTD-CBTD-U layers & 16 16 8 8 & 2 2 1 1$^{*}$\\
{\bf Decoder} & 2 CBTD-CBTD-M layers & 8 8 16 16 1 & 1 1 2 2$^{*}$\\
{\bf Nonlinear correlation} & 1 CBT-C layer & 8 1 & 2 1$^{*}$\\ 
{\bf Linear correlation} & 1 C layer &  1 & 1$^{*}$\\ 
\bottomrule
\end{tabular}
\caption{Architectural details for one dimensional regression network (see Figure~\ref{LMP:fig:low_multifidelity_network}). Layers consist optionally of various components including convolution (C) layer, batch normalization (B) layer, ReLU activation (R), tanh activation (T), max pooling (M) layer, upsampling (U) layer, and DropBlock/dropout (D) layer. (*) a stride of 1 and appropriate padding are used on all convolution layers to produce layer inputs and output of the same size.}
\label{tab:oned_convolution_kernels}
\end{table}

\subsection{Dense regression}\label{sec:app:dense}

\noindent Additional details on the architecture of the dense regression network are reported in Table~\ref{tab:dense_convolution_kernels}. 

\begin{table}[!h]
\centering
\begin{tabular}{cccc}
\toprule
{\bf Component} & {\bf Layers} & {\bf Kernels per convolution layer} & {\bf Kernel}\\
\midrule
{\bf Encoder} & 4 CBRD-CBRD-M layers  & 16 16 32 32 64 64 128 128 & 3x3-s1-p1\\
{\bf Decoder} & 1 CBRD-CBRD-U, 3 CBR-CBR-U, 2 C layers & 64 64 32 32 16 16 32 32 16 1 & 3x3-s1-p1\\
{\bf Low-fidelity outputs} & CBR-C & 16 1 & 3x3-s1-p1\\ 
\bottomrule
\end{tabular}
\caption{Number of convolutional layers and channels for dense regression network in Figure~\ref{LMP:fig:multifidelity_network_2d}. The notation \emph{nxn-sm-pk} indicates a square kernel of size $n$ with stride of $m$ pixels and padding of $k$ pixels.}
\label{tab:dense_convolution_kernels}
\end{table}

\begin{table}[]
    \centering
    \begin{tabular}{cccccc}
    \toprule
         {\bf Fidelity} & {\bf Skip} & {\bf LR} & {\bf LR scheduler steps} & {\bf Drop scheduler steps} & {\bf Drop probability}\\
         \midrule
         Explicit & Concat & $1 \times 10^{-2}$ & 200 & None & 0.1 \\
         Explicit & Add & $5 \times 10^{-2}$ & 200 & None & 0.3 \\
         \midrule
         Implicit & Concat & $1 \times 10^{-2}$ & 500 & 300  & 0.1 \\
         Implicit & Add & $1 \times 10^{-2}$ & 1000 & 300  & 0.3 \\
         \midrule
         HF 32/0 & Concat & $2\times 10^{-2}$ & 1000 & None & 0.3 \\
         HF 32/0 & Add & $1\times 10^{-2}$ & 1000 & None & 0.1 \\
         \midrule
         HF 116/0 & Concat & $5 \times 10^{-2}$ & 500 & None & 0.1 \\
         HF 116/0 & Add & $2 \times 10^{-2}$ & 200 & 300 & 0.1 \\
         \bottomrule
    \end{tabular}
    \caption{Dense regression network. Hyperparameters producing the best validation accuracy on the mean prediction for each HF and MF model, for the case where DropBlocks are independent across channels.}
    \label{tab:hyperparameters_encdec}
\end{table}

\begin{table}[]
    \centering
    \begin{tabular}{cccccc}
         \toprule
         {\bf Fidelity} & {\bf Skip} & {\bf LR} & {\bf LR scheduler steps} & {\bf Drop scheduler steps} & {\bf Drop probability}\\
         \midrule
         Explicit & Concat & $1 \times 10^{-2}$ & 200 & None & 0.1 \\
         Explicit & Add & $1 \times 10^{-2}$ & 200 & 300 & 0.3 \\
         \midrule
         Implicit & Concat & $1\times 10^{-2}$ & 500 & None  & 0.1 \\
         Implicit & Add & $1 \times 10^{-2}$ & 200 & 300  & 0.5 \\
         \midrule
         HF 32/0 & Concat & $2\times 10^{-2}$ & 500 & 300 & 0.3 \\
         HF 32/0 & Add & $1\times 10^{-2}$ & 200 & None & 0.5 \\
         \midrule
         HF 116/0 & Concat & $1 \times 10^{-2}$ & 1000 & 300 & 0.3 \\
         HF 116/0 & Add & $1 \times 10^{-2}$ & 200 & 300 & 0.5 \\
         \bottomrule
    \end{tabular}
    \caption{Dense regression network. Hyperparameters producing the best validation accuracy on the mean prediction for each HF and MF model, for the case where DropBlocks are shared across channels.}
    \label{tab:hyperparameters_encdec_shareddropblock}
\end{table}

\subsection{Low-to-high dimensional regression}

\noindent The number of channels input to the final convolutional layer is defined in the last column of Table~\ref{tab:hyperparameters_decoder}, where preceding layers increase by a factor of 2 (e.g. for a value of $k$, the number of kernels per convolution layer would be $k\times 2^5, k\times 2^5, k\times 2^4, k\times 2^4, k\times 2^3, k\times 2^3, k\times 2^2, k\times 2^2, k\times 2^1, k\times 2^1, k\times 2^0, k\times 2^0, k \times 2^0, 1$). The number of kernels for the two convolutions immediately preceding the low-fidelity outputs are the same as the two kernels immediately preceding the HF prediction. 
Hyperparameters are shown in Table~\ref{tab:hyperparameters_decoder}. Otherwise, we use the same hyperparameters specified in Section~\ref{sec:app:dense}. 
\begin{table}[!h]
\centering
\resizebox{\textwidth}{!}{
\begin{tabular}{cccc}
\toprule
{\bf Component} & {\bf Layers} & {\bf Kernels per convolution layer} & {\bf Kernel}\\
\midrule
{\bf Decoder} & 3 CBRD-CBRD-U, 3 CBR-CBR-U, 1 CBR-C & 128 128 64 64 32 32 16 16 8 8 4 4 4 1 &  3x3-s1-p1\\
{\bf Low-fidelity outputs} & CBR-C & 4 1 & 3x3-s1-p1\\ 
\bottomrule
\end{tabular}}
\caption{Number of convolutional layers and channels for low- to high-dimensional regression network in Figure~\ref{LMP:fig:multifidelity_network_decoder}. The notation \emph{nxn-sm-pk} indicates a square kernel of size $n$ with stride of $m$ pixels and padding of $k$ pixels.}
\label{tab:low-to-high_convolution_kernels}
\end{table}

\begin{table}[]
    \centering
    \begin{tabular}{cccccc}
    \toprule
         {\bf Fidelity}  & {\bf LR} & {\bf LR scheduler steps} & {\bf Drop scheduler steps} & {\bf Drop probability} & {\bf Filters} \\
         \midrule
         Explicit & $2\times 10^{-2}$ & 500 & 300 & 0.1 & 6 \\
         Implicit  & $1 \times 10^{-2}$ & 1000 & None  & 0.3 & 4 \\
         \midrule
         HF 32/0 & $2 \times 10^{-2}$ & 500 & None & 0.5 & 5 \\
         HF 116/0  & $2\times 10^{-2}$ & 1000 & 300 & 0.1 & 4 \\
         \bottomrule
    \end{tabular}
    \caption{Decoder network. Hyperparameters producing the best validation accuracy on the mean prediction for each HF and MF model.}
    \label{tab:hyperparameters_decoder}
\end{table}

\section{Implementation details for DropBlock layers}

\noindent We observed how the increased regularization produced by additional DropBlock layers could act as a valid substitute for $\ell_{1}$ or $\ell_{2}$ regularization applied directly to the loss function. This was found particularly useful for the dataset in Eqs.~\eqref{LMP:equ:example2_lf} and ~\eqref{LMP:equ:example2_hf}, characterized by a lack of overlap between the LF and HF training locations.

Some additional constraints have to be considered in the selection of suitable locations for DropBlock layers, in order to maintain a \emph{shared parameterization} for all HF and LF outputs. We choose not to include DropBlock layers at locations which would induce stochasticity only to a subset of the HF or LF predictors, i.e. we omit DropBlock after any convolution layers following the LF network outputs.
Consistent with these constraints, we apply the same binary mask both prior to the skip connection and prior to the pooling layer at each stage of the encoder of the network in Fig.~\ref{LMP:fig:multifidelity_network_2d}.
%

\subsection{One dimensional regression}

\noindent We consider DropBlocks after the first eight convolution layers (see Fig.~\ref{LMP:fig:low_multifidelity_network}), since these layers precede the LF prediction. However, since these result in very inaccurate predictions, we exclude as few of the dropout layers as results in accurate dropout realizations; for the dataset in Eq.~\eqref{LMP:equ:example1_hf}, we use DropBlock layers 3-8 and for the dataset in Eq.~\eqref{LMP:equ:example2_hf}, we use DropBlock layers 5-8 (Fig.~\ref{LMP:fig:low_multifidelity_network}).
For the DropBlock layers, we use a drop probability of 0.1 and a block size of 1, due to the low dimensionality of the layers. A block size of 1 would be equivalent to a dropout layer, since $\gamma = p$ from Eq.~\eqref{LMP:equ:dropblock_gamma} and single pixels in the feature map are dropped independently. We use no scheduler for the drop probability.
The DropBlock mask is independent across feature channels due to the small dimensionality of the layers. 
Each convolutional layer is followed by a \emph{tanh} activation, with the exception of the final layer where we use a linear activation.

\begin{table}[!ht]
\begin{center}
\resizebox{0.8\textwidth}{!}{
\begin{tabular}{c c c c c | c c c c c}
\toprule
$\boldsymbol{b}$ & $\boldsymbol{p}$ &   $\boldsymbol{F}$ &     $\boldsymbol{\gamma}$ &   {\bf Actual drop ratio} & $\boldsymbol{b}$ & $\boldsymbol{p}$ &  $\boldsymbol{F}$ &     $\boldsymbol{\gamma}$ &   {\bf Actual drop ratio}\\
\midrule
3 &     0.2 &           3 &   0.2   &          0.203 & 5 &         0.2 &           8 &   0.08  &          0.183 \\

3 &     0.9 &           3 &   0.9   &          0.901            & 5 &         0.9 &           8 &   0.36  &          0.631 \\
\midrule
3 & 0.2 &           4 &   0.133 &          0.19  & 5 &         0.2 &          16 &   0.053 &          0.184 \\

3 &     0.9 &           4 &   0.6   &          0.718 & 5 &         0.9 &          16 &   0.24  &          0.609 \\
\midrule
3 &     0.2 &           8 &   0.089 &          0.186 & 7 &         0.2 &           8 &   0.114 &          0.196 \\

3 &     0.9 &           8 &   0.4   &          0.652 &             7 &         0.9 &           8 &   0.514 &          0.701 \\
\midrule
3 &     0.2 &          16 &   0.076 &          0.186 &  7 &         0.2 &          16 &   0.046 &          0.178 \\

3 &     0.9 &          16 &   0.343 &          0.652 &             7 &         0.9 &          16 &   0.206 &          0.589 \\
\bottomrule
\end{tabular}}
\end{center}
\caption{Dropout probability vs. the average ratio of features dropped. This latter is computed across 1000 DropBlock realizations and averaged. This is evaluated for a one-dimensional layer, where there are 8 channels, and masks are independent across feature channels. The feature size $F$ represents the number of dimensions in each channel. We exclude all partial blocks to be more consistent with \cite{ghiasi2018dropblock}. }
\label{LMP:tab:dropout_vs_dropblock}
\end{table}

\subsection{Dense regression}

\noindent We investigated DropBlock layers that are shared or independent across feature channels. Although shared masks were found to work well when only using a single DropBlock layer, in this work we focus on independent masks, with some uncertainty results related to shared masks. Additionally, DropBlock layers use a block size of 3, in combination with a linear scheduler for the drop probability, with parameters as specified in Table~\ref{tab:hyperparameters_encdec} and~\ref{tab:hyperparameters_encdec_shareddropblock}. DropBlock layers are included after the first ten convolution layers, preceding the LF1 output, as shown in Fig.~\ref{LMP:fig:multifidelity_network_2d}.

\subsection{Low-to-high dimensional regression}

\noindent For low- to high-dimensional regression, we apply DropBlocks after the first six convolution layers of Fig.~\ref{LMP:fig:multifidelity_network_decoder} preceding the LF1 output. We use block sizes of 1, 1, 1, 1, 3, 3, respectively, from the first to the last DropBlock layer and no linear scheduler for the drop probability.

\section{Additional implementation details}

\subsection{Hyperparameter search}

\noindent The results in Section~\ref{LMP:sec:results} are obtained using combinations of hyperparameters producing minimal validation losses.
These hyperparameters include the learning rate, step size of the learning rate scheduler, number of filters, regularization penalty, weight initialization scheme, batch size, optimizer, DropBlock location and drop probability.

\subsection{Accuracy metric}\label{sec:accuracy}

\noindent Prediction accuracy is evaluated on the test dataset and reported as the coefficient of determination $R^2$, within the true fluid region, 
\begin{equation}\label{equ:ch3_rsquared}
R^2 = 1 - \text{RSE} = 1 - \frac{\sum_{i=1}^{N}(y_i - \hat{y}_i)^2}{\sum_{i=1}^{N} (y_i - \Bar{y})^2},
\end{equation}
where $\text{RSE}$ is the relative squared error, $y_i$ the true pixel value with mean $\bar{y}$, $\hat{y}_i$ the network mean prediction (mean over $N_{\text{UQ}}=1,000$ DropBlock realizations), and $N$ is calculated across the entire test set (i.e., the number of pixels in the fluid across all test samples).
We also report a normalized accuracy with respect to the cost of generating the training data set, quantified, in this study, as the total number of pixels in the data.
Thus, 116 HF images with resolution $64\times 64$ have an equivalent cost of $475,136$ pixels, while 32 HF images have an equivalent cost of $131,072$ pixels (a cost ratio of 0.276 to the 116 HF case). The multifidelity dataset with 32 HF images and 116 images for each LF would instead result in a cost of $286,976$ pixels (a cost ratio of 0.604 to the 116 HF case).

\subsection{One-dimensional network}

\subsubsection{Dataset preprocessing}

\noindent We use the same training data as detailed in~\cite{meng2020composite} and a test set consisting of 101 equally spaced points in the interval $[0,1]$. Since a validation set does not exist, we also optimize for the best fit over the training set.
In addition, the functions are rescaled to the range $[0,1]$ based on the maximum and minimum value in the training set, which allows for a consistent use of the same optimizer across functions. 

\subsubsection{Regularization}

\noindent In \cite{PartinCSRI2021}, we carefully selected the $\ell_2$ regularization penalty to constrain the relationship between the LF and HF and therefore, in general, its value may depend on the dataset. However, in this paper, the use of extra DropBlock layers precluded the necessity of applying $\ell_2$ regularization. 

\subsubsection{Weight optimization and batch size}

\noindent Training is performed using the Adam optimizer~\cite{kingma2017adam} utilizing a step learning rate scheduler with decay $0.9$, where the step size and initial learning rate is determined by optimizing for the best fit of the given training dataset. We use a learning rate of $9\times 10^{-4}$ and a step size of $450$.
A batch size of 1 was used, since gradient updates based on a small batch size were found to significantly improve prediction accuracy.

\subsubsection{Weight initialization}

\noindent The network weights are initialized from a a uniform distribution $U(-s,s)$, with $s=n k_0 k_1$, $n$ being the number of input channels, and $k = (k_0, k_1)$ being the kernel shape.

\subsection{Two-dimensional networks}

\subsubsection{Dataset preprocessing}

\noindent For the dense and low- to high-dimensional cases, the dataset consists of two-dimensional slices from a Hagen-Poiseuille flow in a cylindrical fluid domain $\Omega_{f}$. 
The solution is axisymmetric and therefore equal with respect to any plane that includes the cylinder axis; therefore a two-dimensional slice is sufficient to fully describe the flow.

\subsubsection{Single- and multifidelity dataset selection}

\noindent The Hagen-Poiseuille flow dataset (denoted as \emph{HF 116/0}) consists of 116 HF realizations corresponding to random values of the maximum velocity and cylinder radius parameters (see Fig.~\ref{LMP:fig:poiseuille_testcase}).
Training, validation and testing datasets are obtained using 60/20/20 split ratios, resulting in 116, 49 and 35 HF images, respectively. 
For each sample, we generate a uniformly random floating point, which determines to which set that sample belongs based on their respective probabilities; therefore, exact ratios are not preserved.
A second dataset (\emph{HF 32/0}) was also generated by randomly subsampling 32 of the 116 HF realizations.

The multifidelity training dataset consists of the 32-sample HF dataset combined with 116 samples from each of three LF resolutions. These resolutions result from subsampling the 116 HF images using dimensions 32x32, 16x16, and 8x8.
This results in a total of $116\cdot 3 + 32=380$ images. For each coarse representation $\text{LF}_i,\,i\in\{1,2,3\}$, we add uniform random noise from $\mathcal{U}(0, 0.05\cdot[\max_{j,k} \text{LF}_i(j,k) - \min_{j,k} \text{LF}_i(j,k)])$.
After training, every dataset's accuracy is evaluated on the same validation and test sets described above, i.e., 49 and 35 HF images, respectively.
Final accuracy results are reported for predictions on the test set, using the model with the lowest validation loss during training.

\subsubsection{Multifidelity loss}

\noindent The training loss consists of the integral of the Mean Square Error (MSE), assembled from equal contributions (penalty 1/4) of all four fidelities. The integral here is obtained by multiplying each pixel's contribution to the MSE by its size, and it is used to compensate for the different number of pixels present at different fidelities.
We also tested a loss formulation where larger penalties were applied to the high-fidelity samples. While this showed some improvements in the final high-fidelity accuracy, the results were not consistently better than with equal penalties across different network weight initializations.

\subsubsection{Weight optimization and batch size}

\noindent Training is performed using the Adam optimizer with a step learning rate scheduler, which decays by a factor 0.9 every $s$ epochs (where $s$ is the LR scheduler step size), with a batch size of 16. Hyperparameters selected are shown in Table~\ref{tab:hyperparameters_encdec} and~\ref{tab:hyperparameters_encdec_shareddropblock}.

\subsubsection{Weight initialization}

\noindent We utilize the Xavier initialization scheme~\cite{pmlr-v9-glorot10a}, where weights are generated using realizations from a normal distribution, with no significant changes in the results.

\section*{Acknowledgments}

\noindent This work was supported by a NSF CAREER award \#1942662 (PI DES), a NSF CDS\&E award \#2104831 (University of Notre Dame PI DES) and used computational resources provided through the Center for Research Computing at the University of Notre Dame.  
Sandia National Laboratories is a multimission laboratory managed and operated by National Technology and Engineering Solutions of Sandia, LLC., a wholly owned subsidiary of Honeywell International, Inc., for the U.S. Department of Energy's National Nuclear Security Administration under contract DE-NA-0003525. The views expressed in the article do not necessarily represent the views of the U.S. Department of Energy or the United States Government. 
The authors would like to thank Ishani Aniruddha Karmarkar for her assistance with testing fully connected multifidelity network implementations proposed in the literature.

\bibliographystyle{unsrtnat}
\bibliography{main}

\end{document}